\documentclass[runningheads]{llncs}
\usepackage[T1]{fontenc}
\usepackage{cite}
\usepackage{amsmath,amssymb,amsfonts}

\usepackage{algorithmic}
\usepackage{textcomp}
\usepackage{xcolor}
\usepackage[algoruled,linesnumbered,noend,noline]{algorithm2e}%
\usepackage{colortbl}
\usepackage{makecell}
\usepackage{multirow}
\usepackage{enumitem}
\usepackage{hhline}
\usepackage{hyperref}
\usepackage{color}

\usepackage{tikz}
\usepackage{booktabs}
\usepackage{subfigure}
\usepackage{bbding}
\usepackage{graphicx}
\usepackage{color}

\usetikzlibrary{automata,arrows,positioning}
\definecolor{mygray}{gray}{.85}

\usepackage[skip=6pt,font=small,labelfont=bf]{caption}

\newcommand{\bs}[1]{{\mathbf{#1}}}
\newcommand{\mb}{\mathcal{N}}
\newcommand{\tool}{{\sf QEBVerif}\xspace} 
\newcommand{\up}{\mathcal{C}^\text{ub}_h}

\newcommand{\SinQ}{S^{in}(\tilde{\bs{x}}^i_j)}
\newcommand{\SinD}{S^{in}(\bs{x}^i_j)}

\newcommand{\hl}[1]{\textcolor{blue}{#1}}

\begin{document}
%
\title{\tool: Quantization Error Bound Verification of Neural Networks} 
%
%
\author{Yedi Zhang\inst{1} \and
	Fu Song\inst{1,2,3}\textsuperscript{(\Envelope)} \and
	Jun Sun\inst{4}}
\authorrunning{Y. Zhang et al.}
%
\institute{ShanghaiTech University,  Shanghai 201210, China  
 \and 
Institute of Software, Chinese Academy of Sciences \& University of Chinese Academy of
Sciences, Beijing 100190, China \and Automotive Software Innovation Center, Chongqing 400000, China \and
	Singapore Management University, Singapore 178902} 
\maketitle              
\begin{abstract}
To alleviate the practical constraints for deploying deep neural networks (DNNs) on edge devices, quantization is widely regarded as one promising technique. It reduces the resource requirements for computational power and storage space by quantizing the weights and/or activation tensors of a DNN into lower bit-width fixed-point numbers, resulting in quantized neural networks (QNNs).
While it has been empirically shown to introduce minor accuracy loss, critical verified properties of a DNN might become invalid once quantized. Existing verification methods focus on either individual neural networks (DNNs or QNNs) or quantization error bound for \emph{partial} quantization. In this work, we propose a quantization error bound verification method, named \tool, where both weights and activation tensors are quantized. \tool consists of two parts, i.e., a differential reachability analysis (DRA) and a mixed-integer linear programming (MILP) based verification method. DRA performs difference analysis between the DNN and its quantized counterpart layer-by-layer to compute a tight quantization error interval efficiently. If DRA fails to prove the error bound, then we encode the verification problem into an equivalent MILP problem which can be solved by off-the-shelf solvers. Thus, \tool is sound, complete, and reasonably efficient. We implement \tool and conduct extensive experiments, showing its effectiveness and efficiency.

\end{abstract}

\section{Introduction}\label{sec:intro}
In the past few years, the development of deep neural networks (DNNs) has grown at an impressive pace owing to their outstanding performance in solving various complicated tasks~\cite{KTSLSF14,HDYDMJSVNSK12}. However, modern DNNs are often large in size and contain a great number of 32-bit floating-point parameters to achieve competitive performance. Thus, they often result in high computational costs and excessive storage requirements, hindering their deployment on resource-constrained embedded devices, e.g., edge devices.
A promising solution is to quantize the weights and/or activation tensors as fixed-point numbers of lower bit-width~\cite{HanMD15,LinTA16,DSQ,JacobKCZTHAK18}. For example, TensorFlow Lite~\cite{TFLiteWebpage} supports quantization of weights and/or activation tensors to reduce model size and latency, and Tesla FSD-chip~\cite{FSDChip} stores all the data and weights of a network in the form of 8-bit integers.

In spite of the empirically impressive results which show there is only minor accuracy loss, quantization does not necessarily preserve properties such as robustness~\cite{GiacobbeHL20}.
Even worse, input perturbation can be amplified by quantization~\cite{lin2018defensive,duncan2020relative}, worsening the robustness of quantized neural networks (QNNs) compared to their DNN counterparts.
Indeed, existing neural network quantization methods focus on minimizing its impact on model accuracy (e.g., by formulating it as an optimization problem that aims to maximize the accuracy~\cite{jung2019learning,nagel2020up}).
However, they cannot guarantee that the final quantization error is always lower than a given error bound, especially when some specific safety-critical input regions are concerned. This is concerning as such errors may lead to catastrophes when the quantized networks are deployed in safety-critical applications~\cite{EEF0RXPKS18,julian2019deep}.
Furthermore, analyzing (in particular, quantifying) such errors can also help us understand how 
quantization affect the network behaviors~\cite{li2021robust}, and provide insights on, for instance, how to choose appropriate quantization bit sizes without introducing too much error. Therefore, a method that soundly quantifies the errors between DNNs and their quantized counterparts is highly desirable.

There is a large and growing body of work on developing verification methods for DNNs~\cite{PT10,Ehl17,KBDJK17,KatzHIJLLSTWZDK19,HKWW17,SGPV19,ElboherGK20,YLLHWSXZ20,LiuSZW20,GuoWZZSW21,LXSSXM22,AndersonPDC19,GMDTCV18,LiLYCHZ19,SinghGPV19,WPWYJ18,TranBXJ20,TranLMYNXJ19} and QNNs~\cite{narodytska2018verifying,NPAQ19,GiacobbeHL20,AWBK20,BDD4BNN,scaleQNN21,zhang2022qvip},
aiming to establish a formal guarantee on the network behaviors. However, all the above-mentioned methods focus exclusively on verifying individual neural networks.
Recently, Paulsen et al.~\cite{paulsen2020reludiff,PaulsenWWW20} proposed differential verification methods,
aimed to establish formal guarantees on the difference between two DNNs. Specifically, given two DNNs $\mathcal{N}_1$ and $\mathcal{N}_2$ with the same network topology and inputs,
they try to prove that $|\mathcal{N}_1(\bs{x})-\mathcal{N}_2(\bs{x})|<\epsilon$ for all possible inputs $\bs{x}\in \mathcal{X}$, where $\mathcal{X}$ is the interested input region.
They presented fast and sound difference propagation techniques followed by a refinement of the input region until the property can be successfully verified, i.e., the property is either proved or falsified by providing a counterexample. This idea has been extended to handle recurrent neural networks (RNNs)~\cite{MohammadinejadP21} though the refinement is not considered therein.
Although their methods~\cite{paulsen2020reludiff,PaulsenWWW20,MohammadinejadP21} can be used to analyze the error bound introduced by quantizing weights (called \emph{partially} QNNs), they are not complete and
cannot handle the cases where both the weights and activation tensors of a DNN are quantized to lower bit-width fixed-point numbers (called \emph{fully} QNNs).
 We remark that fully QNN can significantly reduces energy-consumption (floating-point-operations consume much more energy than integer-only-operations)~\cite{FSDChip}.


\noindent
{\bf Main Contributions.} We propose a sound and complete
{\bf Q}uantization {\bf E}rror {\bf B}ound {\bf Verif}ication method (\mbox{\tool}) to efficiently and effectively verify if the quantization error
of a \textit{fully} QNN w.r.t. an input region and its original DNN is always lower than an error bound (a.k.a. robust error bound~\cite{li2021robust}).
\tool first conducts a novel reachability analysis to quantify the quantization errors, which is referred to as \emph{differential reachability analysis} (DRA). Such an analysis yields two results: (1) \emph{Proved}, meaning that the quantization error is proved to be always less than the given error bound; or (2) \emph{Unknown}, meaning that it fails to prove the error bound, possibly due to a conservative approximation of the quantization error. If the outcome is \emph{Unknown}, we further encode this quantization error bound verification problem into an equivalent mixed-integer linear programming (MILP) problem,
which can be solved by off-the-shelf solvers.

There are two main technical challenges that must be addressed for DRA. First, the activation tensors in a fully QNN are discrete values and contribute additional rounding errors to the final quantization errors, which are hard to propagate symbolically and make it difficult to establish relatively accurate difference intervals.
%
Second, much more activation-patterns (i.e., $3\times 6=18$) have to consider in a forward propagation,
while 9 activation-patterns are sufficient in~\cite{paulsen2020reludiff,PaulsenWWW20},
where an activation-pattern indicates the status of the output range of a neuron. A neuron in a DNN under an input region has 3 patterns: always-active (i.e., output $\ge 0$), always-inactive (i.e., output $< 0$), or both possible. A neuron in a QNN has 6 patterns due to the clamp function (cf. Definition~\ref{def:qnn}). We remark that handling these different combinations efficiently and soundly is highly nontrivial.
To tackle the above challenges, we propose sound transformations for the affine and activation functions to propagate quantization errors of two networks layer-by-layer.
Moreover, for the affine transformation, we provide two alternative solutions: \emph{interval-based} and \emph{symbolic-based}.
The former directly computes sound difference intervals via interval analysis~\cite{moore2009introduction},
while the latter leverages abstract interpretation~\cite{CousotC77} to compute sound and symbolic difference intervals, using the polyhedra abstract domain. In comparison, the symbolic-based one is usually more accurate but less efficient than the interval-based one.
Note that though existing tools can obtain quantization error intervals by independently computing the output intervals of two networks followed by interval subtractions, such an approach is often too conservative.
%

To resolve those problems that cannot be proved via our DRA, we resort to the sound and complete MILP-based verification method. Inspired by the MILP encoding of DNN and QNN verification~\cite{LomuscioM17,zhang2022qvip,mistry2022milp}, we propose a novel MILP encoding for verifying quantization error bounds. \tool represents both the computations of the QNN and the DNN in mixed-integer linear constraints which are further simplified using their own output intervals. Moreover, we also encode the output difference intervals of hidden neurons from our DRA as mixed-integer linear constraints to boost the verification.

We implement our method as an end-to-end tool and use Gurobi~\cite{Gurobi} as our back-end MILP solver.
We extensively evaluate it on a large set of verification tasks using neural networks for ACAS Xu~\cite{julian2019deep} and MNIST~\cite{MNIST}, where the number of neurons varies from 310 to 4890,
the number of bits for quantizing weights and activation tensors ranges from 4 to 10 bits,
and the number of bits for quantizing inputs is fixed to 8 bits. For DRA,
we compare \tool with a naive method that first independently computes the output intervals of DNNs and QNNs using
the existing state-of-the-art (symbolic) interval analysis~\cite{SGPV19,scaleQNN21}, and then conducts an interval subtraction.
The experimental results show that both our interval- and symbolic-based approaches are much more accurate and can successfully verify much more tasks without the MILP-based
verification.
We also find that the quantization error interval returned by DRA is getting tighter with the increase of the quantization bit size.
The experimental results also confirm the effectiveness of our MILP-based verification method, which can help verify many tasks that
cannot be solved by DRA solely. Finally, our results also allow us to study the potential correlation of quantization errors and robustness for QNNs using~\tool.

We summarize our contributions as follows:
\begin{itemize}[leftmargin=*]
    \item We introduce the first sound, complete and reasonably efficient quantization error bound verification method \tool for fully QNNs by cleverly combining novel DRA and MILP-based verification methods.
    \item We propose a novel DRA to compute sound and tight quantization error intervals accompanied by an abstract domain tailored to QNNs, which can significantly and soundly tighten the quantization error intervals.
    \item We implement \tool as an end-to-end open-source tool~\cite{toolWEB} and conduct extensive evaluation on various verification tasks,  demonstrating its effectiveness and efficiency.
\end{itemize}

\smallskip
\noindent
{\bf Outline.}
Section~\ref{sec:pre} defines our problem and briefly recap {\scshape DeepPoly}~\cite{SGPV19}.
Section~\ref{sec:qveb} presents our quantization error bound verification method \tool and Section~\ref{sec:abstract} gives our symbolic-based approach used in \tool.
Section~\ref{sec:exp} reports experimental results.
Section~\ref{sec:relatedwork} discusses related work. Finally, we conclude this work in Section~\ref{sec:conclu}.

The source code of our tool and benchmarks are available at
\begin{center}
\url{https://github.com/S3L-official/QEBVerif}.
\end{center}

\section{Preliminaries}\label{sec:pre}
We denote by $\mathbb{R}, \mathbb{Z}, \mathbb{N}$ and $\mathbb{B}$ the sets of real-valued numbers, integers, natural numbers, and Boolean values, respectively.
Let $[n]$ denote the integer set $\{1, \ldots, n\}$ for given $n\in\mathbb{N}$. We use {\bf BOLD UPPERCASE} (e.g., $\bs{W}$) and {\bf bold lowercase} (e.g., $\bs{x}$) to denote  matrices and vectors, respectively.
We denote by $\bs{W}_{i,j}$ the $j$-entry in the $i$-th row of the matrix $\bs{W}$,
and by $\bs{x}_i$ the $i$-th entry of the vector $\bs{x}$. Given a matrix $\bs{W}$ and a vector $\bs{x}$, we use $\widehat{\bs{W}}$ and $\hat{\bs{x}}$ (resp. $\widetilde{\bs{W}}$ and $\tilde{\bs{x}}$) to denote their quantized/integer (resp. fixed-point) counterparts.

\subsection{Neural Networks}

A deep neural network (DNN) consists of a sequence of layers, where the first layer is the \emph{input layer}, the last layer is the \emph{output layer} and the others are called \emph{hidden layers}. Each layer contains one or more neurons. A DNN is \emph{feed-forward} if all the neurons in each non-input layer only receives inputs from the neurons in the preceding layer.

\begin{definition}[Feed-forward Deep Neural Network]\label{def:dnn}
    A feed-forward DNN $\mathcal{N}:\mathbb{R}^n \rightarrow \mathbb{R}^s$ with $d$ layers can be seen as a composition of $d$ functions such that $\mathcal{N}=l_d \circ l_{d-1} \circ \cdots \circ l_1$. Then, given an input $\bs{x}\in\mathbb{R}^n$, the output of the DNN $\bs{y}=\mathcal{N}(\bs{x})$ can be obtained by the following recursive computation:
    \begin{itemize}[leftmargin=*]
        \item Input layer $l_1: \mathbb{R}^n\rightarrow \mathbb{R}^{n_1}$ is the identity function, i.e., $\bs{x}^1=l_1(\bs{x})=\bs{x}$;
        \item Hidden layer $l_i:\mathbb{R}^{n_{i-1}}\rightarrow \mathbb{R}^{n_{i}}$ for $2\le i \le d-1$ is the function such that $\bs{x}^i=l_i(\bs{x}^{i-1})=\phi(\bs{W}^i\bs{x}^{i-1}+\bs{b}^i)$;
        \item Output layer $l_d: \mathbb{R}^{n_{d-1}}\rightarrow \mathbb{R}^s$ is the function such that $\bs{y}=\bs{x}^d=l_d(\bs{x}^{d-1})=\bs{W}^d\bs{x}^{d-1}+\bs{b}^d$;
    \end{itemize}
    where $n_1=n$, $\bs{W}^i$ and $\bs{b}^i$ are the weight matrix and bias vector in the $i$-th layer, and $\phi(\cdot)$ is the activation function which acts element-wise on an input vector.
\end{definition}
In this work, we focus on feed-forward DNNs with the most commonly used activation functions: the rectified linear unit (ReLU) function, defined as $\text{ReLU}(x)=\text{max}(x,0)$.



A quantized neural network (QNN) is structurally similar to its real-valued counterpart, except that all the parameters, inputs of the QNN, and outputs of all the hidden layers are quantized into integers according to the given quantization scheme. Then, the computation over real-valued arithmetic in a DNN can be replaced by the computation using integer arithmetic, or equally, fixed-point arithmetic.
In this work, we consider the most common quantization scheme, i.e., symmetric uniform quantization~\cite{nagel2021white}. We first give the concept of quantization configuration which effectively defines a quantization scheme.

A \emph{quantization configuration} $\mathcal{C}$ is a tuple $ \langle \tau, Q, F\rangle$, where $Q$ and $F$ are the total bit size and the fractional bit size allocated to a value, respectively,
and $\tau\in\{+,\pm\}$ indicates if the quantized value is unsigned or signed.
%
Given a real number $x\in\mathbb{R}$ and a quantization configuration $\mathcal{C}=\langle \tau, Q,F\rangle$, its quantized integer counterpart $\hat{x}$ and the fixed-point counterpart $\tilde{x}$ under the symmetric uniform quantization scheme are:
\begin{center}
    $\hat{x}=\text{clamp}(\lfloor 2^F\cdot x \rceil, \ \mathcal{C}^\text{lb}, \ \mathcal{C}^\text{ub})$ \ and \ $\tilde{x} = \hat{x}/2^F$
\end{center}
where $\mathcal{C}^\text{lb}=0$ and $\mathcal{C}^\text{ub}=2^Q-1$ if $\tau=+$,  $\mathcal{C}^\text{lb}=-2^{Q-1}$ and $\mathcal{C}^\text{ub}=2^{Q-1}-1$ otherwise,
and $\lfloor \cdot \rceil$ is the round-to-nearest integer operator. The \emph{clamping function} $\text{clamp}(x,a,b)$ with a lower bound $a$ and an upper bound $b$ is defined as:
\begin{center}
    $\text{clamp}(x,a,b)=
    \begin{cases}
      a,  & \text{if} \ x<a;\\
      x,      &  \text{if} \ a \le x \le b;\\
      b,   &  \text{if} \ x>b.
    \end{cases}
    $
\end{center}


\begin{definition}[Quantized Neural Network]\label{def:qnn}
    Given quantization configurations for the weights, biases, output of the input layer and each hidden layer as $\mathcal{C}_w=\langle \tau_w, Q_w, F_w\rangle$, $\mathcal{C}_b=\langle \tau_b, Q_b, F_b\rangle$, $\mathcal{C}_{in}=\langle \tau_{in}, Q_{in}, F_{in}\rangle$, $\mathcal{C}_h=\langle \tau_h, Q_h, F_h\rangle$, the quantized version (i.e., QNN) of a DNN $\mathcal{N}$ with $d$ layers
is a function $\widehat{\mathcal{N}}: \mathbb{Z}^n \rightarrow \mathbb{R}^s$ such that $\widehat{\mathcal{N}}=\hat{l}_d \circ \hat{l}_{d-1} \circ \cdots \circ \hat{l}_1$. Then, given a quantized input $\hat{\bs{x}}\in \mathbb{Z}^n$, the output of the QNN $\hat{\bs{y}}=\widehat{\mathcal{N}}(\hat{\bs{x}})$ can be obtained by the following recursive computation:
\begin{itemize}[leftmargin=*]
    \item Input layer $\hat{l}_1: \mathbb{Z}^n\rightarrow\mathbb{Z}^{n_1}$ is the identity function, i.e., $\hat{\bs{x}}^1=\hat{l}_1(\hat{\bs{x}})=\hat{\bs{x}}$;
    \item Hidden layer $\hat{l}_i: \mathbb{Z}^{n_{i-1}}\rightarrow \mathbb{Z}^{n_{i}}$ for $2\le i \le d-1$ is the function such that for each $j\in[n_{i}]$,
    \begin{center}
        $\hat{\bs{x}}_j^i=\text{\emph{clamp}}(\lfloor 2^{F_i}\widehat{\bs{W}}^i_{j,:}\cdot{\hat{\bs{x}}}^{i-1} + 2^{F_h-F_b} \hat{\bs{b}}_j^i \rceil, 0, \ \mathcal{C}_h^{\text{\emph{ub}}})$,
        \end{center}
        where $F_i$ is $F_h-F_w-F_{in}$ if $i=2$, and $-F_w$ otherwise;
    \item Output layer $\hat{l}_d:\mathbb{Z}^{n_{d-1}}\rightarrow \mathbb{R}^s$ is the function such that $\hat{\bs{y}}=\hat{\bs{x}}^d = \hat{l}_d (\hat{\bs{x}}^{d-1})= 2^{-F_w}\widehat{\bs{W}}^d\hat{\bs{x}}^{d-1}+ 2^{F_h-F_b}\hat{\bs{b}}^d$;
\end{itemize}
where for every $2\le i \le d$ and $k\in[n_{i-1}]$, $\widehat{\bs{W}}^i_{j,k}=\text{\emph{clamp}}(\lfloor 2^{F_w}\bs{W}^i_{j,k}\rceil, \mathcal{C}_w^{\text{\emph{lb}}}, \mathcal{C}_w^{\text{\emph{ub}}})$ is the quantized
        weight and $\hat{\bs{b}}^i_j=\text{\emph{clamp}}(\lfloor 2^{F_b} \bs{b}^i_j \rceil, \mathcal{C}_b^\text{\emph{lb}},\mathcal{C}_b^\text{\emph{ub}})$ is the quantized bias.
\end{definition}
We remark that $2^{F_i}$ and $2^{F_h-F_b}$ in Definition~\ref{def:qnn} are used to align the precision between the inputs and outputs of hidden layers, and $F_i$ for $i=2$ and $i>2$ because quantization bit sizes for the outputs of the input layer and hidden layers can be different.


\begin{figure}[t]
	\centering
	\subfigure[DNN $\mb_{e}$]{\label{fig:dnnDemo}
		\begin{minipage}[b]{0.365\textwidth}\centering
			\includegraphics[width=.9\textwidth]{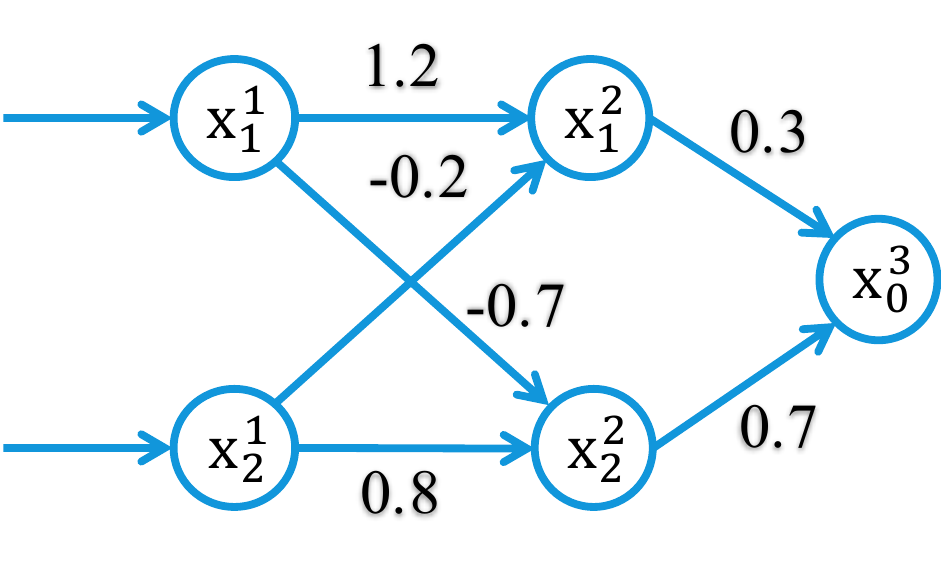}
		\end{minipage}	
	}
	\subfigure[QNN $\widehat{\mb}_{e}$]{\label{fig:qnnDemo}
		\begin{minipage}[b]{0.36\textwidth}\centering
			\includegraphics[width=0.9\textwidth]{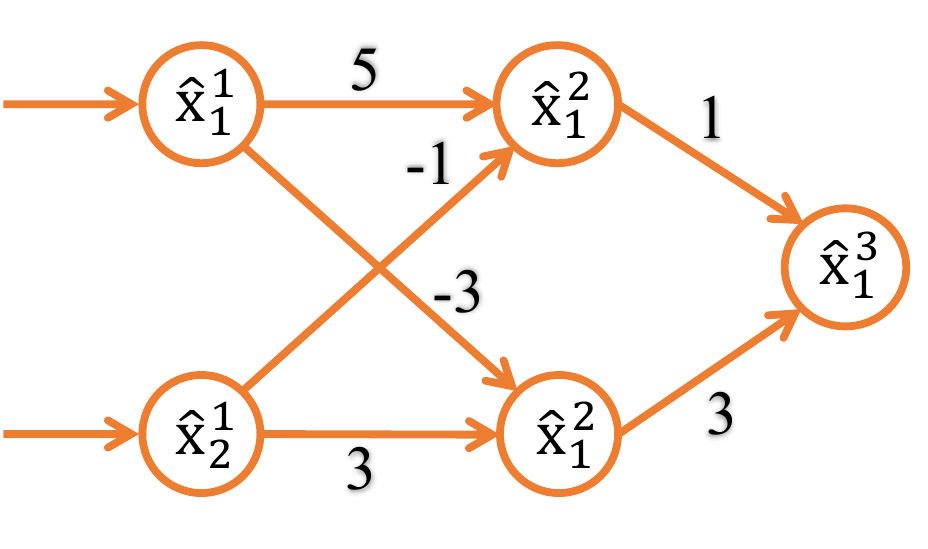}
		\end{minipage}	
	}
	 \vspace*{-1mm}
	\caption{A 3-layer DNN ${\mb}_{e}$ and its quantized version $\widehat{\mb}_{e}$.}
	 \vspace*{-3mm}
    \label{fig:nnVS}
\end{figure}

 \subsection{Quantization Error Bound and its Verification Problem}
We now give the formal definition of the quantization error bound verification problem considered in this work as follows.


\begin{definition}[Quantization Error Bound]\label{def:reb}
    Given a DNN $\mathcal{N}:\mathbb{R}^n\rightarrow \mathbb{R}^s$, the corresponding QNN $\widehat{\mathcal{N}}:\mathbb{Z}^n\rightarrow \mathbb{R}^s$, a quantized input $\hat{\bs{x}}\in\mathbb{Z}^n$, a radius $r\in\mathbb{N}$ and an error bound $\epsilon\in\mathbb{R}$. The QNN $\widehat{\mathcal{N}}$ has a quantization error bound of $\epsilon$ w.r.t. the input region $R(\hat{\bs{x}},r)=\{\hat{\bs{x}}'\in\mathbb{Z}^n \mid || \hat{\bs{x}}'-\hat{\bs{x}}||_\infty \le r \}$ if for every $\hat{\bs{x}}'\in R(\hat{\bs{x}},r)$, we have $||2^{-F_h}\widehat{\mathcal{N}}(\hat{\bs{x}}')- \mathcal{N}(\bs{x}')||_\infty<\epsilon$, where $\bs{x}'=\hat{\bs{x}}'/(\mathcal{C}_{in}^\text{\emph{ub}}-\mathcal{C}_{in}^\text{\emph{lb}})$.
\end{definition}
Intuitively, quantization-error-bound is the bound of the output difference of the DNN and its quantized counterpart for all the inputs in the input region. In this work, we obtain the input for DNN via dividing $\hat{\bs{x}}'$ by $(\mathcal{C}_{in}^\text{ub}-\mathcal{C}_{in}^\text{lb})$ to allow input normalization. Furthermore, $2^{-F_h}$ is used to align the precision between the outputs of QNN and DNN.



\begin{example}
    Consider the DNN $\mathcal{N}_e$ with 3 layers (one input layer, one hidden layer, and one output layer) given in Figure~\ref{fig:nnVS}, where weights are associated with the edges and all the biases are 0. The quantization configurations for the weights, the output of the input layer and hidden layer are $\mathcal{C}_w=\langle \pm, 4,2\rangle$, $\mathcal{C}_{in}=\langle +, 4,4\rangle$ and $\mathcal{C}_h=\langle +, 4,2\rangle$. Its QNN $\widehat{\mathcal{N}}_e$ is shown in Figure~\ref{fig:nnVS}.

    Given a quantized input $\hat{\bs{x}}=(9,6)$ and a radius $r=1$, the input region for QNN $\widehat{\mathcal{N}}_e$ is $R((9,6),1)=\{(x,y)\in \mathbb{Z}^2 \mid 8\le x\le 10, 5\le y\le 7\}$.
    Since $\mathcal{C}_{in}^\text{\emph{ub}}=15$ and $\mathcal{C}_{in}^\text{\emph{lb}}=0$,
    by Definitions~\ref{def:dnn}, \ref{def:qnn}, and \ref{def:reb}, we have the maximum quantization error as max$(2^{-2}\widehat{\mathcal{N}}_e(\hat{\bs{x}}')-\mathcal{N}_e(\hat{\bs{x}}'/15))=0.067$ for $\hat{\bs{x}}' \in R((9,6),1)$.
    Then, $\widehat{\mathcal{N}}_e$ has a quantization error bound of $\epsilon$ w.r.t. input region $R((9,6),1)$ for any $\epsilon>0.067$.

    We remark that if only weights are quantized and the activation tensors are floating-point numbers, the maximal quantization error of $\widehat{\mathcal{N}}_e$ for the input region $R((9,6),1)$ is 0.04422, which implies that existing methods~\cite{paulsen2020reludiff,PaulsenWWW20} cannot be used to analyze the error bound for a fully QNN.

\end{example}

In this work, we focus on the quantization error bound verification problem for classification tasks. Specifically, for a classification task, we only focus on the output difference of the predicted class instead of all the classes. Hence,
given a DNN $\mathcal{N}$, a corresponding QNN $\widehat{\mathcal{N}}$, a quantized input $\hat{\bs{x}}$ which is classified to class $g$
by the DNN $\mathcal{N}$, a radius $r$ and an error bound $\epsilon$,  the quantization error bound property $P(\mathcal{N},\widehat{\mathcal{N}},\hat{\bs{x}},r,\epsilon)$ for a classification task
can be defined as follows:
\begin{center}
$\bigwedge_{\hat{\bs{x}}'\in R(\hat{\bs{x}},r)} \big ( |2^{-F_h}\widehat{\mathcal{N}}(\hat{\bs{x}}')_g- \mathcal{N}(\bs{x}')_g|<\epsilon \big ) \wedge \big ( \bs{x}'=\hat{\bs{x}}'/(\mathcal{C}_{in}^\text{ub}-\mathcal{C}_{in}^\text{lb})\big )$.
\end{center}
Note that $\mathcal{N}(\cdot)_g$ denotes the $g$-th entry of the vector $\mathcal{N}(\cdot)$.

\subsection{{\scshape DeepPoly}}\label{sec:deepPoly}
We briefly recap {\scshape DeepPoly}~\cite{SGPV19}, which will be leveraged in this work 
for computing the output of each neuron in a DNN. 

The core idea of {\scshape DeepPoly} is to give each neuron an abstract domain in the form of a linear combination of the variables preceding the neuron. To achieve this, each hidden neuron $\bs{x}^i_j$ (the $j$-th neuron in the $i$-th layer) in a DNN is seen as two nodes $\bs{x}^i_{j,0}$ and $\bs{x}^i_{j,1}$, 
such that $\bs{x}^i_{j,0}=\sum_{k=1}^{n_{i-1}}\bs{W}^i_{j,k}\bs{x}^{i-1}_{k,1}+\bs{b}^i_j$ (affine function) and $\bs{x}^i_{j,1} = \text{ReLU}(\bs{x}^i_{j,0})$ (ReLU function). 
Then, the affine function
is characterized as an abstract transformer using an upper polyhedral computation and a lower polyhedral computation
in terms of the variables $\bs{x}^{i-1}_{k,1}$.
Finally, it recursively substitutes the variables in
the upper and lower polyhedral computations with the corresponding upper/lower polyhedral computations of the variables
until they only contain the input variables from which the concrete intervals are computed.

Formally, the abstract element $\mathcal{A}^i_{j,s}$ for the node $\bs{x}^i_{j,s}$ ($s\in\{0,1\}$) is a tuple $\mathcal{A}^i_{j,s}=\langle\bs{a}^{i,\le}_{j,s},
\bs{a}_{j,s}^{i,\ge}, l^i_{j,s},u^i_{j,s}\rangle$, where $\bs{a}_{j,s}^{i,\le}$ and $\bs{a}_{j,s}^{i,\ge}$
are respectively the lower and upper polyhedral computations in the form of a linear combination of the variables $\bs{x}^{i-1}_{k,1}$'s (if $s=0$) or $\bs{x}^{i}_{k,0}$'s if $s=1$,
$l_{j,s}^i\in\mathbb{R}$ and $u^i_{j,s}\in\mathbb{R}$ are the concrete lower and upper bound of the neuron.
Then, the concretization of the abstract element $\mathcal{A}^i_{j,s}$ is
$\Gamma (\mathcal{A}^i_{j,s})=\{x\in\mathbb{R}\mid \bs{a}^{i,\le}_{j,s} \le x \wedge x\le \bs{a}^{i,\ge}_{j,s}\}$.

Concretely, $\bs{a}_{j,0}^{i,\le}$ and $\bs{a}_{j,0}^{i,\ge}$ are defined as $\bs{a}_{j,0}^{i,\le} =\bs{a}_{j,0}^{i,\ge} = \sum_{k=1}^{n_{i-1}}\bs{W}^i_{j,k}\bs{x}^{i-1}_{k,1}+\bs{b}^i_j$. Furthermore, we can repeatedly substitute every variable in $\bs{a}_{j,0}^{i,\le}$ (resp. $\bs{a}_{j,0}^{i,\ge}$) with its lower or upper polyhedral computation according to the coefficients until no further substitution is possible.
Then, we can get a sound lower (resp. upper) bound in the form of a linear combination of the input variables based on which $l^i_{j,0}$ (resp. $u^i_{j,0}$) can be computed immediately from the given input region.

For ReLU function $\bs{x}^i_{j,1}=\text{ReLU}(\bs{x}^i_{j,0})$, there are three cases to consider of the abstract element $\mathcal{A}^i_{j,1}$:
\begin{itemize}
	\item If $u^i_{j,0} \le 0$, then $\bs{a}_{j,1}^{i,\le}=\bs{a}_{j,1}^{i,\ge}=0$, $l^i_{j,1}=u^i_{j,1}=0$;
	\item If $l^i_{j,0} \ge 0$, then $\bs{a}_{j,1}^{i,\le}= \bs{a}_{j,0}^{i,\le}$, $\bs{a}_{j,1}^{i,\ge}= \bs{a}_{j,0}^{i,\ge}$, $l^i_{j,1}=l^i_{j,0}$ and $u^i_{j,1}=u^i_{j,0}$;
	\item If $l^i_{j,0}<0\wedge u^i_{j,0}>0$, then $\bs{a}_{j,1}^{i,\ge}=\frac{u^i_{j,0}(\bs{x}^i_{j,0}-l^i_{j,0})}{u^i_{j,0}-l^i_{j,0}}$, $\bs{a}_{j,1}^{i,\le}=\lambda \bs{x}^i_{j,0}$ where $\lambda\in\{0,1\}$ such that the area of resulting shape by $\bs{a}_{j,1}^{i,\le}$ and $\bs{a}_{j,1}^{i,\ge}$ is minimal, $l^i_{j,1}=\lambda l^i_{j,0}$ and $u^i_{j,1}=u^i_{j,0}$.
\end{itemize}

Note that {\scshape DeepPoly} also introduces transformers for other functions, such as sigmoid, tanh and maxpool functions. In this work, we only consider DNNs with only ReLU as non-linear operators. A simple example of {\scshape DeepPoly} is given in Appendix~\ref{app_sec:deepPoly}.

\section{Methodology of \tool}\label{sec:qveb}

In this section, we first give an overview of our quantization error bound verification method, \tool, and then give the detailed design of each component. First, let review

\subsection{Overview of \tool}
\begin{figure}[t]
	\centering
	\includegraphics[width=.7\textwidth]{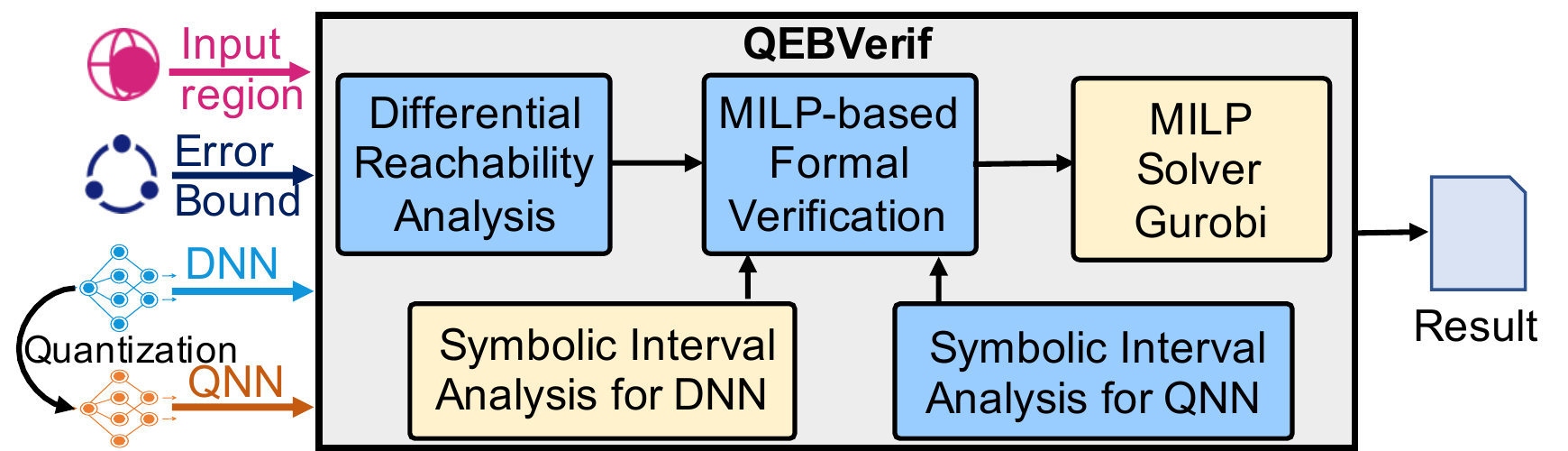}
	\caption{An overview of \tool.}\label{fig:overview}
\end{figure}

An overview of \tool is shown in Figure~\ref{fig:overview}. Given a DNN $\mathcal{N}$, its QNN $\widehat{\mathcal{N}}$, a quantization error bound $\epsilon$
and an input region consisting of a quantized input $\hat{\bs{x}}$ and a radius $r$,
to verify the quantization error bound property $P(\mathcal{N},\widehat{\mathcal{N}},\hat{\bs{x}},r,\epsilon)$,
\tool first performs a differential reachability analysis (DRA) to compute a sound output difference interval for the two networks. Note that, the difference intervals of all the neurons are also recorded for later use.
If the output difference interval of the two networks is contained in $[-\epsilon, \epsilon]$, then the property is proved and \tool outputs ``Proved''.
Otherwise, \tool leverages our MILP-based quantization error bound verification method by encoding the problem into an equivalent mixed integer linear programming (MILP) problem which can be solved by off-the-shelf solvers. To reduce the size of mixed integer linear constraints and boost the verification,
\tool independently applies symbolic interval analysis on the two networks based on which some activation patterns could be omitted.
We further encode the difference intervals of all the neurons from DRA as mixed integer linear constraints and add
them to the MILP problem. Though it increases the number of mixed integer linear constraints, it is very helpful for solving hard verification tasks.
Therefore, the whole verification process is sound, complete yet reasonably efficient. We remark that the MILP-based verification method is often more time-consuming and thus the first step allows us to quickly verify many tasks first.

\subsection{Differential Reachability Analysis}
Naively, one could use an existing verification tool in the literature to independently compute the output intervals for both the QNN and the DNN, and then compute their output difference directly by interval subtraction. However, such an approach would be ineffective due to the significant precision loss.

Recently, Paulsen et al. \cite{paulsen2020reludiff} proposed {\scshape ReluDiff} and showed that the accuracy of output difference for two DNNs can be greatly improved by propagating the difference intervals layer-by-layer. For each hidden layer, they first compute the output difference of affine functions (before applying the ReLU), and then they use a ReLU transformer to compute the output difference after applying the ReLU functions.
The reason why {\scshape ReluDiff} outperforms the naive method is that {\scshape ReluDiff} first computes part of the difference before it accumulates. {\scshape ReluDiff} is later improved to tighten the approximated difference intervals~\cite{PaulsenWWW20}.
However, as mentioned previously, they do not support \emph{fully} quantified neural networks.
Inspired by their work, we design a difference propagation algorithm for our setting.
We use $S^{in}(\bs{x}^i_j)$ (resp. $S^{in}(\hat{\bs{x}}^i_j)$) to denote the interval of the $j$-th neuron in the $i$-th layer in the DNN (resp. QNN) before applying the ReLU function (resp. clamp function), and use $S(\bs{x}^i_j)$ (resp. $S(\hat{\bs{x}}^i_j)$) to denote the output interval after applying the ReLU function (resp. clamp function).
We use $\delta^{in}_{i}$ (resp. $\delta_{i}$) to denote the difference interval for the $i$-th layer before (resp. after) applying the activation functions, and use $\delta^{in}_{i,j}$ (resp. $\delta_{i,j}$) to denote the interval for the $j$-th neuron of the $i$-th layer.
We denote by $\text{LB}(\cdot)$ and $\text{UB}(\cdot)$ the concrete lower and upper bounds accordingly.

\begin{algorithm}[t]
	\scriptsize
	\SetKwInOut{Input}{Input}
	\SetKwInOut{Output}{output}
	\SetInd{1em}{1em}
	\Input{DNN $\mathcal{N}$, QNN $\widehat{\mathcal{N}}$, input region $R(\hat{x},r)$}
	\Output{Output difference interval $\delta$}
	Compute $S^{in}(\bs{x}_j^i)$ and $S(\bs{x}_j^i)$  for $i\in[d-1]$, $j\in[n_i]$ using {\scshape DeepPoly}\;
	Compute $S^{in}(\hat{\bs{x}}_j^i)$ and $S(\hat{\bs{x}}_j^i)$ for $i\in[d-1]$, $j\in[n_i]$ by applying interval analysis~\cite{scaleQNN21}\;

	Initialize the difference: $\delta_1=(2^{-F_{in}}-1/(\mathcal{C}_{in}^\text{ub}-\mathcal{C}_{in}^\text{lb}))S(\hat{\bs{x}}^1)$\;

	\For( \textcolor{gray}{// propagate in hidden layers}){i \emph{in} $2,\ldots,d-1$ }{
	\hspace*{-2mm}	\For{j \emph{in} $1,\ldots, n_i$}{
	\hspace*{-4mm}	$\Delta\bs{b}^i_j= 2^{-F_b}\hat{\bs{b}}^i_j-\bs{b}^i_j$; \ $\xi=2^{-F_h-1}$\;

	\hspace*{-4mm}	$\delta^{in}_{i,j} = \text{\scshape AffTrs}(\bs{W}^i_{j,:},2^{-F_w}\widehat{\bs{W}}^i_{j,:}, \Delta\bs{b}^i_j, S(\bs{x}^{i-1}),\delta_{i-1},\xi)$;

	\hspace*{-4mm}	$\delta_{i,j} = \text{\scshape ActTrs}(\delta^{in}_{i,j}, S^{in}(\bs{x}^i_j), 2^{-F_h}S^{in}(\hat{\bs{x}}^i_j))$\;
		}
	}

    \textcolor{gray}{// propagate in the output layer}

	\For{j \emph{in} $1,\ldots, n_d$}{
	\hspace*{-4mm}	$\Delta\bs{b}^d_j= 2^{-F_b}\hat{\bs{b}}^d_j-\bs{b}^d_j$\;

	\hspace*{-4mm}	  \mbox{$\delta_{d,j}= \delta_{d,j}^{in}=\text{\scshape AffTrs}(\bs{W}^d_{j,:},2^{-F_w}\widehat{\bs{W}}^d_{j,:}, \Delta\bs{b}^d_j, S(\bs{x}^{d-1}),\delta_{d-1},0)$\;}
	}
    \Return{$(\delta_{i,j})_{2\leq i\leq d, 1\leq j\leq n_d}$}
	\caption{Forward Difference Propagation}
	\label{alg:overall}
\end{algorithm}
 
\begin{algorithm}[t]
	\scriptsize
	\SetKwInOut{Input}{Input}
	\SetKwInOut{Output}{output}
	
	\Input{Weight vector $\bs{W}^i_{j,:}$, weight vector $\widetilde{\bs{W}}^i_{j,:}$, bias difference $\Delta \bs{b}^i_j$, neuron interval $S(\bs{x}^{i-1})$, difference interval $\delta_{i-1}$, rounding error $\xi$}
	\Output{Difference interval $\delta^{in}_{i,j}$}
	$lb=\text{LB}\big(\widetilde{\bs{W}}^i_{j,:}\delta_{i-1}+(\widetilde{\bs{W}}^i_{j,:}-\bs{W}^i_{j,:})S(\bs{x}^{i-1})\big)+\Delta\bs{b}^i_j-\xi$\;
	$ub=\text{UB}\big(\widetilde{\bs{W}}^i_{j,:}\delta_{i-1}+(\widetilde{\bs{W}}^i_{j,:}-\bs{W}^i_{j,:})S(\bs{x}^{i-1})\big)+\Delta\bs{b}^i_j+\xi$\;
	\Return{$[lb,ub]$\;}
	\caption{{\scshape AffTrs} Function}
	\label{alg:AffTrs}
\end{algorithm} 

Based on the above notations, we give our difference propagation in Alg.~\ref{alg:overall}. It works as follows.
Given a DNN $\mathcal{N}$, a QNN $\widehat{\mathcal{N}}$ and a quantized input region $R(\hat{\bs{x}},r)$, we first compute
intervals $S^{in}(\bs{x}^i_j)$ and $S(\bs{x}^i_j)$ for neurons in $\mathcal{N}$ using \emph{symbolic} 
interval analysis {\scshape DeepPoly}, and compute interval  $S^{in}(\hat{\bs{x}}^i_j)$ and $S(\hat{\bs{x}}^i_j)$ for neurons in $\widehat{\mathcal{N}}$ using concrete interval analysis method~\cite{scaleQNN21}.
 Remark that no symbolic interval analysis for QNNs exists.
By Definition~\ref{def:reb}, for each quantized input $\hat{\bs{x}}'$ for QNN, we obtain the input for DNN as $\bs{x}'=\hat{\bs{x}}'/(\mathcal{C}_{in}^\text{ub}-\mathcal{C}_{in}^\text{lb})$.
 After precision alignment,  we get the input difference as $2^{-F_{in}}\hat{\bs{x}}'-\bs{x}'=(2^{-F_{in}}-1/(\mathcal{C}_{in}^\text{ub}-\mathcal{C}_{in}^\text{lb}))\hat{\bs{x}}'$. 
 Hence, given an input region, we get the output difference of the input layer: $\delta_1=(2^{-F_{in}}-1/(\mathcal{C}_{in}^\text{ub}-\mathcal{C}_{in}^\text{lb}))S(\hat{\bs{x}}^1)$. Then, we compute the output difference  $\delta_i$ of each hidden layer iteratively by applying the affine transformer and activation transformer given in Alg.~\ref{alg:AffTrs} and Alg.~\ref{alg:ActTrs}. Finally, we get the output difference for the output layer using only
the affine transformer.

\smallskip
\noindent{\bf Affine Transformer.}
The difference before applying the activation function for the $j$-th neuron in the $i$-th layer is:
$\delta^{in}_{i,j}=2^{-F_h}\lfloor 2^{F_i}\widehat{\bs{W}}^i_{j,:}S(\hat{\bs{x}}^{i-1}) + 2^{F_h-F_b} \hat{\bs{b}}_j^i \rceil - \bs{W}^i_{j,:} S(\bs{x}^{i-1})-\bs{b}^i_j$
where $2^{-F_h}$ is used to align the precision between the outputs of the two networks (cf. Section~\ref{sec:pre}).
Then, we soundly remove the rounding operators and give constraints for upper bounds as well as lower bounds of $\delta^{in}_{i,j}$ as follows:
\[
\begin{array}{l}
\text{UB}(\delta^{in}_{i,j}) \le \text{UB}(2^{-F_h} (2^{F_i}\widehat{\bs{W}}^i_{j,:}S(\hat{\bs{x}}^{i-1}) + 2^{F_h-F_b} \hat{\bs{b}}_j^i +0.5)- \bs{W}^i_{j,:} S(\bs{x}^{i-1})-\bs{b}^i) \\
\text{LB}(\delta^{in}_{i,j}) \ge \text{LB}(2^{-F_h} (2^{F_i}\widehat{\bs{W}}^i_{j,:}S(\hat{\bs{x}}^{i-1}) + 2^{F_h-F_b} \hat{\bs{b}}_j^i -0.5)- \bs{W}^i_{j,:} S(\bs{x}^{i-1})-\bs{b}^i)
\end{array}
\]
Finally, we have $\text{UB}(\delta^{in}_{i,j}) \le \text{UB}\big(\widetilde{\bs{W}}^i_{j,:} S(\tilde{\bs{x}}^{i-1})- \bs{W}^i_{j,:} S(\bs{x}^{i-1})\big)+\Delta\bs{b}^i_j+\xi$ and $\text{LB}(\delta^{in}_{i,j}) \ge \text{LB}\big(\widetilde{\bs{W}}^i_{j,:} S(\tilde{\bs{x}}^{i-1})- \bs{W}^i_{j,:} S(\bs{x}^{i-1})\big)+\Delta\bs{b}^i_j-\xi$, which can be further reformulated as follows:
\[
\begin{array}{l}
\text{UB}(\delta^{in}_{i,j}) \le \text{UB}\big(\widetilde{\bs{W}}^i_{j,:} \delta_{i-1} +\Delta\bs{W}^i_{j,:} S(\bs{x}^{i-1})\big)+\Delta\bs{b}^i_j+\xi \\
\text{LB}(\delta^{in}_{i,j}) \ge \text{LB}\big(\widetilde{\bs{W}}^i_{j,:} \delta_{i-1} +\Delta\bs{W}^i_{j,:} S(\bs{x}^{i-1})\big)+\Delta\bs{b}^i_j-\xi
\end{array}
\]
where $S(\tilde{\bs{x}}^{i-1})=2^{-F_{in}}S(\hat{\bs{x}}^{i-1})$ if $i=2$, and $2^{-F_{h}}S(\hat{\bs{x}}^{i-1})$ otherwise. $\widetilde{\bs{W}}^i_{j,:}=2^{-F_w}\widehat{\bs{W}}^i_{j,:}$, $\Delta\bs{W}^i_{j,:}=\widetilde{\bs{W}}^i_{j,:}-\bs{W}^i_{j,:}$, $\Delta\bs{b}^i_j = 2^{-F_b}\hat{\bs{b}}^i_j- \bs{b}^i_j$ and $\xi=2^{-F_h-1}$.

\smallskip
\noindent
{\bf Activation Transformer.}
Now we give our activation transformer in Alg.~\ref{alg:ActTrs} which computes
 the difference interval  $\delta_{i,j}$ from the difference interval  $\delta_{i,j}^{in}$. Note that, the neuron interval $S(\hat{\bs{x}}^i_j)$ for the QNN has already been converted to the fixed-point counterpart $S(\tilde{\bs{x}}^i_j)=2^{-F_h}S(\hat{\bs{x}}^i_j)$ as an input parameter, as well as the clamping upper bound ($t=2^{-F_h}\up$). Different from {\scshape ReluDiff}  \cite{paulsen2020reludiff} which focuses on the subtraction of two ReLU functions, here we investigate the subtraction of the clamping function and ReLU function.

\begin{algorithm}[t]
	\scriptsize\SetInd{1em}{1em}
	\SetKwData{lbDnn}{LB$(S^{in}(\bs{x}^i_j))$}
	\SetKwData{ubDnn}{UB$(S^{in}(\bs{x}^i_j))$}

	\SetKwData{lbQnn}{LB$(S^{in}(\tilde{\bs{x}}_j^i))$}
	\SetKwData{ubQnn}{UB$(S^{in}(\tilde{\bs{x}}^i_j))$}
	
	\SetKwData{lbDiff}{LB$(\delta^{in}_{i,j})$}
	\SetKwData{ubDiff}{UB$(\delta^{in}_{i,j})$}
	
	\SetKwData{lbOut}{LB$(\delta_{i,j})$}
	\SetKwData{ubOut}{UB$(\delta_{i,j})$}
	
	\SetKwData{diffOut}{$\delta_{i,j}$}
	\SetKwData{diffIn}{$\delta^{in}_{i,j}$}
	\SetKwData{dnnIn}{$S^{in}(\bs{x}^i_j)$}
	\SetKwData{qnnIn}{$S^{in}(\tilde{\bs{x}}^i_j)$}

	\SetKwData{qnnUP}{$t$}
	\SetKwData{tup}{$[\mathcal{C}^{\text{ub}},\mathcal{C}^{\text{ub}}]$}
	\SetKwInOut{Input}{Input}
	\SetKwInOut{Output}{output}
	
	\Input{Difference interval $\delta^{in}_{i,j}$, neuron interval $S^{in}(\bs{x}^i_j)$, neuron interval $S^{in}(\tilde{\bs{x}}^i_j)$, clamp upper bound \qnnUP}
	\Output{Difference interval $\delta_{i,j}$}
	\BlankLine
	
	\lIf{\ubDnn$\le0$}{ $lb$ = clamp(\lbQnn,0, \qnnUP); \	$ub$ = clamp(\ubQnn,0, \qnnUP)
	}
	\ElseIf{\lbDnn$\ge0$}{ 
\hspace*{-4mm}		\lIf{\ubQnn$\le$ \qnnUP \textbf{\emph{and}} \lbQnn$\ge0$}{
	$lb$ = \lbDiff; \ \ $ub$ = \ubDiff
		}
\hspace*{-4mm}	    \ElseIf{\lbQnn$\ge$ \qnnUP \textbf{\emph{or}} \ubQnn$\le0$}{
\hspace*{-6mm}			$lb$ = clamp(\lbQnn, 0, \qnnUP)$-$\ubDnn\;
\hspace*{-6mm}	        $ub$ = clamp(\ubQnn, 0, \qnnUP)$-$\lbDnn \;
		}
\hspace*{-4mm}	   \ElseIf{\ubQnn$\le$ \qnnUP}{
\hspace*{-6mm}			$lb$ = max($-$\ubDnn, \lbDiff); \ $ub$ = max($-$\lbDnn, \ubDiff)\;
		}
\hspace*{-4mm}	   \ElseIf{\lbQnn$\ge$ 0}{
\hspace*{-6mm}			$lb$ = min($\qnnUP-\ubDnn$, \lbDiff); \ $ub$ = min($\qnnUP-\lbDnn$, \ubDiff)\;
		}
\hspace*{-4mm}	    \Else{
\hspace*{-6mm}			$lb$ = max($-$\ubDnn, min($\qnnUP-\ubDnn$, \lbDiff))\;
			
\hspace*{-6mm}			$ub$ = max($-$\lbDnn,  min(\qnnUP$-$\lbDnn, \ubDiff))\;
		}
	}
   \Else{ 
\hspace*{-4mm}	\If{\ubQnn$\le$ \qnnUP \textbf{\emph{and}} \lbQnn$\ge0$}{
\hspace*{-6mm}		$lb$ = min(\lbQnn, \lbDiff); \ $ub$ = min(\ubQnn, \ubDiff)\;
		}
\hspace*{-4mm}		\ElseIf{\lbQnn$\ge$ \qnnUP \textbf{\emph{or}} \ubQnn$\le0$}{
\hspace*{-6mm}			$lb$ = clamp(\lbQnn, 0, \qnnUP)$-$\ubDnn; \ 
$ub$ = clamp(\ubQnn, 0, \qnnUP)\;
		}
\hspace*{-4mm}		\ElseIf{\ubQnn$\le$ \qnnUP}{
\hspace*{-6mm}			$lb$ = max(\lbDiff, $-$\ubDnn); \ $ub$ = min(\ubDiff, \ubQnn)\;
\hspace*{-6mm}			\lIf{\ubDiff$\le 0$}{$ub$ = $0$}
\hspace*{-6mm}			\lIf{\lbDiff$\ge 0$}{	$lb$ = $0$}
		}
\hspace*{-4mm}		\ElseIf{\lbQnn$\ge$ 0}{
\hspace*{-6mm}			$lb$ = min(\lbDiff, \lbQnn, \qnnUP$-$\ubDnn); \ $ub$ = min(\ubDiff, \qnnUP)\;

		}
\hspace*{-4mm}		\Else{
\hspace*{-6mm}			$lb$ = min(\qnnUP$-$\ubDnn,0,max(\lbDiff,$-$\ubDnn)); \ $ub$ = clamp(\ubDiff, 0, \qnnUP)\;
		}
	}
	\Return{$[lb,ub]\cap\big( (S^{in}(\tilde{\bs{x}}^i_j) \cap [0,t]) - (S^{in}(\bs{x}^i_j)\cap [0, +\infty))\big)$}\;
	\caption{\text{\scshape ActTrs} function}
	\label{alg:ActTrs}
\end{algorithm}

\begin{theorem}\label{them:framework}
	If $\tau_h=+$, then Alg.~\ref{alg:overall} is sound.
\end{theorem}
Recall that we consider the ReLU activation function in this work,
thus $\tau_h$ should be $+$, i.e., unsigned. Proof refers to Appendix~\ref{sec:proofthem:framework}.

\begin{example}
We exemplify Alg.~\ref{alg:overall} using the networks $\mb_{e}$ and $\widehat{\mb}_{e}$ shown in Fig.\ref{fig:nnVS}. Given quantized input region $R((9,6),3)$ and the corresponding real-valued input region $R((0.6,0.4),0.2)$, we have $S(\hat{\bs{x}}^1_1)=[6,12]$ and $S(\hat{\bs{x}}^1_2)=[3,9]$. 

First, we get $S^{in}(\bs{x}^2_1)=S(\bs{x}^2_1)=[0.36,0.92]$, $S^{in}(\bs{x}^2_2)=[-0.4,0.2]$, $S(\bs{x}^2_2)=[0,0.2]$ based on {\scshape DeepPoly} (cf.~Appendix \ref{app_sec:deepPoly}) and get $S^{in}(\hat{\bs{x}}^2_1)=S(\hat{\bs{x}}^2_1)=[1,4]$, $S^{in}(\hat{\bs{x}}^2_2)=[-2,1]$, $S(\hat{\bs{x}}^2_2)=[0,1]$ via following concrete
interval analysis:
$\text{LB}(S^{in}(\hat{\bs{x}}^2_1))=\lfloor (5\text{LB}(\hat{\bs{x}}^1_1)-\text{UB}(\hat{\bs{x}}^1_2))/2^{-4}\rceil=1$,  $\text{UB}(S^{in}(\hat{\bs{x}}^2_1))=\lfloor (5\text{UB}(\hat{\bs{x}}^1_1)-\text{LB}(\hat{\bs{x}}^1_2))/2^{-4}\rceil=4$, $\text{LB}(S^{in}(\hat{\bs{x}}^2_2))=\lfloor (-3\text{UB}(\hat{\bs{x}}^1_1)+3\text{LB}(\hat{\bs{x}}^1_2))/2^{-4}\rceil=-2$, and $\text{UB}(S^{in}(\hat{\bs{x}}^2_2))=\lfloor (-3\text{LB}(\hat{\bs{x}}^1_1)+3\text{UB}(\hat{\bs{x}}^1_2))/2^{-4}\rceil=1$.
By Line 3 in Alg.~\ref{alg:overall}, we have $\delta_{1,1}=-\frac{1}{16\times 15}S(\hat{\bs{x}}^1_1)=[-0.05,-0.025]$, $\delta_{1,2}=-\frac{1}{16\times 15}S(\hat{\bs{x}}^1_2)=[-0.0375,-0.0125]$.

Then, we compute the difference interval before the activation functions. The rounding error is $\xi=2^{-F_h-1}=0.125$.
We obtain the difference intervals $\delta^{in}_{2,1}=[-0.194375, 0.133125]$ and  $\delta^{in}_{2,2}=[-0.204375, 0.123125]$ as follows based on Alg.~\ref{alg:AffTrs}:
\begin{itemize}
	\item $\text{LB}(\delta^{in}_{2,1})=\text{LB}(\widetilde{\bs{W}}^1_{1,1} \delta_{1,1} + \widetilde{\bs{W}}^1_{1,2} \delta_{1,2} +  \Delta \bs{W}^1_{1,1} S(\bs{x}^1_1)+\Delta \bs{W}^1_{1,2} S(\bs{x}^1_2))-\xi=1.25\times \text{LB}(\delta_{1,1}) -0.25 \times \text{UB}(\delta_{1,2}) + (1.25-1.2)\times \text{LB}(S(\bs{x}^1_1)) +(-0.25+0.2)\times \text{UB}(S(\bs{x}^1_2))-0.125$,
	$\text{UB}(\delta^{in}_{2,1})=\text{UB}(\widetilde{\bs{W}}^1_{1,1} \delta_{1,1} + \widetilde{\bs{W}}^1_{1,2} \delta_{1,2} +  \Delta \bs{W}^1_{1,1} S(\bs{x}^1_1)+\Delta \bs{W}^1_{1,2} S(\bs{x}^1_2))+\xi=1.25\times \text{UB}(\delta_{1,1}) -0.25 \times \text{LB}(\delta_{1,2}) +(1.25-1.2)\times \text{UB}(S(\bs{x}^1_1)) +(-0.25+0.2)\times \text{LB}(S(\bs{x}^1_2))+0.125$;
	\item $\text{LB}(\delta^{in}_{2,2})=\text{LB}(\widetilde{\bs{W}}^1_{2,1} \delta_{1,1} + \widetilde{\bs{W}}^1_{2,2} \delta_{1,2} +  \Delta \bs{W}^1_{2,1} S(\bs{x}^1_1)+\Delta \bs{W}^1_{2,2} S(\bs{x}^1_2))-\xi=-0.75\times \text{UB}(\delta_{1,1})+ 0.75\times \text{LB}(\delta_{1,2})+(-0.75+0.7)\times \text{UB}(S(\bs{x}^1_1))  +(0.75-0.8)\times \text{UB}(S(\bs{x}^1_2))-0.125$,
	$\text{UB}(\delta^{in}_{2,2})=\text{UB}(\widetilde{\bs{W}}^1_{2,1} \delta_{1,1} + \widetilde{\bs{W}}^1_{2,2} \delta_{1,2} +  \Delta \bs{W}^1_{2,1} S(\bs{x}^1_1)+\Delta \bs{W}^1_{2,2} S(\bs{x}^1_2))+ \xi=-0.75\times \text{LB}(\delta_{1,1})+0.75\times \text{UB}(\delta_{1,2}) +(-0.75+0.7)\times \text{LB}(S(\bs{x}^1_1)) + (0.75-0.8)\times \text{LB}(S(\bs{x}^1_2))+0.125$.
\end{itemize}
By Lines 20$\sim$22 in Alg.~\ref{alg:ActTrs}, we get the difference intervals after the activation functions for the hidden layer as: $\delta_{2,1}=\delta^{in}_{2,1}=[-0.194375, 0.133125]$, $\delta_{2,1}=[\text{max}\big(\text{LB}(\delta^{in}_{2,2}),-\text{UB}(S^{in}(\bs{x}^2_2))\big), \text{min}\big(\text{UB}(\delta^{in}_{2,2}),\text{UB}(S^{in}(\tilde{\bs{x}}^2_2))\big)]=[-0.2,0.123125]$.

Next, we compute the output difference interval of the networks using Alg.~\ref{alg:AffTrs} again but with $\xi=0$:
	$\text{LB}(\delta^{in}_{3,1})=\text{LB}(\widetilde{\bs{W}}^2_{1,1} \delta_{2,1} + \widetilde{\bs{W}}^2_{1,2} \delta_{2,2} +  \Delta \bs{W}^2_{1,1} S(\bs{x}^2_1)+\Delta \bs{W}^2_{1,2} S(\bs{x}^2_2))=0.25\times \text{LB}(\delta_{2,1})+0.75\times \text{LB}(\delta_{2,2})+(0.25-0.3)\times \text{UB}(S(\bs{x}^2_1))+ (0.75-0.7)\times \text{LB}(S(\bs{x}^2_2))$,
	 $\text{UB}(\delta^{in}_{3,2})=\text{UB}(\widetilde{\bs{W}}^2_{1,1} \delta_{2,1} + \widetilde{\bs{W}}^2_{1,2} \delta_{2,2} +  \Delta \bs{W}^2_{1,1} S(\bs{x}^2_1)+\Delta \bs{W}^2_{1,2} S(\bs{x}^2_2))=0.25\times \text{UB}(\delta_{2,1})+ 0.75\times \text{UB}(\delta_{2,2})+(0.25-0.3)\times \text{LB}(S(\bs{x}^2_1))+(0.75-0.7)\times \text{UB}(S(\bs{x}^2_2))$.
%
Finally, the quantization error interval is [-0.24459375, 0.117625].
\end{example}


\subsection{MILP Encoding of the Verification Problem}
If DRA fails to prove the property,
we encode the problem as an equivalent MILP problem.
Specifically, we encode both the QNN and DNN as sets of (mixed integer) linear constraints,
and quantize the input region as a set of integer linear constraints.
We adopt the MILP encodings of DNNs~\cite{LomuscioM17} and QNNs~\cite{mistry2022milp}
to transform the DNN and QNN into a set of linear constraints. We use (symbolic) intervals to further reduce the size of linear constraints similar to~\cite{LomuscioM17} while \cite{mistry2022milp} did not.
We suppose that the sets of constraints encoding the QNN, DNN, and quantized input region are $\Theta_{\widehat{\mathcal{N}}}$, $\Theta_{\mathcal{N}}$, and $\Theta_{R}$, respectively. Next, we give the MILP encoding of the robust error bound property.

Recall that, given a DNN $\mathcal{N}$, an input region $R(\hat{\bs{x}},r)$
such that $\bs{x}$ is classified to class $g$ by $\mathcal{N}$, a QNN $\widehat{\mathcal{N}}$ has a quantization error bound $\epsilon$ w.r.t. $R(\hat{\bs{x}},r)$ if for every $\hat{\bs{x}}'\in R(\hat{\bs{x}},r)$, we have $|2^{-F_h}\widehat{\mathcal{N}}(\hat{\bs{x}}')_g- \mathcal{N}(\bs{x}')_g|<\epsilon$.
%
Thus, it suffices to check if $|2^{-F_h}\widehat{\mathcal{N}}(\hat{\bs{x}}')_g- \mathcal{N}(\bs{x}')_g|\ge\epsilon$ for some $\hat{\bs{x}}'\in  R(\hat{\bs{x}},r)$.

Let $\hat{\bs{x}}^d_g$ (resp. $\bs{x}^d_g$) be the $g$-th output of $\widehat{\mathcal{N}}$ (resp. $\mathcal{N}$).
We introduce a real-valued variable $\eta $ and a Boolean variable $v$ such that $\eta = \text{max}(2^{-F_h}\hat{\bs{x}}^d_g - \bs{x}^d_g,0)$ can be encoded by the set
$\Theta_g$ of constraints with an extremely large number \textbf{M}: $\Theta_g = \big\{\eta \ge 0, \ \eta \ge 2^{-F_h}\hat{\bs{x}}^d_g - \bs{x}^d_g, \ \eta \le \textbf{M} \cdot v, \ \le 2^{-F_h}\hat{\bs{x}}^d_g - \bs{x}^d_g + \textbf{M}\cdot (1-v)\big\}$.
%
%
As a result, $|2^{-F_h}\hat{\bs{x}}^d_g - \bs{x}^d_g|\ge \epsilon$ iff the set of linear constraints $\Theta_{\epsilon} = \Theta_g \cup \{ 2\eta- (2^{-F_h}\hat{\bs{x}}^d_g - \bs{x}^d_g)\ge \epsilon\}$ holds.

Finally, the quantization error bound verification problem is equivalent to the solving of the constraints: $\Theta_{P} = \Theta_{\widehat{\mathcal{N}}} \cup \Theta_{\mathcal{N}} \cup \Theta_{R} \cup \Theta_{\epsilon}$.
Remark that the output difference intervals of hidden neurons obtained from Alg.~\ref{alg:overall} can be encoded as linear constraints
which are added into the set $\Theta_{P}$ to boost the solving.

\section{An Abstract Domain for Symbolic-based DRA}\label{sec:abstract}
While Alg.~\ref{alg:overall} can compute difference intervals,
the affine transformer explicitly adds a concrete rounding error interval to each neuron, which accumulates to significant precision loss over the subsequent layers.
To alleviate this problem,
we introduce an abstract domain based on {\scshape DeepPoly} which helps to compute sound symbolic approximations for the lower and upper bounds of each difference interval, hence computing tighter difference intervals.

\subsection{An Abstract Domain for QNNs}\label{sec:abstractQNN}
We first introduce transformers for affine transforms with rounding operators and clamp functions in QNNs.
Recall that the activation function in a QNN $\widehat{\mathcal{N}}$ is also a min-ReLU function: $\text{min}(\text{ReLU}(\lfloor\cdot\rceil),\up)$.
Thus, we regard each hidden neuron $\hat{\bs{x}}^i_j$  in a QNN as three nodes $\hat{\bs{x}}^i_{j,0}$, $\hat{\bs{x}}^i_{j,1}$, and $\hat{\bs{x}}^i_{j,2}$
 such that $\hat{\bs{x}}^i_{j,0}=\lfloor 2^{F_i}\sum_{k=1}^{n_{i-1}}\widehat{\bs{W}}^i_{j,k}{\hat{\bs{x}}}^{i-1}_{k,2} + 2^{F_h-F_b} \hat{\bs{b}}_j^i  \rceil$ (affine function),
 $\hat{\bs{x}}^i_{j,1} = \text{max}(\hat{\bs{x}}^i_{j,0},0)$ (ReLU function) and
 $\hat{\bs{x}}^i_{j,2}=\text{min}(\hat{\bs{x}}^i_{j,1},\up)$ (min function).
We now give the abstract domain $\widehat{\mathcal{A}}^i_{j,p}=\langle \hat{\bs{a}}^{i,\le}_{j,p}, \hat{\bs{a}}^{i,\ge}_{j,p}, \hat{l}^i_{j,p}, \hat{u}^i_{j,p}\rangle$ for each neuron $\hat{\bs{x}}^i_{j,p}$ ($p\in\{0,1,2\}$) in a QNN as follows.

Following {\scshape DeepPoly}, $\hat{\bs{a}}^{i,\le}_{j,0}$ and $\hat{\bs{a}}^{i,\ge}_{j,0}$
for the affine function of $\hat{\bs{x}}^i_{j,0}$ with rounding operators are defined as
$\hat{\bs{a}}^{i,\le}_{j,0}= 2^{F_i}\sum_{k=1}^{n_{i-1}}\widehat{\bs{W}}^i_{j,k}{\hat{\bs{x}}}^{i-1}_{k,2} + 2^{F_h-F_b} \hat{\bs{b}}_j^i -0.5$ and
$\hat{\bs{a}}^{i,\ge}_{j,0}= 2^{F_i}\sum_{k=1}^{n_{i-1}}\widehat{\bs{W}}^i_{j,k}{\hat{\bs{x}}}^{i-1}_{k,2} + 2^{F_h-F_b} \hat{\bs{b}}_j^i +0.5$.
%
We remark that $+0.5$ and $-0.5$ here are added to soundly encode the rounding operators and have no effect on the preservance of invariant since the rounding operators will add/subtract 0.5 at most to round each floating-point number into its nearest integer.
The abstract transformer for the ReLU function $\bs{x}^i_{j,1} = \text{ReLU}(\bs{x}^i_{j,0})$ is defined the same as {\scshape DeepPoly}.

\begin{figure}[t]
	\centering
		\subfigure[$(\gamma,\mu)=(0,\up)$]{
			\label{fig:deepPolyQ1}
			\includegraphics[width=0.25\textwidth]{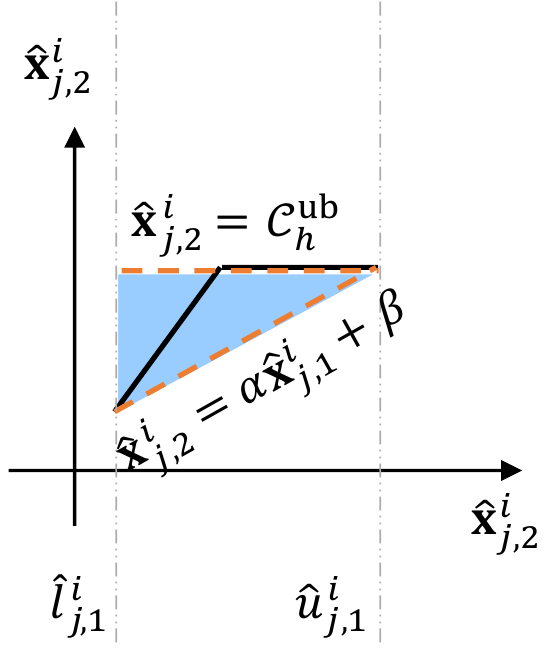}
		}
\qquad
		\subfigure[$(\gamma,\mu)=(1,0)$]{
			\label{fig:deepPolyQ2}
			\includegraphics[width=0.25\textwidth]{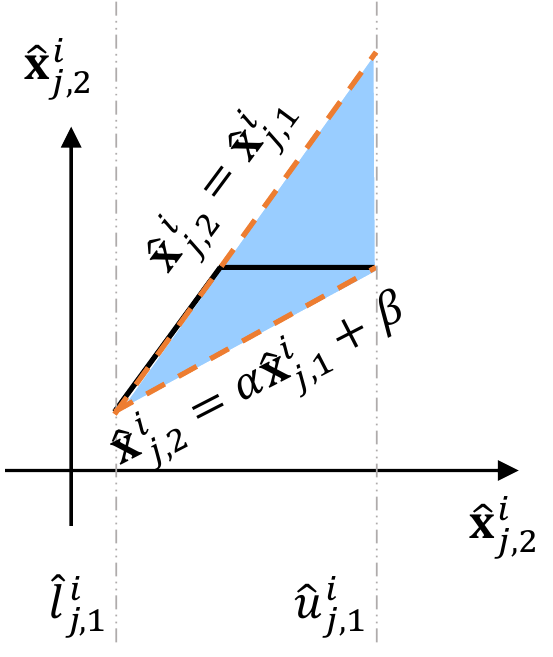}
		}
		\caption{Convex approximation for the min function in QNNs, where Figure~\ref{fig:deepPolyQ1} and Figure~\ref{fig:deepPolyQ2} show the two ways where $\alpha=\frac{\up-\hat{l}^i_{j,1}}{\hat{u}^i_{j,1}-\hat{l}^i_{j,1}}$ and $\beta=\frac{(\hat{u}^i_{j,1}-\up)}{\hat{u}^i_{j,1}-\hat{l}^i_{j,1}}$.}
		\label{fig:deepPolyQ}
\end{figure}

For the min function $\hat{\bs{x}}^i_{j,2}=\text{min}(\hat{\bs{x}}^i_{j,1},\up)$, there are three cases for $\widehat{\mathcal{A}}^i_{j,2}$:
\begin{itemize}
	\item If $\hat{l}^i_{j,1}\ge \up$, then $\hat{\bs{a}}_{j,2}^{i,\le}=\hat{\bs{a}}_{j,2}^{i,\ge}=\up$, $\hat{l}^i_{j,2}=\hat{u}^i_{j,2}=\up$;
	\item If $\hat{u}^i_{j,1}\le \up$, then $\hat{\bs{a}}_{j,2}^{i,\le}= \hat{\bs{a}}_{j,1}^{i,\le}$, $\hat{\bs{a}}_{j,2}^{i,\ge}=\hat{\bs{a}}_{j,1}^{i,\ge}$, $\hat{l}^i_{j,2}=\hat{l}^i_{j,1}$ and $\hat{u}^i_{j,2}=\hat{u}^i_{j,1}$;
	\item If $\hat{l}^i_{j,1}<\up\wedge \hat{u}^i_{j,1}>\up$, then $\hat{\bs{a}}_{j,2}^{i,\ge}=\lambda\hat{\bs{x}}^i_{j,1}+\mu $ and $\hat{\bs{a}}_{j,2}^{i,\le}=\frac{\up-\hat{l}^i_{j,1}}{\hat{u}^i_{j,1}-\hat{l}^i_{j,1}}\hat{\bs{x}}^i_{j,1}+\frac{(\hat{u}^i_{j,1}-\up)}{\hat{u}^i_{j,1}-\hat{l}^i_{j,1}} \hat{l}^i_{j,1}$, where $(\lambda,\mu)\in\{(0,\up),(1,0)\}$ such that the area of resulting shape by $\hat{\bs{a}}_{j,2}^{i,\le}$ and $\hat{\bs{a}}_{j,2}^{i,\ge}$ is minimal, $\hat{l}^i_{j,2}=\hat{l}^i_{j,1}$ and $\hat{u}^i_{j,2}= \lambda\hat{u}^i_{j,1}+\mu$. We show the two ways of approximation in Figure~\ref{fig:deepPolyQ}.
\end{itemize}

\begin{theorem}\label{them:abQNN}
	The min abstract transformer preserves the following invariant: $\Gamma(\widehat{\mathcal{A}}^i_{j,2})\subseteq [\hat{l}^i_{j,2}, \hat{u}^i_{j,2}]$.
\end{theorem}


From our abstract domain for QNNs, we get
a symbolic interval analysis, similar to the one for DNNs using {\scshape DeepPoly},
to replace Line 2 in Alg.~\ref{alg:overall}.
Proof refers to Appendix~\ref{sec:proofthem:abQNN}.

\subsection{Symbolic Quantization Error Computation}\label{sec:diffSym}

Recall that to compute tight bounds of QNNs or DNNs via symbolic interval analysis,
variables in upper and lower polyhedral computations are recursively
substituted with the corresponding upper/lower polyhedral
computations of variables until they only contain the input
variables from which the concrete intervals are computed.
This idea motivates us to design a symbolic difference computation approach
for differential reachability analysis based on the abstract domain {\scshape DeepPoly} for DNNs
and our abstract domain for QNNs.

Consider two hidden neurons $\bs{x}^i_{j,s}$ and $\hat{\bs{x}}^i_{j,s}$ from the DNN $\mathcal{N}$ and the QNN $\widehat{\mathcal{N}}$.
Let $\mathcal{A}^{i,\ast}_{j,s}=\langle \bs{a}^{i,\le,\ast}_{j,s},\bs{a}^{i,\ge,\ast}_{j,s}, l^{i,\ast}_{j,s}, u^{i,\ast}_{j,s}\rangle$ and $\widehat{\mathcal{A}}^i_{j,p}=\langle \hat{\bs{a}}^{i,\le,\ast}_{j,p},\hat{\bs{a}}^{i,\ge,\ast}_{j,p}, \hat{l}^{i,\ast}_{j,p}, \hat{u}^{i,\ast}_{j,p}\rangle$ be their abstract elements, respectively, where all the polyhedral computations are linear combinations of the input variables of the DNN and QNN, respectively, i.e., 
\begin{itemize}
\item $\bs{a}^{i,\le,\ast}_{j,s}=\sum_{k=1}^m \bs{w}^{l,\ast}_k \bs{x}^1_k+\bs{b}^{l,\ast}_j$,  $\bs{a}^{i,\ge,\ast}_{j,s}=\sum_{k=1}^m \bs{w}^{u,\ast}_k \bs{x}^1_k+\bs{b}^{u,\ast}_j$
\item $\hat{\bs{a}}^{i,\le,\ast}_{j,p}=\sum_{k=1}^m \hat{\bs{w}}^{l,\ast}_k \hat{\bs{x}}^1_k+\hat{\bs{b}}^{l,\ast}_j$,  $\hat{\bs{a}}^{i,\ge,\ast}_{j,p}=\sum_{k=1}^m \hat{\bs{w}}^{u,\ast}_k \hat{\bs{x}}^1_k+\hat{\bs{b}}^{u,\ast}_j$.
\end{itemize}
Then, the sound lower bound $\Delta l^{i,\ast}_{j,s}$ and upper $\Delta u^{i,\ast}_{j,s}$ bound of the difference can be derived as follows, where $p=2s$:
\begin{itemize}
	\item $\Delta l^{i,\ast}_{j,s} =\text{LB}(2^{-F_h}\hat{\bs{x}}^i_{j,p}-\bs{x}^i_{j,s}) = 2^{-F_h}\hat{\bs{a}}^{i,\le,\ast}_{j,p}-\bs{a}^{i,\ge,\ast}_{j,s}$
	\item $\Delta u^{i,\ast}_{j,s}=\text{UB}(2^{-F_h}\hat{\bs{x}}^i_{j,p}-\bs{x}^i_{j,s})=2^{-F_h}\hat{\bs{a}}^{i,\ge,\ast}_{j,p}-\bs{a}^{i,\le,\ast}_{j,s}$
\end{itemize}


Given a quantized input $\hat{\bs{x}}$ of the QNN $\widehat{\mathcal{N}}$, the input difference of two networks is $2^{-F_{in}}\hat{\bs{x}}-\bs{x} = (2^{-F_{in}}\up -1 )\bs{x}$.
Therefore, we have $\Delta^1_k=\tilde{\bs{x}}^1_k-\bs{x}^1_k= 2^{-F_{in}}\hat{\bs{x}}^1_k-\bs{x}^1_k=(2^{-F_{in}}\up -1 )\bs{x}$. Then, the lower bound of difference can be reformulated as follows which only contains the input variables of DNN $\mathcal{N}$: $\Delta l^{i,\ast}_{j,s} = \Delta\bs{b}^{l,\ast}_j+ \sum_{k=1}^m (- \bs{w}^{u,\ast}_k + 2^{-F_{in}}\up\tilde{\bs{w}}^{l,\ast}_k ) \bs{x}^1_k$,
where $\Delta\bs{b}^{l,\ast}_j=2^{-F_h}\hat{\bs{b}}^{l,\ast}_j-\bs{b}^{u,\ast}_j$, $F^\ast=F_{in}-F_h$, $\Delta^1_k=\tilde{\bs{x}}^1_k-\bs{x}^1_k$ and $\tilde{\bs{w}}^{l,\ast}_k=2^{F^\ast}\hat{\bs{w}}_k^{l,\ast}$.

Similarly, we can reformulated the upper bound $\Delta u^{i,\ast}_{j,s}$ as follows using the input variables
of the DNN: $
\Delta u^{i,\ast}_{j,s} =\Delta\bs{b}^{u,\ast}_j+ \sum_{k=1}^m (- \bs{w}^{l,\ast}_k + 2^{-F_{in}}\up\tilde{\bs{w}}^{u,\ast}_k ) \bs{x}^1_k
$,
where $\Delta\bs{b}^{u,\ast}_j=2^{-F_h}\hat{\bs{b}}^{u,\ast}_j-\bs{b}^{l,\ast}_j$, $F^\ast=F_{in}-F_h$, and $\tilde{\bs{w}}^{u,\ast}_k=2^{F^\ast}\hat{\bs{w}}_k^{u,\ast}$.

Finally, we compute the concrete input difference interval $\delta_{i,j}^{in}$ based on the given input region as $\delta_{i,j}^{in}=[\text{LB}(\Delta l^{i,\ast}_{j,0}), \text{UB}(\Delta u^{i,\ast}_{j,0})]$, with which we can replace the {\scshape AffTrs} functions in Alg.~\ref{alg:overall} directly. An illustrating example is given in Appendix~\ref{app_sec:sym_DRA}.




\section{Evaluation}\label{sec:exp}

We have implemented our method \tool as an end-to-end tool written in Python, where we use Gurobi~\cite{Gurobi} as our back-end MILP solver. All floating-point numbers used in our tool are 32-bit. Experiments are conducted on a 96-core machine with Intel(R) Xeon(R) Gold 6342 2.80GHz CPU and 1 TB main memory. We allow Gurobi to use up to 24 threads. The time limit for each verification task is 1 hour.




\smallskip
\noindent
{\bf Benchmarks.}
We first build 45*4 QNNs from the 45 DNNs of ACAS Xu~\cite{julian2019deep}, following a \emph{post-training quantization scheme}~\cite{nagel2021white} and using quantization configurations $\mathcal{C}_{in}=\langle \pm, 8,8\rangle$, $\mathcal{C}_w=\mathcal{C}_b=\langle \pm , Q,Q-2\rangle$, $\mathcal{C}_h=\langle + , Q,Q-2\rangle$, where $Q\in\{4,6,8,10\}$. We then train 5 DNNs with different architectures using the MNIST dataset~\cite{MNIST} and build 5*4 QNNs following the same quantization scheme and quantization configurations except that we set $\mathcal{C}_{in}=\langle +, 8,8\rangle$ and $\mathcal{C}_w=\langle \pm , Q,Q-1\rangle$ for each DNN trained on MNIST. Details on the networks trained on the MNIST dataset are presented in Table~\ref{tab:bench}. 
Column 1 gives the name and architecture of each DNN, where $A$blk\_$B$ means that the network has $A$ hidden layers with each hidden layer size $B$ neurons, 
Column 2 gives the number of parameters in each DNN, and Columns 3-7 list the accuracy of these networks. Hereafter, we denote by P$x$-$y$ (resp. A$x$-$y$) the QNN using the architecture P$x$ (using the $x$-th DNN) and quantization bit size $Q=y$ for MNIST (resp. ACAS Xu), and by P$x$-Full (resp. A$x$-Full) the DNN of architecture P$x$ for MNIST (resp. the $x$-th DNN in ACAS Xu).

\begin{table}[t]
	\centering
	\caption{Benchmarks for QNNs and DNNs on MNIST.}
	\label{tab:bench}\setlength{\tabcolsep}{5pt} 
	\scalebox{0.85}{
		\begin{tabular}{c|c|cccc|c}
			\toprule
			& & \multicolumn{4}{c|}{QNNs} & ~ \\ 
			\multirow{-2}*{Arch} & \multirow{-2}*{\#Paras} & $Q=4$ & $Q=6$  & $Q=8$ & $Q=10$ & \multirow{-2}*{DNNs} \\ \midrule
			\rowcolor{gray!20}
			P1: 1blk\_100 & $\approx$ 79.5k &96.38\% & 96.79\% & 96.77\% & 96.74\% & 96.92\% \\
			P2: 2blk\_100 & $\approx$ 89.6k &96.01\% & 97.04\%	& 97.00\% & 97.02\% & 97.07\% \\
			
			\rowcolor{gray!20}
			P3: 3blk\_100 & $\approx$ 99.7k &95.53\% & 96.66\% & 96.59\% &	96.68\% & 96.71\% \\
			P4: 2blk\_512 & $\approx$ 669.7k &96.69\% & 97.41\% & 97.35\% & 97.36\% & 97.36\% \\
			
			\rowcolor{gray!20}
			P5: 4blk\_1024 & $\approx$ 3,963k &97.71\% & 98.05\% & 98.01\% & 98.04\% & 97.97\% \\
			
			\bottomrule
		\end{tabular}
		
	}
\end{table}

\subsection{Effectiveness and Efficiency of DRA}\label{sec:RQ1}
We first implement a naive method using existing state-of-the-art reachability analysis methods for QNNs and DNNs.
Specifically, we use the symbolic interval analysis of {\scshape DeepPoly}~\cite{SGPV19} to compute the output intervals for a DNN, and use interval analysis of~\cite{scaleQNN21} to compute the output intervals for a QNN.
Then, we compute quantization error intervals via interval subtraction.
Note that no existing methods can directly verify quantization error bounds and the methods in~\cite{paulsen2020reludiff,PaulsenWWW20} are not applicable. Finally, we compare the quantization error intervals computed by the naive method against DRA in \tool, using DNNs A$x$-Full, P$y$-Full and QNNs A$x$-$z$, P$y$-$z$ for $x=1$, $y\in\{1,2,3,4,5\}$ and $z\in\{4,6,8,10\}$. We use the same adversarial input regions (5 input points with radius $r=\{3,6,13,19,26\}$ for each point) as in~\cite{KBDJK17} for ACAS Xu, and set the quantization error bound $\epsilon=\{0.05,0.1,0.2,0.3,0.4\}$, i.e., resulting 25 tasks for each radius.
For MNIST, we randomly select 30 input samples from the test set of MNIST and set radius $r=3$ for each input sample and quantization error bound $\epsilon=\{1,2,4,6,8\}$, resulting in a total of 150 tasks for each pair of DNN and QNN of same architecture for MNIST.

Table~\ref{tab:DRA} reports the analysis results for ACAS Xu (above) and MNIST (below). Column 2 lists different analysis methods, where
\tool(Int) is Alg.~\ref{alg:overall}
and \tool(Sym) uses a symbolic-based method for the affine transformation in Alg.~\ref{alg:overall} (cf. Section~\ref{sec:diffSym}).
Columns (H\_Diff) (resp. O\_Diff) averagely give the sum ranges of the difference intervals of all the hidden neurons (resp. output neurons of the predicted class) for the 25 verification tasks for ACAS Xu and 150 verification tasks for MNIST. Columns (\#S/T) list the number of tasks (\#S) successfully proved by DRA and average computation time (T) in seconds, respectively, where the best ones (i.e., solving the most tasks) are highlighted in blue. Note that Table~\ref{tab:DRA} only reports the number of true propositions proved by DRA while the exact number is unknown.

\begin{table}[t]
	\centering
	\caption{Differential Reachability Analysis on ACAS Xu and MNIST.}
	\label{tab:DRA}
	\setlength{\tabcolsep}{1pt} 
	\scalebox{0.59}{
		\begin{tabular}{c|c|ccc|ccc|ccc|ccc|ccc}
			\toprule
			~ &  ~ &\multicolumn{3}{c|}{$r=3$}&\multicolumn{3}{c|}{$r=6$}&\multicolumn{3}{c|}{$r=13$}&\multicolumn{3}{c|}{$r=19$}&\multicolumn{3}{c}{$r=26$}\\
			\multirow{-2}*{$Q$} &  \multirow{-2}*{Method} & H\_Diff & O\_Diff & \#S/T & H\_Diff & O\_Diff & \#S/T & H\_Diff & O\_Diff & \#S/T & H\_Diff & O\_Diff & \#S/T & H\_Diff & O\_Diff & \#S/T  \\ \midrule
			
			
			\cellcolor{white} &  Naive & 270.5 & 0.70 & \hl{15/0.47} &423.7 & 0.99 & \hl{9/0.52} & 1,182 & 4.49 & 0/0.67 & 6,110 & 50.91 & 0/0.79 & 18,255 & 186.6 & 0/0.81 \\
			\rowcolor{gray!20}
			\cellcolor{white}{4} & \tool (Int) & 270.5 & 0.70 & \hl{15/0.49} & 423.4 & 0.99 & \hl{9/0.53} & 1,181 & 4.46 & 0/0.70 & 6,044 & 50.91 & 0/0.81 & 17,696 & 186.6 & 0/0.85 \\
			
			\cellcolor{white}& \tool (Sym) & 749.4 & 145.7 & 0/2.02 & 780.9 & 150.2 & 0/2.11 & 1,347 & 210.4 & 0/2.24 & 6,176 & 254.7 & 0/2.35 & 18,283 & 343.7 & 0/2.39 \\
			\midrule
			
			\cellcolor{white} &  Naive & 268.3 & 1.43 & 5/0.47 & 557.2 & 4.00 & 0/0.51 & 1,258 & 6.91 & 0/0.67 & 6,145 & 53.29 & 0/0.77 & 18,299 & 189.0 & 0/0.82 \\
			\rowcolor{gray!20}
			\cellcolor{white}{6} & \tool (Int) & 268.0 & 1.41 & 5/0.50 & 555.0 & 3.98 & 0/0.54 & 1,245 & 6.90 & 0/0.69 & 6,125 & 53.28 & 0/0.80 & 18,218 & 189.0 & 0/0.83 \\
			\cellcolor{white} &  \tool (Sym)  & 299.7 & 2.58 & \hl{10/1.48} & 365.1 & 3.53 & \hl{9/1.59} & 1,032 & 7.65 & \hl{5/1.91} & 5,946 & 85.46 & \hl{4/2.15} & 18,144 & 260.5 & 0/2.27 \\
			\midrule
			
			\cellcolor{white} &  Naive & 397.2 & 3.57 & 0/0.47 & 587.7 & 5.00 & 0/0.51 & 1,266 & 7.90 & 0/0.67 & 6,160 & 54.27 & 0/0.78 & 18,308 & 190.0 & 0/0.81 \\
			\rowcolor{gray!20}
			\cellcolor{white}{8} &\tool (Int)  & 388.4 & 3.56 & 0/0.49 & 560.1 & 5.00 & 0/0.53 & 1,222 & 7.89 & 0/0.69 & 6,103 & 54.27 & 0/0.79 & 18,212 & 190.0 & 0/0.83 \\
			
			\cellcolor{white} & \tool (Sym)  & 35.75 & 0.01 & \hl{24/1.10} & 93.78 & 0.16 & \hl{18/1.19} & 845.2 & 5.84 & \hl{8/1.65} & 5,832 & 58.73 & \hl{5/1.97} & 18,033 & 209.6 & \hl{5/2.12}  \\
			\midrule
			
			\cellcolor{white} & Naive & 394.5 & 3.67 & 0/0.49 & 591.4 & 5.17 & 0/0.51 & 1,268 & 8.04 & 0/0.68 & 6,164 & 54.42 & 0/0.78 & 18,312 & 190.1 & 0/0.80 \\
			\rowcolor{gray!20}
			\cellcolor{white}{10} & \tool (Int) & 361.9 & 3.67 & 0/0.50 & 546.2 & 5.17 & 0/0.54 & 1,209 & 8.04 & 0/0.68 & 6,083 & 54.42 & 0/0.79 & 18,182 & 190.1 & 0/0.83 \\
			\cellcolor{white} & \tool (Sym) & 15.55 & 0.01 & \hl{25/1.04} & 54.29 & 0.06 & \hl{22/1.15} & 764.6 & 4.53 & \hl{9/1.52} & 5,780 & 57.21 & \hl{5/1.91} & 18,011 & 228.7 & \hl{5/2.08} \\
			\midrule \midrule
			
			~ &  ~ &\multicolumn{3}{c|}{P1 }&\multicolumn{3}{c|}{P2 }&\multicolumn{3}{c|}{P3}&\multicolumn{3}{c|}{P4}&\multicolumn{3}{c}{P5 }\\
			\multirow{-2}*{$Q$} &  \multirow{-2}*{Method} & H\_Diff & O\_Diff & \#S/T & H\_Diff & O\_Diff & \#S/T & H\_Diff & O\_Diff & \#S/T & H\_Diff & O\_Diff & \#S/T & H\_Diff & O\_Diff & \#S/T \\ \midrule
			
			\cellcolor{white} &  Naive & 64.45 & 7.02 & 61/0.77 & 220.9 & 20.27 & 0/1.53 & 551.6 & 47.75 & 0/2.38 & 470.1 & 22.69 & 2/11.16 & 5,336 & 140.4 & 0/123.0 \\
			
			\rowcolor{gray!20}
			\cellcolor{white}{4} & \tool (Int)  & 32.86 & 6.65 & 63/0.78 & 194.8 & 20.27 & 0/1.54 & 530.9 & 47.75 & 0/2.40 & 443.3 & 22.69 & 2/11.23 & 5,275 & 140.4 & 0/123.4 \\
			
			\cellcolor{white} & \tool (Sym)  & 32.69 & 3.14 & \hl{88/1.31} & 134.9 & 7.11 & \hl{49/2.91} & 313.8 & 14.90 & \hl{1/5.08} & 365.2 & 11.11 & \hl{35/22.28} & 1,864 & 50.30 & \hl{1/310.2} \\
			\midrule
			
			\cellcolor{white} &  Naive & 68.94 & 7.89 & 66/0.77 & 249.5 & 24.25 & 0/1.52 & 616.2 & 54.66 & 0/2.38 & 612.2 & 31.67 & 1/11.18 & 7,399 & 221.0 & 0/125.4 \\
			
			\rowcolor{gray!20}
			\cellcolor{white}{6} & \tool (Int)  & 10.33 & 2.19 & 115/0.78 & 89.66 & 12.81 & 14/1.54 & 466.0 & 52.84 & 0/2.39 & 307.6 & 20.22 & 5/11.28 & 7,092 & 221.0 & 0/125.1 \\
			
			\cellcolor{white}  &  \tool (Sym)  & 10.18 & 1.46 & \hl{130/1.34} & 55.73 & 3.11 & \hl{88/2.85} & 131.3 & 5.33 & \hl{70/4.72} & 158.5 & 3.99 & \hl{102/21.85} & 861.9 & 12.67 & \hl{22/279.9} \\
			\midrule
			
			\cellcolor{white} &  Naive & 69.15 & 7.95 & 64/0.77 & 251.6 & 24.58 & 0/1.52 & 623.1 & 55.42 & 0/2.38 & 620.6 & 32.43 & 1/11.29 & 7,542 & 226.1 & 0/125.3 \\
			
			\rowcolor{gray!20}
			\cellcolor{white}{8} &\tool (Int)   & 4.27 & 0.89 & 135/0.78 & 38.87 & 5.99 & 66/1.54 & 320.1 & 40.84 & 0/2.39 & 134.0 & 8.99 & 50/11.24 & 7,109 & 226.1 & 0/125.7 \\
			
			\cellcolor{white} &   \tool (Sym)  & 4.13 & 1.02 & \hl{136/1.35} & 34.01 & 2.14 & \hl{108/2.82} & 82.90 & 3.48 & \hl{86/4.61} & 96.26 & 2.39 & \hl{128/21.45} & 675.7 & 6.20 & \hl{27/273.6} \\
			\midrule
			
			\cellcolor{white} &  Naive & 69.18 & 7.96 & 65/0.77 & 252.0 & 24.63 & 0/1.52 & 624.0 & 55.55 & 0/2.36 & 620.4 & 32.40 & 1/11.19 & 7,559 & 226.9 & 0/124.2 \\
			
			\rowcolor{gray!20}
			\cellcolor{white}{10} & \tool (Int) & 2.72 & 0.56 & \hl{139/0.78} & 25.39 & 4.15 & 79/1.53 & 260.9 & 34.35 & 0/2.40 & 84.12 & 5.75 & 73/11.26 & 7,090 & 226.9 & 0/125.9 \\
			
			\cellcolor{white} &  \tool (Sym) & 2.61 & 0.92 & 139/1.35 & 28.59 & 1.91 & \hl{112/2.82} & 71.33 & 3.06 & \hl{92/4.56} & 81.08 & 2.01 & \hl{131/21.48} & 646.5 & 5.68 & \hl{31/271.5} \\
			
			\bottomrule
		\end{tabular}
	}
\end{table}

Unsurprisingly, \tool(Sym) is less efficient than the others but is still in the same order of magnitude. However, we can observe that \tool(Sym) solves the most tasks for both ACAS Xu and MNIST and produces the most accurate difference intervals of both hidden neurons and output neurons for almost all the tasks in MNIST, except for P1-8 and P1-10 where \tool(Int) performs better on the intervals for the output neurons.
We also find that \tool(Sym) may perform worse than the naive method when the quantization bit size is small for ACAS Xu.
It is because: 
 (1) the rounding error added into the abstract domain of the affine function in each hidden layer of QNNs is large due to the small bit size, and (2) such errors can accumulate and magnify layer by layer, in contrast to the naive approach where we directly apply the interval subtraction. We remark that symbolic-based reachability analysis methods for DNNs become less accurate 
 as the network gets deeper and the input region gets larger. It means that for a large input region, the output intervals of hidden/output neurons computed by symbolic interval analysis for DNNs can be very large. However, the output intervals of their quantized counterparts are always limited by the quantization grid limit, i.e., $[0,\frac{2^Q-1}{2^{Q-2}}]$. Hence, the difference intervals computed in Table~\ref{tab:DRA} can be very conservative for large input regions and deeper networks. 
 More experimental results of DRA on MNIST and ACAS Xu  are given in Appendix~\ref{app_sec:minist_DRA} and Appendix~\ref{app_sec:acas_DRA}, respectively.

\subsection{Effectiveness and Efficiency of \tool}\label{sec:exp2}

\begin{figure}[t]
	\centering
	\subfigure[ACAS Xu: $Q=4$]{\label{fig:verMQ4}
		\begin{minipage}[b]{0.23\textwidth}
			\includegraphics[width=1.0\textwidth]{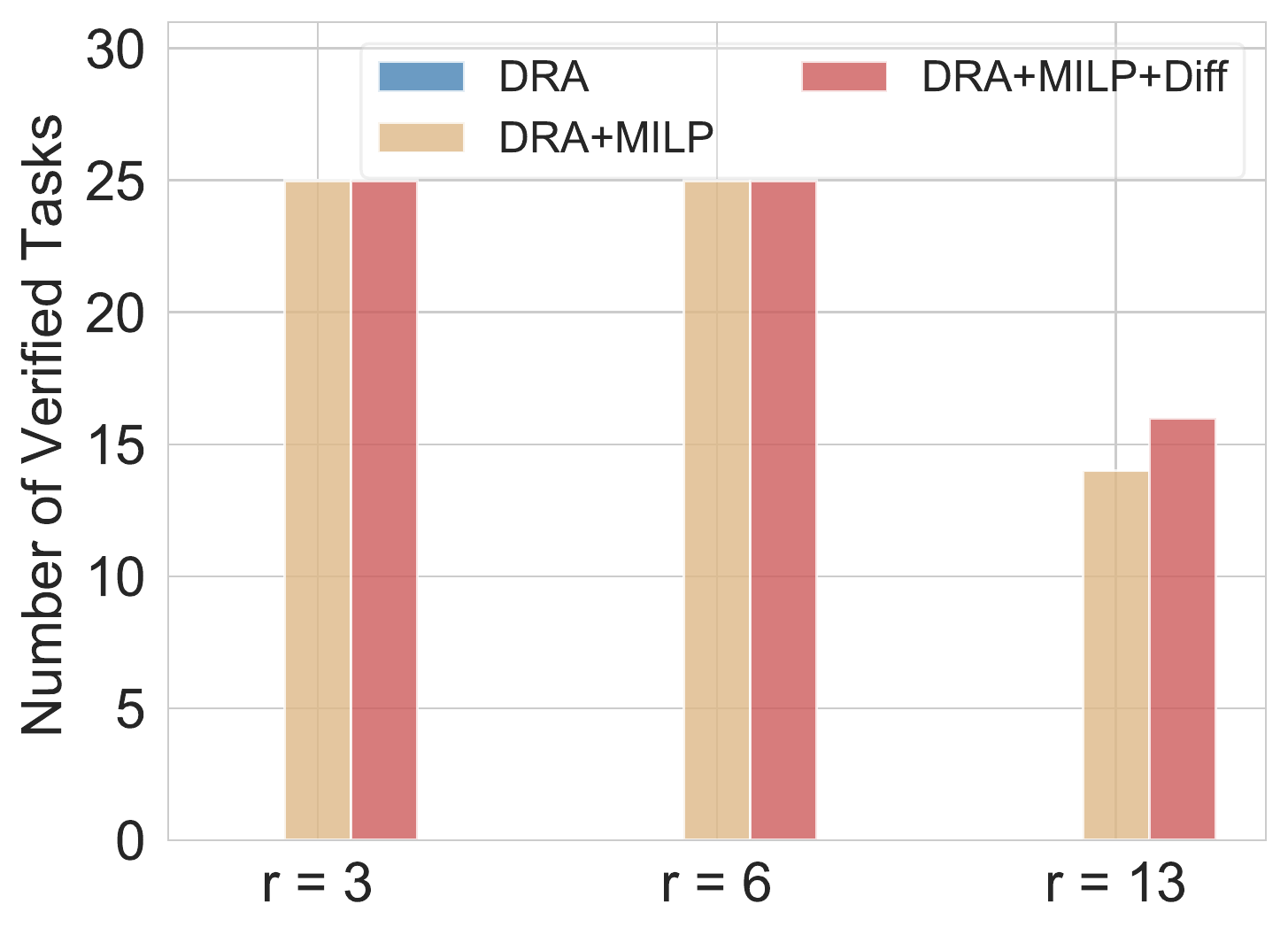}
		\end{minipage}	
	}\vspace*{-3mm}
	\subfigure[ACAS Xu: $Q=6$]{\label{fig:verMQ6}
		\begin{minipage}[b]{0.23\textwidth}
			\includegraphics[width=1.0\textwidth]{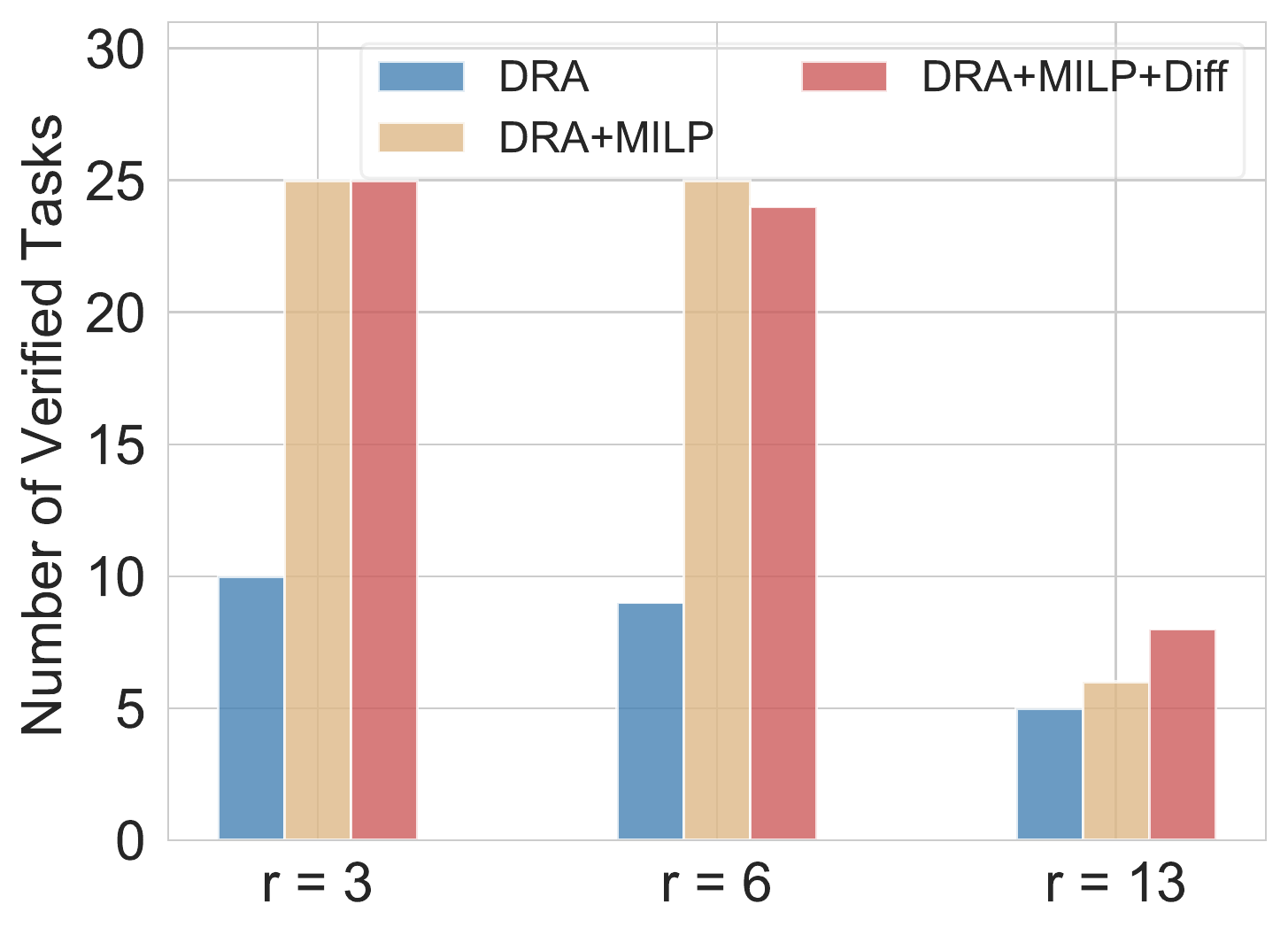}
		\end{minipage}	
	}
	\subfigure[ACAS Xu: $Q=8$]{\label{fig:verMQ8}
		\begin{minipage}[b]{0.23\textwidth}
			\includegraphics[width=1.0\textwidth]{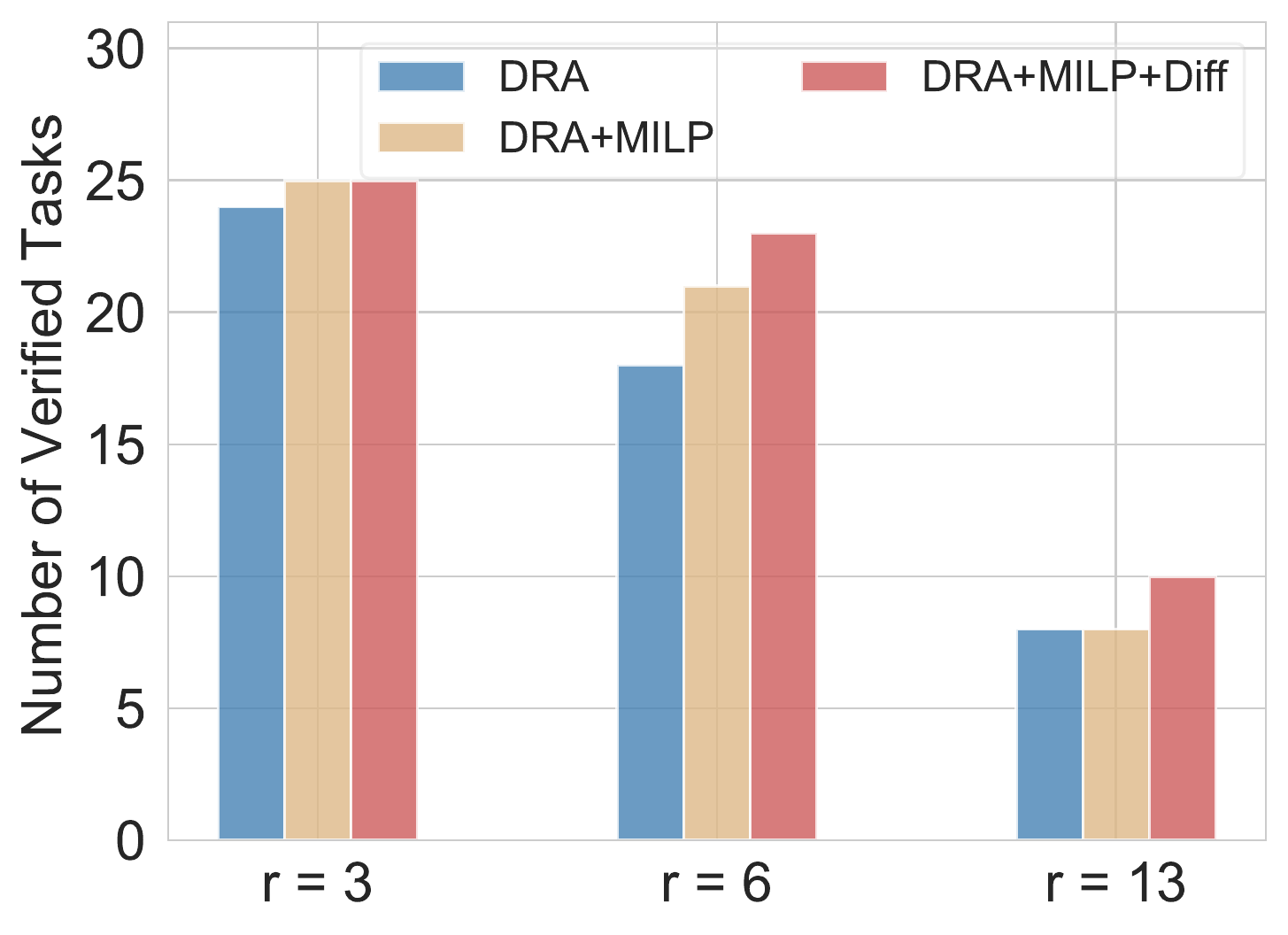}
		\end{minipage}	
	}
	\subfigure[ACAS Xu: $Q=10$]{\label{fig:verMQ10}
	\begin{minipage}[b]{0.23\textwidth}
		\includegraphics[width=1.0\textwidth]{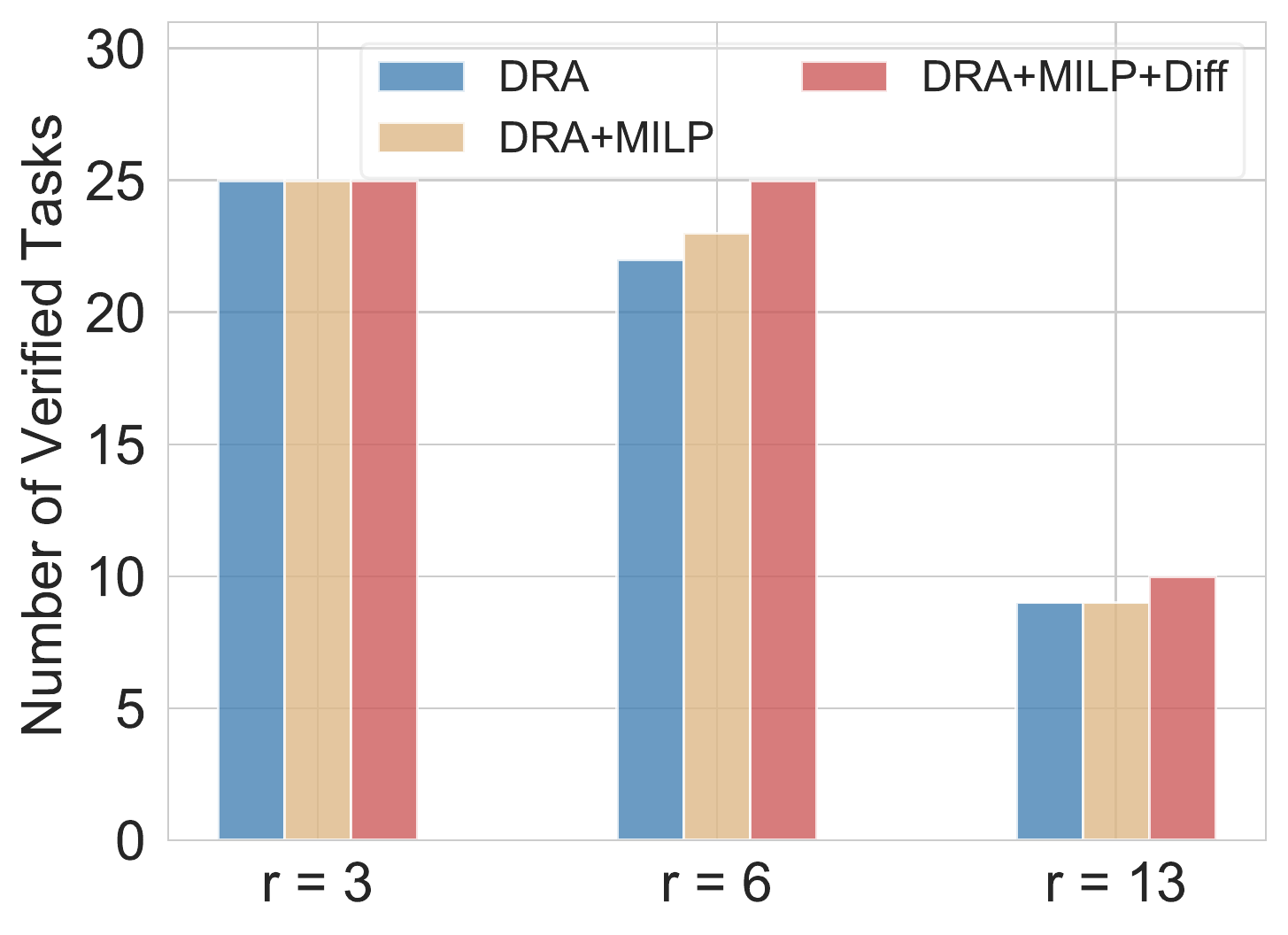}
	\end{minipage}	
	}
	\subfigure[MNIST: $Q=4$]{\label{fig:verMQ4}
		\begin{minipage}[b]{0.23\textwidth}
			\includegraphics[width=1.0\textwidth]{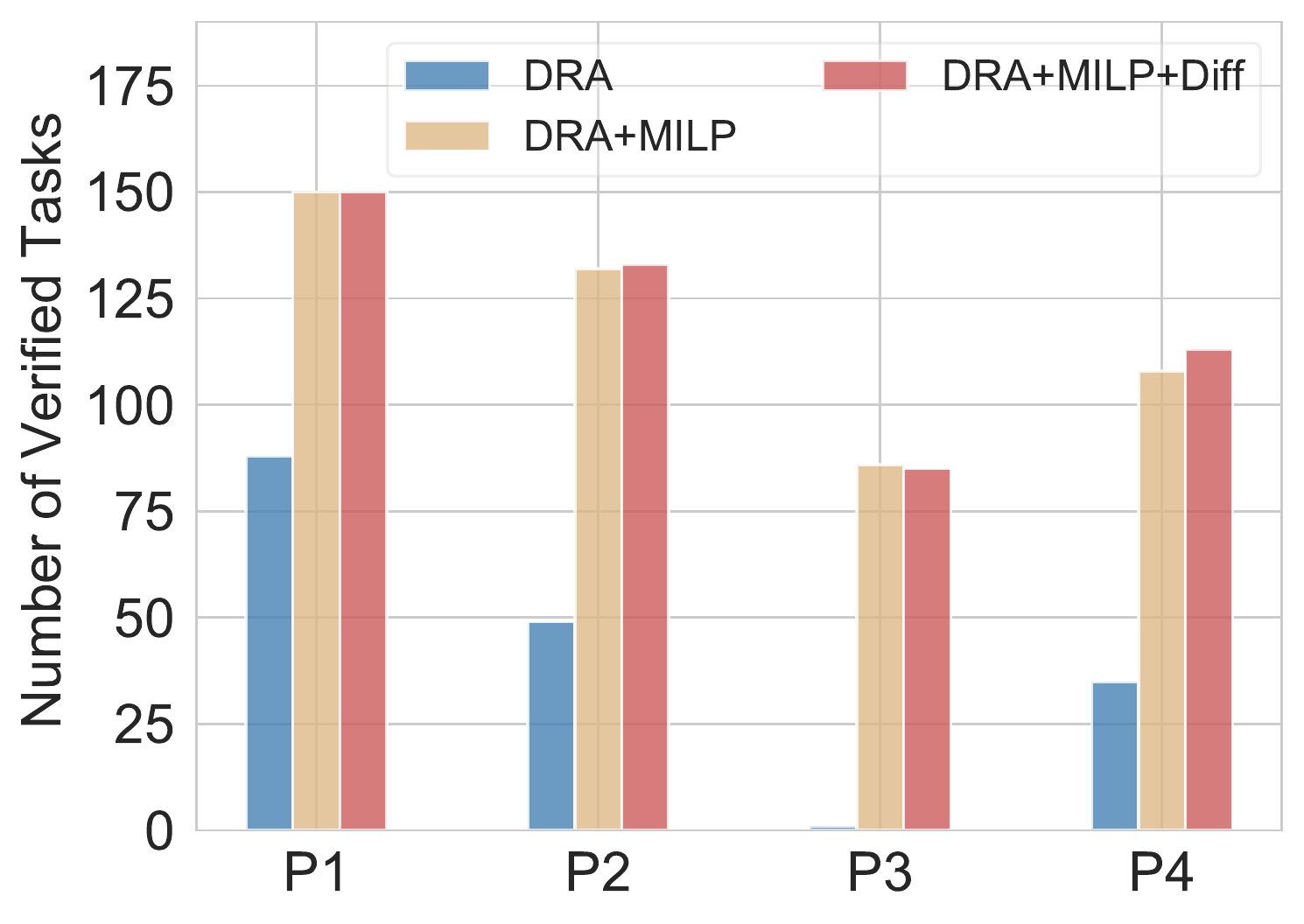}
		\end{minipage}	
	}
	\subfigure[MNIST: $Q=6$]{\label{fig:verMQ6}
		\begin{minipage}[b]{0.23\textwidth}
			\includegraphics[width=1.0\textwidth]{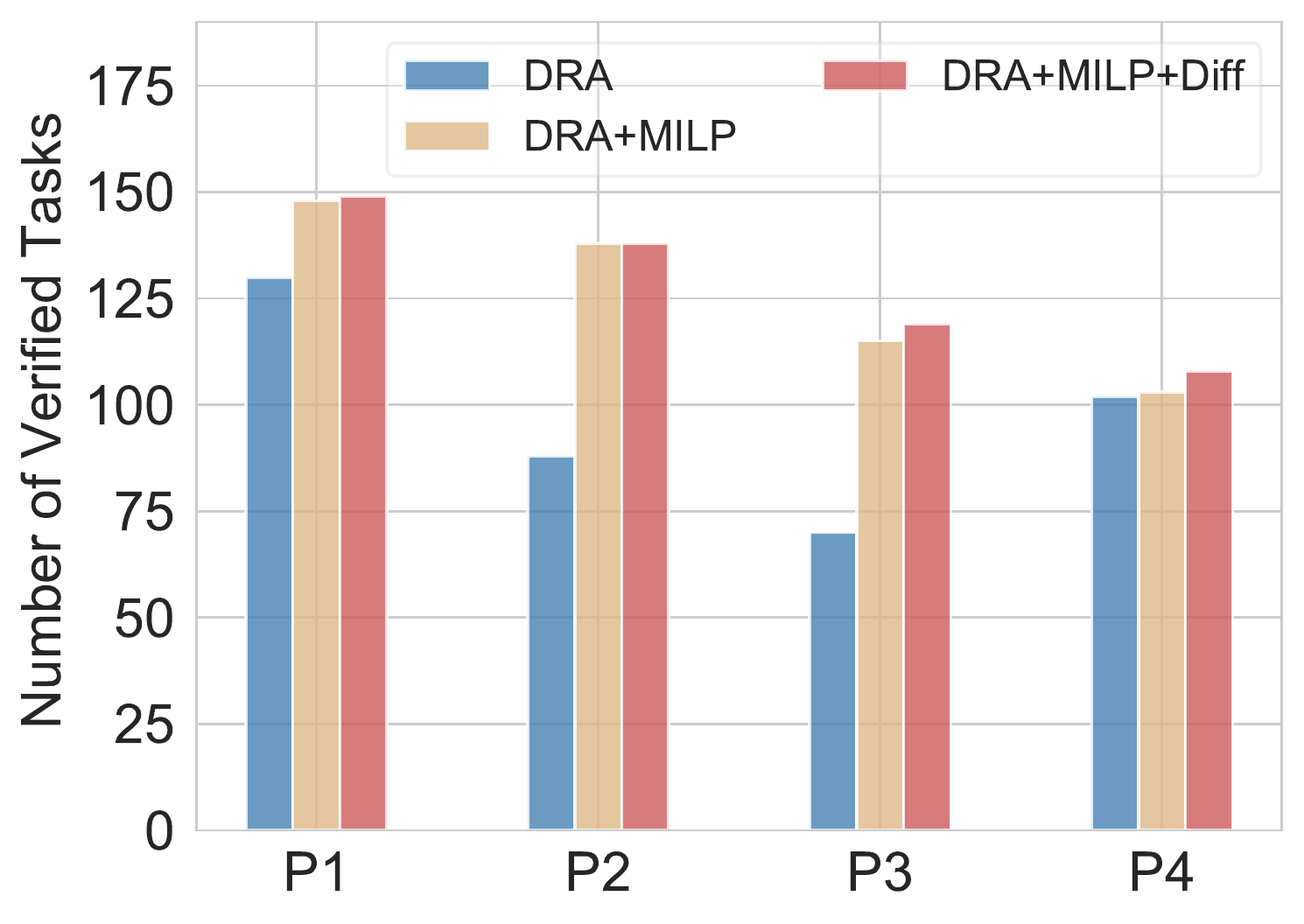}
		\end{minipage}	
	}
	\subfigure[MNIST: $Q=8$]{\label{fig:verMQ8}
		\begin{minipage}[b]{0.23\textwidth}
			\includegraphics[width=1.0\textwidth]{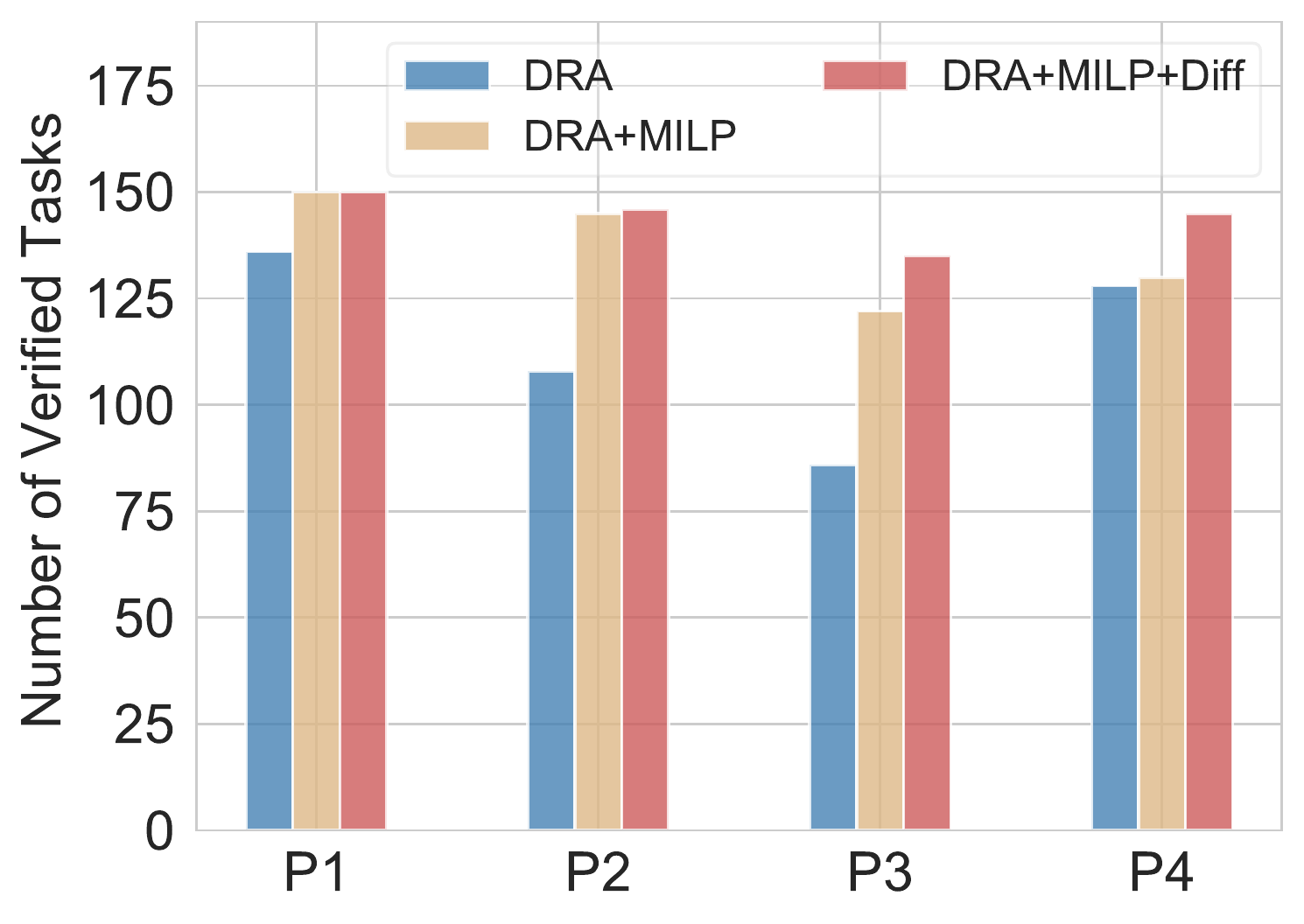}
		\end{minipage}	
	}
	\subfigure[MNIST: $Q=10$]{\label{fig:verMQ10}
	\begin{minipage}[b]{0.23\textwidth}
		\includegraphics[width=1.0\textwidth]{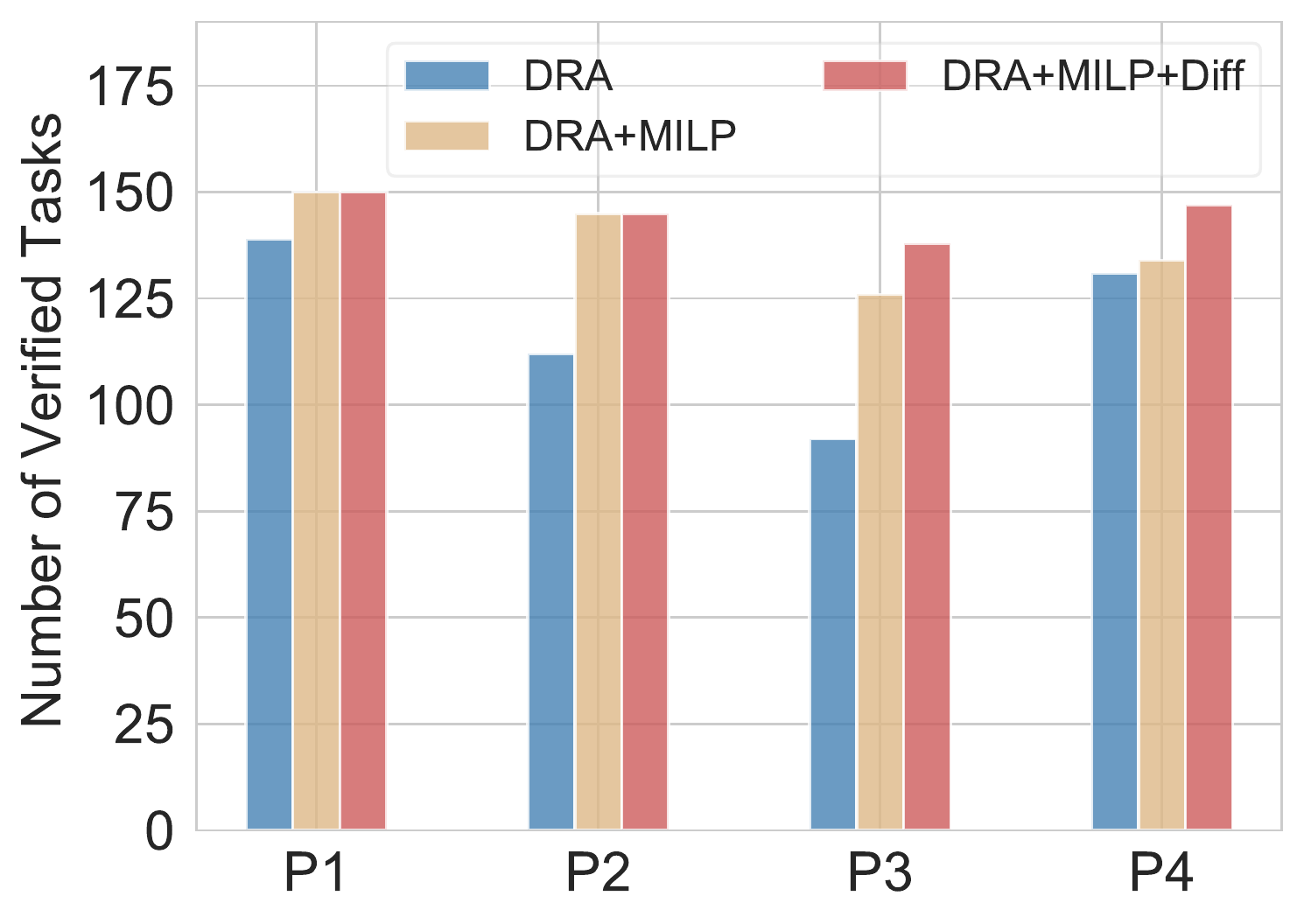}
	\end{minipage}	
	}
	\caption{Verification Results of \tool on ACAS Xu and MNIST}
	\vspace*{-3mm}
    \label{fig:qebVerif}
\end{figure}

We evaluate \tool on QNNs A$x$-$z$, P$y$-$z$ for $x=1$, $y\in\{1,2,3,4\}$ and $z\in\{4,6,8,10\}$, as well as DNNs correspondingly. We use the same input regions and error bounds as in Section~\ref{sec:RQ1} except that we consider $r=\{3,6,13\}$ for each input point for ACAS Xu.
Note that, we omit the other two radii for ACAS Xu and use medium-sized QNNs for MNIST as our evaluation benchmarks of this experiment for the sake of time and computing resources.

Figure~\ref{fig:qebVerif} shows the verification results of \tool within 1 hour per task, which gives the number of successfully verified tasks with three methods
(Complete results refer to Appendix~\ref{app_sec:complete_res}). Note that only the number of successfully proved tasks is given in Figure~\ref{fig:qebVerif} for DRA due to its incompleteness.
The blue bars show the results using only the symbolic differential reachability analysis, i.e., \tool(Sym). The yellow bars give the results by a full verification process in \tool as shown in Figure~\ref{fig:overview}, i.e., we first use DRA and then use MILP solving if DRA fails. The red bars are similar to the yellow ones except that linear constraints of the difference intervals of hidden neurons got from DRA are added into the MILP encoding.

Overall, although DRA successfully proved most of the tasks (60.19\% with DRA solely), our MILP-based verification method can help further verify many tasks
on which DRA fails, namely, 85.67\% with DRA+MILP and 88.59\% with DRA+MILP+Diff. Interestingly, we find that the effectiveness of the added linear constraints of the difference intervals varies
on the MILP solving efficiency on different tasks. Our conjecture is that there are some heuristics in the Gurobi solving algorithm for which the additional constraints may not always be helpful.
However, those difference linear constraints allow the MILP-based verification method to verify more tasks, i.e., 79 tasks more in total.
More experimental results on ACAS Xu refer to Appendix~\ref{app_sec:acas_tool}.

\subsection{Correlation of Quantization Errors and Robustness}

We use \tool to verify a set of properties $\Psi = \{P(\mathcal{N},\widehat{\mathcal{N}},\hat{\bs{x}},r,\epsilon)\}$, where
$\mathcal{N}=\text{P1-Full}$, $\widehat{\mathcal{N}}\in\{\text{P1-4, P1-8}\} $,  $\hat{\bs{x}}\in\mathcal{X}$ and $\mathcal{X}$ is the set of the 30 samples from MNIST as above, $r\in\{3,5,7\}$ and $\epsilon\in \Omega =\{0.5,1.0,1.5,2.0,2.5,3.0,3.5,4.0,5.0\}$. We solve all the above tasks and process all the results to obtain the tightest range of quantization error bounds $[a,b]$ for each input region such that $a,b\in\Omega$.
It allows us to obtain intervals that are tighter than those obtained via DRA.
Finally, we implemented a robustness verifier for QNNs in a way similar to~\cite{mistry2022milp} to check the robustness of P1-4 and P1-8 w.r.t. the input regions given in $\Psi$.

\begin{figure}[t]
	\centering
	\subfigure[Robustness Results for P1-4.]{\label{fig:robQ4}
		\begin{minipage}[b]{0.225\textwidth}
			\includegraphics[width=1.0\textwidth]{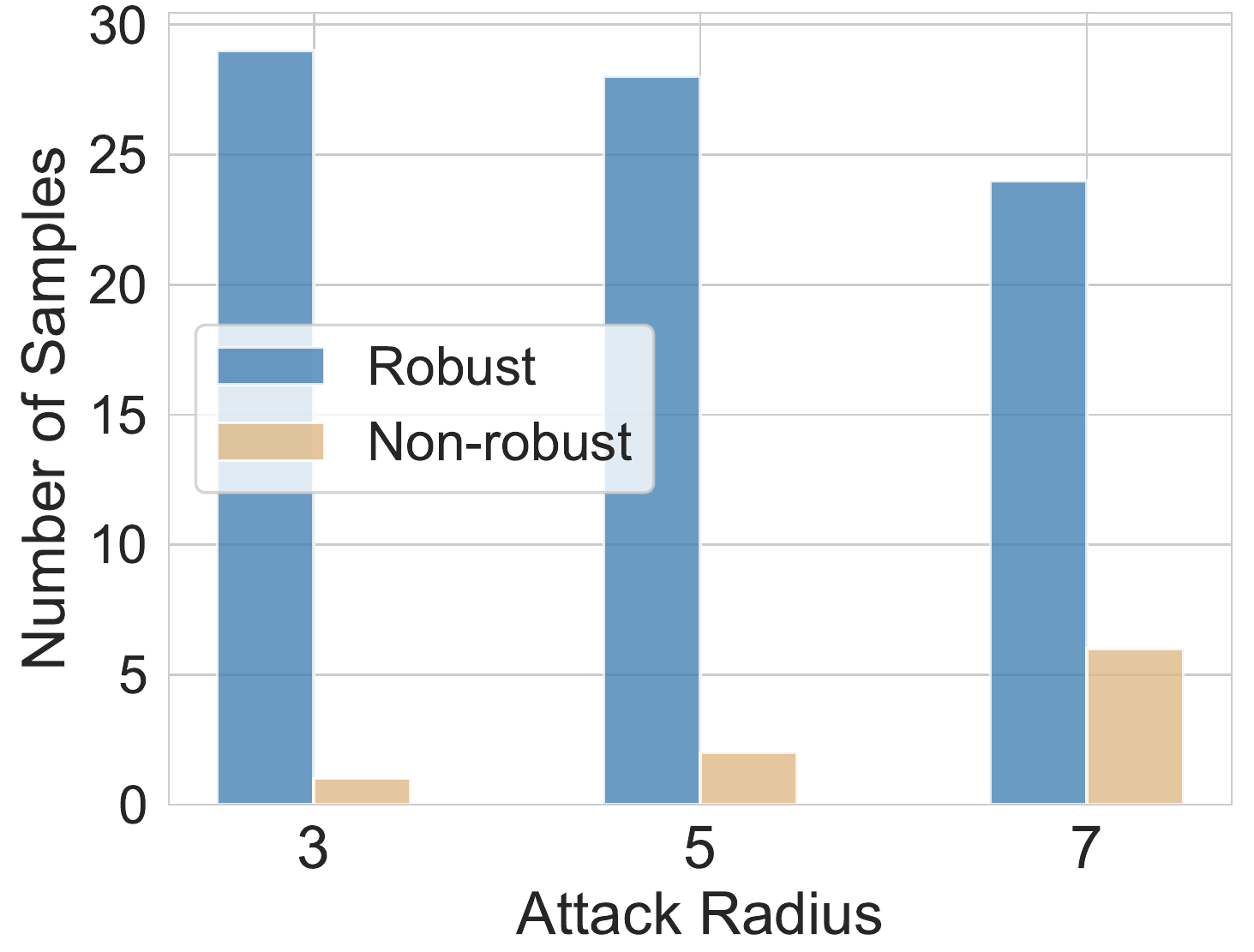}
		\end{minipage}	
	}
	\subfigure[Errors for P1-4 under $r=3$.]{\label{fig:errQ4R3}
		\begin{minipage}[b]{0.225\textwidth}
			\includegraphics[width=1.0\textwidth]{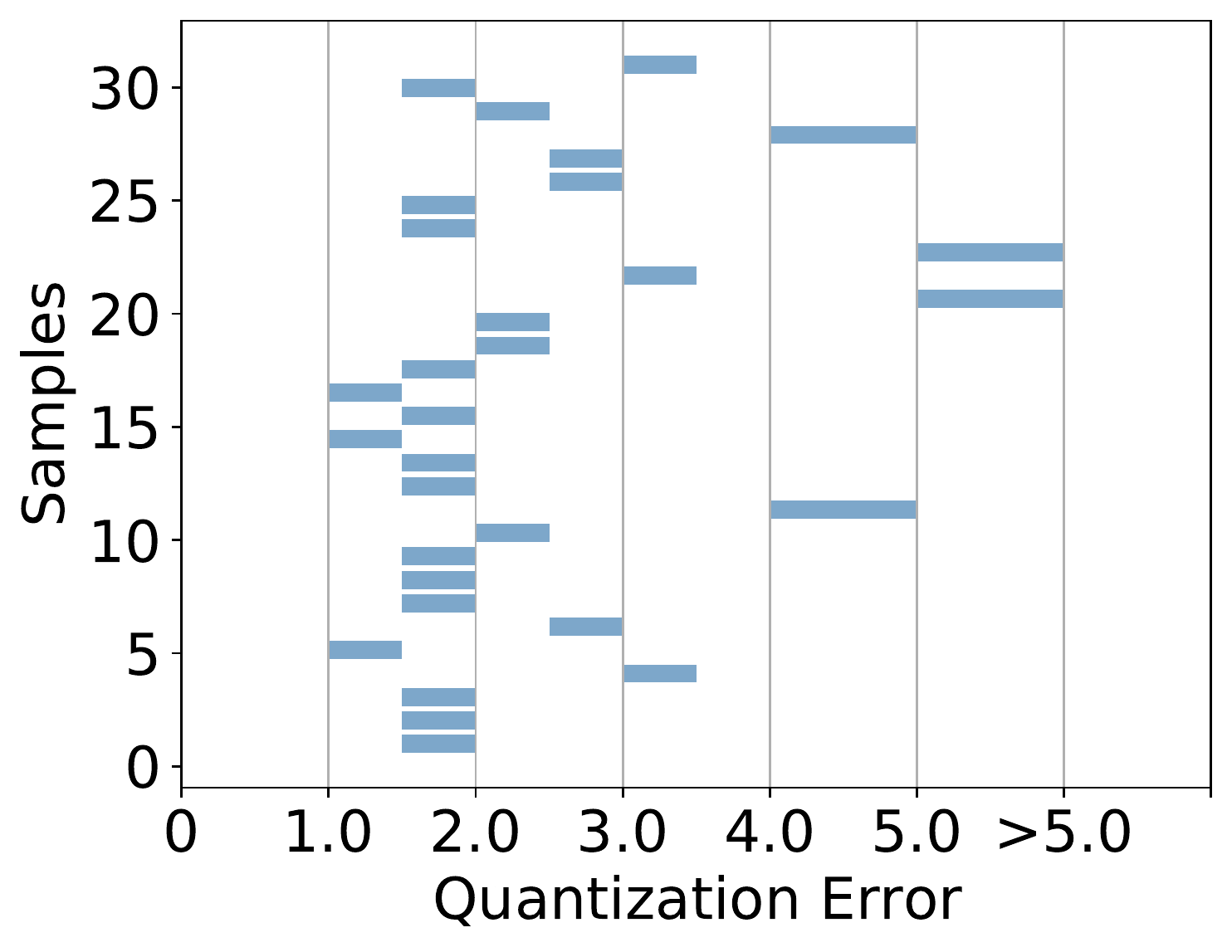}
		\end{minipage}	
	}
	\subfigure[Errors for P1-4 under $r=5$.]{\label{fig:errQ4R5}
		\begin{minipage}[b]{0.225\textwidth}
			\includegraphics[width=1.0\textwidth]{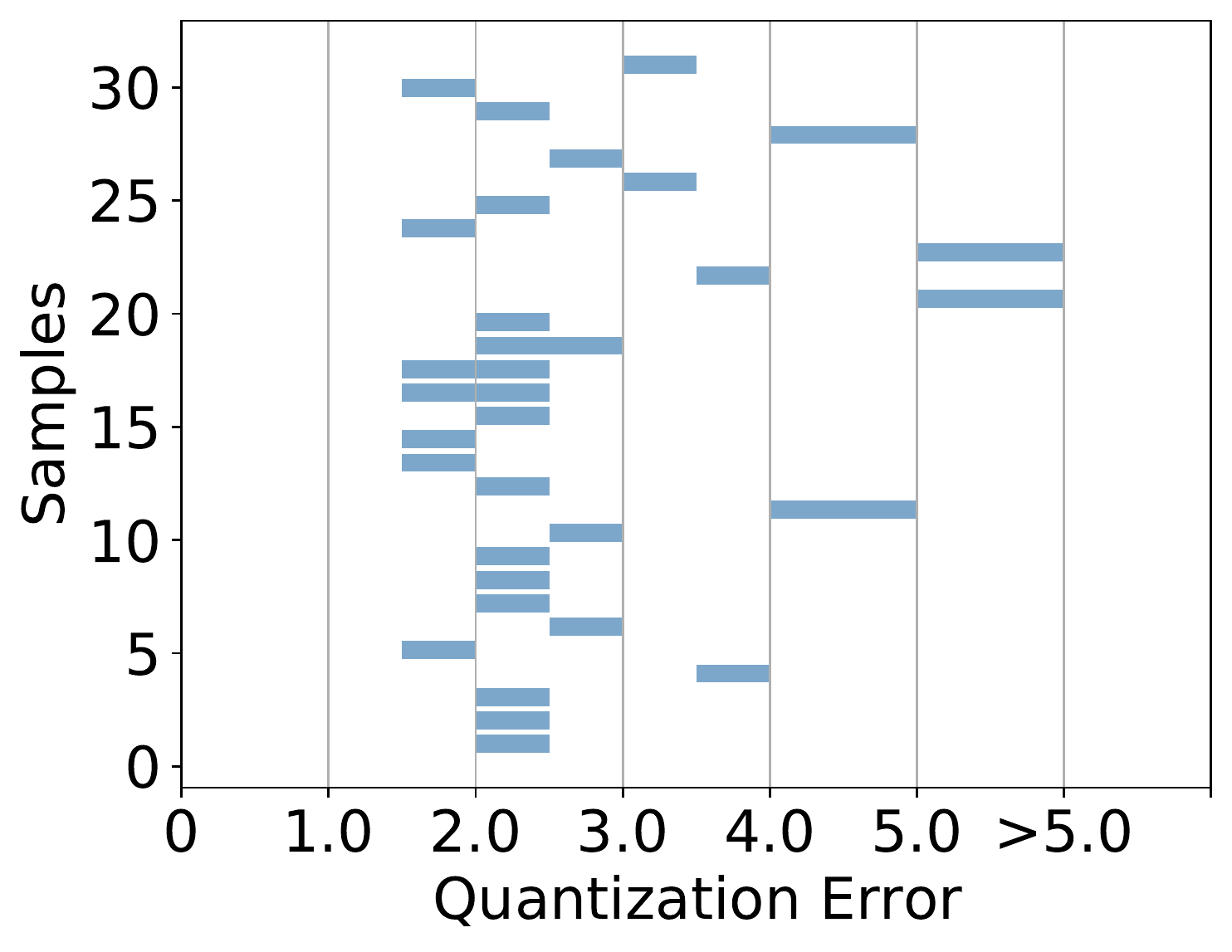}
		\end{minipage}	
	}
	\subfigure[Errors for P1-4 under $r=7$.]{\label{fig:errQ4R7}
	\begin{minipage}[b]{0.225\textwidth}
		\includegraphics[width=1.0\textwidth]{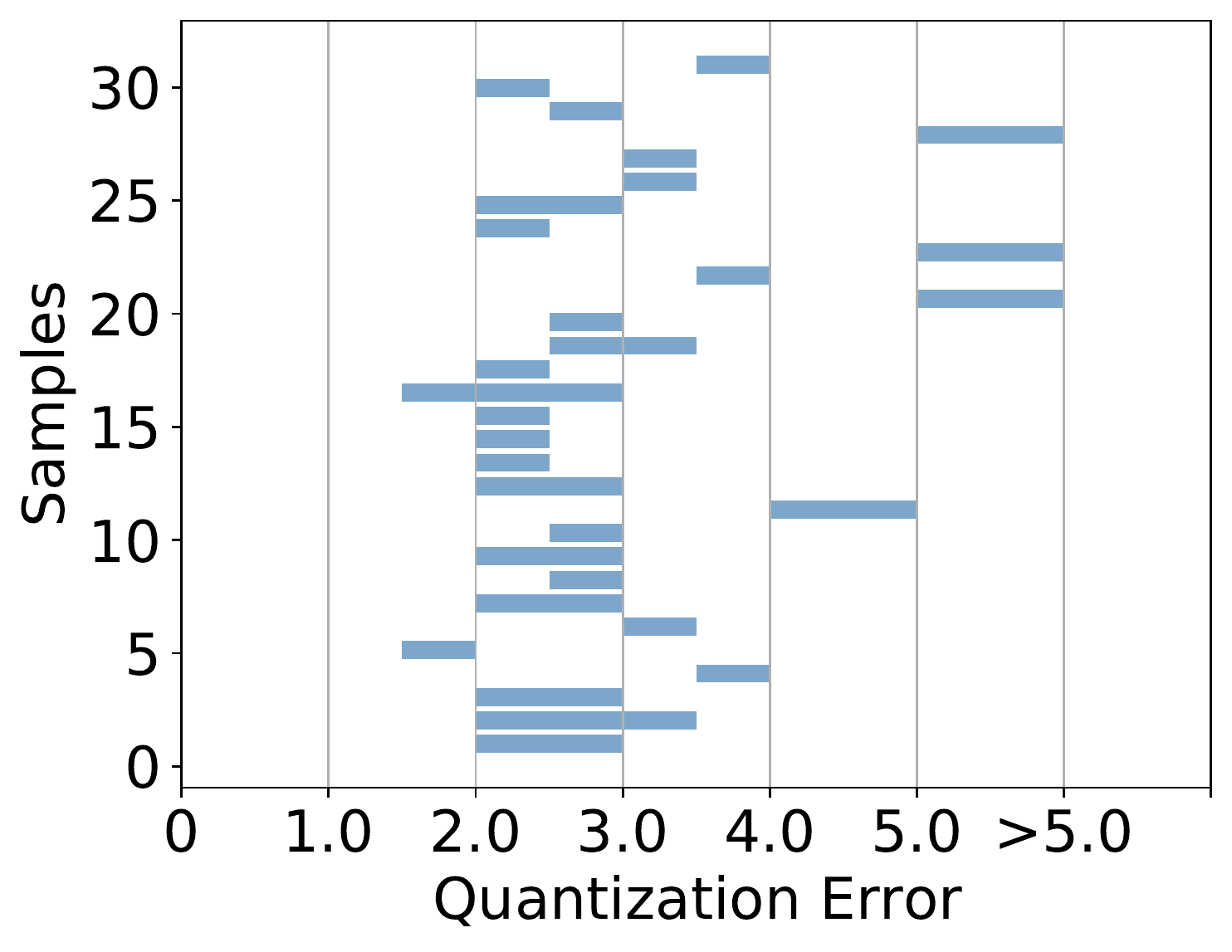}
	\end{minipage}	
	}
	\subfigure[Robustness Results for P1-8.]{\label{fig:robQ8}
	\begin{minipage}[b]{0.225\textwidth}
		\includegraphics[width=1.0\textwidth]{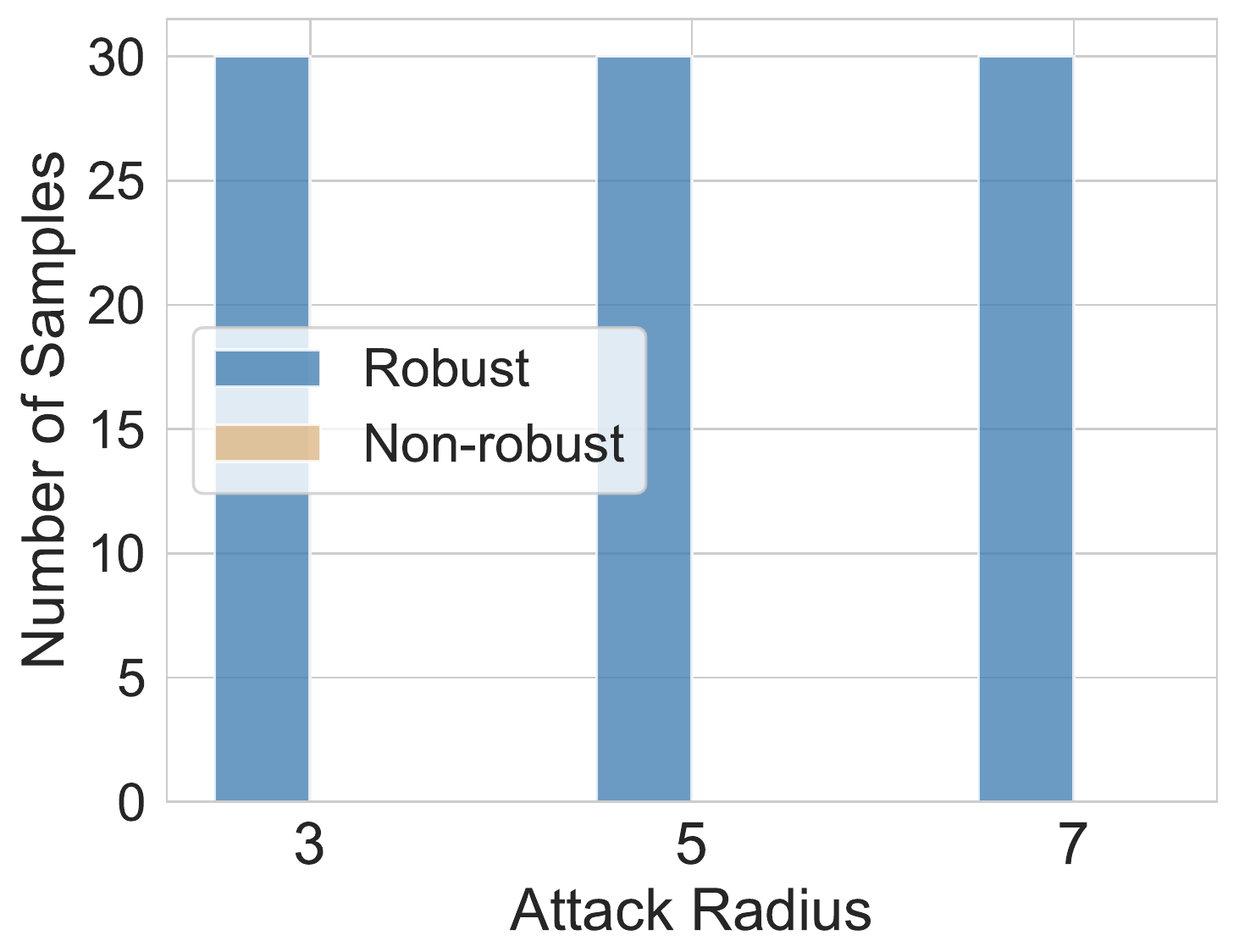}
	\end{minipage}	
	}
	\subfigure[Errors for P1-8 under $r=3$.]{\label{fig:errQ8R3}
		\begin{minipage}[b]{0.225\textwidth}
			\includegraphics[width=1.0\textwidth]{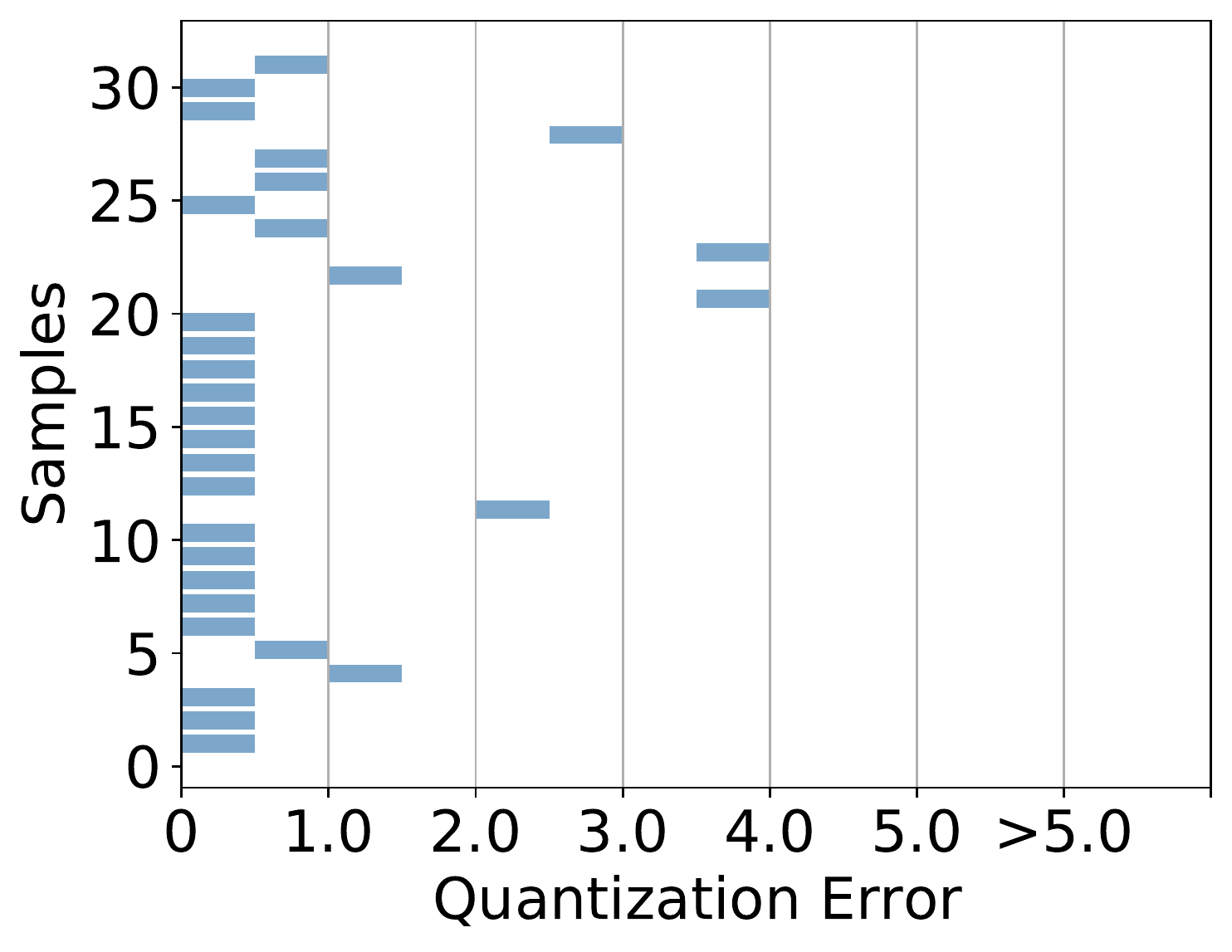}
		\end{minipage}	
	}
	\subfigure[Errors for P1-8 under $r=5$.]{\label{fig:errQ8R5}
		\begin{minipage}[b]{0.225\textwidth}
			\includegraphics[width=1.0\textwidth]{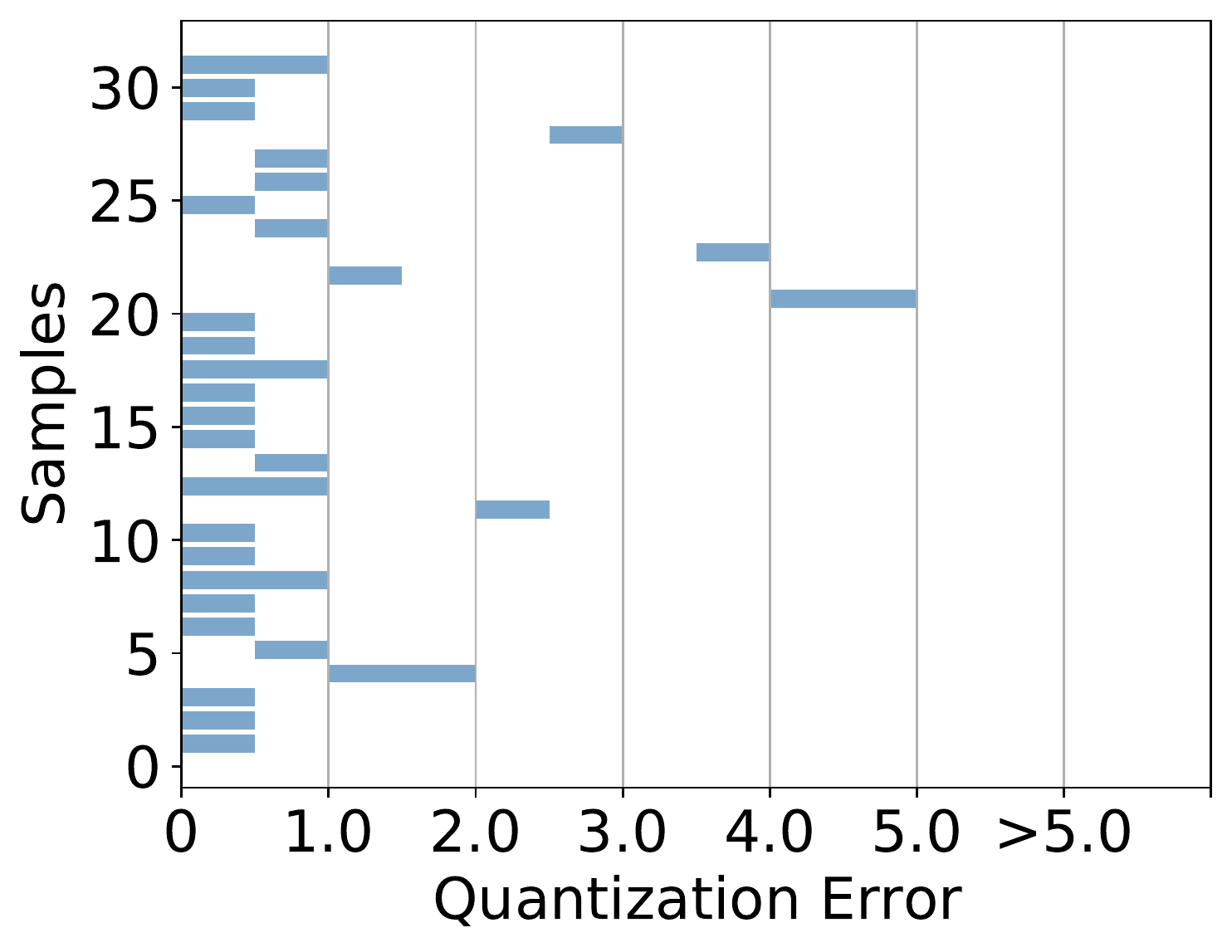}
		\end{minipage}	
	}
	\subfigure[Errors for P1-8 under $r=7$.]{\label{fig:errQ8R7}
	\begin{minipage}[b]{0.225\textwidth}
		\includegraphics[width=1.0\textwidth]{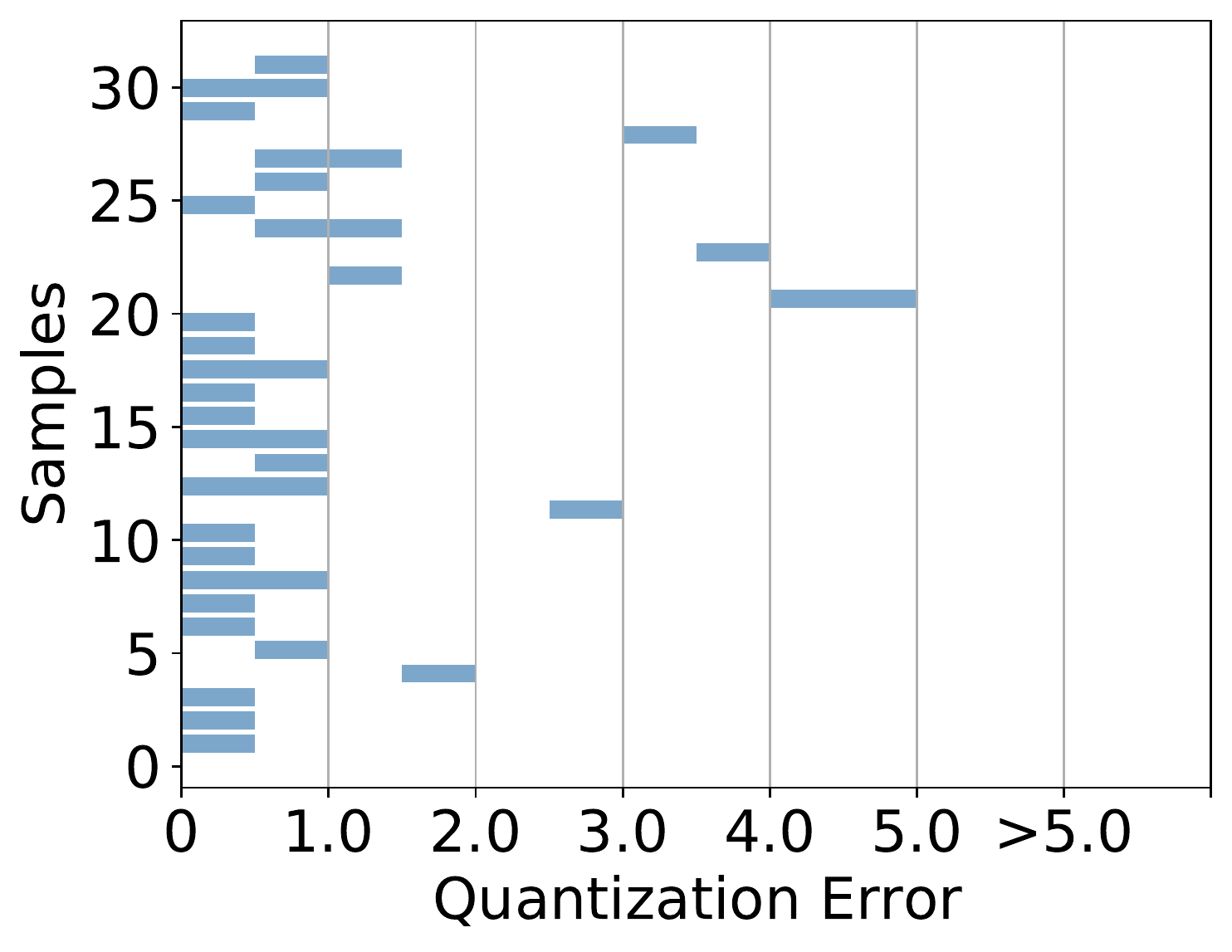}
	\end{minipage}	
	}
	\caption{Distribution of (non-)robust samples and Quantization Errors under radius $r$ and quantization bits $Q$.}
	\vspace*{-2mm}
    \label{fig:relation}
\end{figure}

Figure~\ref{fig:relation} gives the experimental results. The blue (resp. yellow) bars in Figures~\ref{fig:robQ4} and \ref{fig:robQ8} show the number of robust (resp. non-robust) samples among the
30 verification tasks, and blue bars in the other 6 figures demonstrate the quantization error interval for each input region.
By comparing the results of P1-8 and P1-4, we observe that P1-8 is more robust than P1-4 w.r.t. the 90 input regions and its quantization errors
are also generally much smaller than that of P1-4.
Furthermore, we find that P1-8 remains consistently robust as the radius increases, and its quantization error interval changes very little.
However, P1-4 becomes increasingly less robust as the radius increases and its quantization error also increases significantly.
 Thus, we speculate that there may be some correlation between network robustness and quantization error in QNNs. Specifically, as the quantization bit size
 decreases, the quantization error increases and the QNN becomes less robust.
The reason we suspect ``the fewer bits, the less robust'' is that
with fewer bits, a perturbation may easily cause significant change on hidden neurons (i.e., the change is magnified by the loss of precision) and consequently the output.
Furthermore, the correlation between the quantization error bound and the empirical robustness of the QNN suggests that it is indeed possible to apply our method to compute the quantization error bound and use it as a guide for identifying the best quantization scheme which balances the size of the model and its robustness.

\section{Related Work}\label{sec:relatedwork}


While there is a large and growing body of work on quality assurance techniques for neural networks
 including testing (e.g.,~\cite{CW17b,PCYJ17,TianPJR18,OdenaOAG19,ZhangHML22,ZhaoCWYS021,ChenCFDZSL21,CZSCFL22,DZBS21,SongLCFL21}) 
and formal verification (e.g.,~\cite{PT10,Ehl17,KBDJK17,KatzHIJLLSTWZDK19,HKWW17,SGPV19,ElboherGK20,YLLHWSXZ20,LiuSZW20,GuoWZZSW21,LXSSXM22,AndersonPDC19,GMDTCV18,LiLYCHZ19,LiLHY0ZXH20,SinghGPV19,WPWYJ18,TranBXJ20,TranLMYNXJ19,ZZCSCL22,chen2022towards}).
Testing techniques are often effective in finding violations, but they cannot prove their absence.
While formal verification can prove their absence, existing methods typically target real-valued neural networks, i.e., DNNs, and are not effective in verifying quantization error bound~\cite{paulsen2020reludiff}.
In this section, we mainly discuss the existing verification techniques for QNNs.

Early work on formal verification of QNNs typically focuses on 1-bit quantized neural networks (i.e., BNNs)~\cite{narodytska2018verifying,NPAQ19,CSSDvnn19,ddlearning19A,ddlearning19B,BDD4BNN,ZZCSC23}. Narodytska et al.~\cite{narodytska2018verifying} first proposed to reduce the verification problem of BNNs to a satisfiability problem of a Boolean formula or an integer linear programming problem.
Baluta et al.~\cite{NPAQ19} proposed a PAC-style quantitative analysis framework for BNNs via approximate SAT model-counting solvers.
 Shih et al. proposed a quantitative verification framework for BNNs~\cite{ddlearning19A,ddlearning19B} via a BDD learning-based method~\cite{Nakamura05}.
 Zhang et al.~\cite{BDD4BNN,ZZCSC23} proposed a BDD-based verification framework for BNNs, which exploits the internal structure of the BNNs to construct BDD models instead of BDD-learning.
Giacobbe et al.~\cite{GiacobbeHL20} pushed this direction further by introducing the first formal verification for multiple-bit quantized DNNs (i.e., QNNs)
by encoding the robustness verification problem into an SMT formula based on the first-order theory of quantifier-free bit-vector.
Later, Henzinger et al.~\cite{scaleQNN21} explored several heuristics to improve the efficiency and scalability of \cite{GiacobbeHL20}.
Very recently, \cite{zhang2022qvip} and \cite{mistry2022milp} proposed an ILP-based method and an MILP-based verification method for QNNs, respectively, and both outperforms the SMT-based verification approach~\cite{scaleQNN21}.
Though these works can directly verify QNNs or BNNs, they cannot verify quantization error bounds.

There are also some works focusing on exploring the properties of two neural networks which are most closely related to our work.
Paulsen et al.~\cite{paulsen2020reludiff,PaulsenWWW20}
proposed differential verification methods to verify two DNNs with the same network topology.
This idea has been extended to handle recurrent neural networks~\cite{MohammadinejadP21}.
The difference between \cite{paulsen2020reludiff,PaulsenWWW20,MohammadinejadP21} and our work has been discussed throughout this work, i.e.,
they focus on quantized weights and cannot handle quantized activation tensors. Moreover, their methods are not complete, thus would fail to prove tighter error bounds.
%
 Semi-definite program was used to analyze the different behaviors of DNNs and \emph{fully} QNNs~\cite{li2021robust}.
 Different from our work focusing on verification, they aim at generating an upper bound for the worst-case error induced by quantization. 
 Furthermore, \cite{li2021robust} only scales to tiny QNNs, e.g., 1 input neuron, 1 output neuron, and 10 neurons per hidden layer (up to 4 hidden layers).
 In comparison, our differential reachability analysis scales to much larger QNNs, e.g., QNN with 4890 neurons.

\section{Conclusion}\label{sec:conclu}
In this work, we proposed a novel quantization error bound verification method \tool which is sound, complete, and arguably efficient. We implemented it as an end-to-end tool and conducted thorough experiments on various QNNs with different quantization bit sizes. Experimental results showed the effectiveness and the efficiency of \tool. We also investigated the potential correlation between robustness and quantization errors for QNNs and found that as the quantization error increases the QNN might become less robust. For further work, it would be interesting to investigate the verification method for other activation functions and network architectures, towards which this work makes a significant step.

\subsubsection{Acknowledgements.} This work is supported by the National Key Research Program
(2020AAA0107800), National Natural Science Foundation of China (62072309), CAS Project for Young Scientists in Basic
Research (YSBR-040), ISCAS New Cultivation Project (ISCAS-PYFX-202201),
and the Ministry of Education, Singapore under its Academic
Research Fund Tier 3 (MOET32020-0004). Any opinions,
findings and conclusions or recommendations expressed in this
material are those of the authors and do not reflect the views of the
Ministry of Education, Singapore.

%
%
%

\bibliographystyle{splncs04}
\bibliography{yd-bibfile}

\begin{thebibliography}{10}
\providecommand{\url}[1]{\texttt{#1}}
\providecommand{\urlprefix}{URL }
\providecommand{\doi}[1]{https://doi.org/#1}

\bibitem{AWBK20}
Amir, G., Wu, H., Barrett, C.W., Katz, G.: An smt-based approach for verifying
  binarized neural networks. In: Proceedings of the 27th International
  Conference on Tools and Algorithms for the Construction and Analysis of
  Systems. pp. 203--222 (2021)

\bibitem{AndersonPDC19}
Anderson, G., Pailoor, S., Dillig, I., Chaudhuri, S.: Optimization and
  abstraction: a synergistic approach for analyzing neural network robustness.
  In: Proceedings of the 40th {ACM} {SIGPLAN} Conference on Programming
  Language Design and Implementation. pp. 731--744 (2019)

\bibitem{NPAQ19}
Baluta, T., Shen, S., Shinde, S., Meel, K.S., Saxena, P.: Quantitative
  verification of neural networks and its security applications. In:
  Proceedings of the 2019 ACM SIGSAC Conference on Computer and Communications
  Security. pp. 1249--1264 (2019)

\bibitem{DZBS21}
Bu, L., Zhao, Z., Duan, Y., Song, F.: Taking care of the discretization
  problem: {A} comprehensive study of the discretization problem and a
  black-box adversarial attack in discrete integer domain. {IEEE} Trans.
  Dependable Secur. Comput.  \textbf{19}(5),  3200--3217 (2022)

\bibitem{CW17b}
Carlini, N., Wagner, D.A.: Towards evaluating the robustness of neural
  networks. In: Proceedings of the 2017 {IEEE} Symposium on Security and
  Privacy. pp. 39--57 (2017)

\bibitem{ChenCFDZSL21}
Chen, G., Chen, S., Fan, L., Du, X., Zhao, Z., Song, F., Liu, Y.: Who is real
  {Bob}? adversarial attacks on speaker recognition systems. In: Proceedings of
  the 42nd {IEEE} Symposium on Security and Privacy. pp. 694--711 (2021)

\bibitem{CZSCFL22}
Chen, G., Zhao, Z., Song, F., Chen, S., Fan, L., Liu, Y.: {AS2T}: Arbitrary
  source-to-target adversarial attack on speaker recognition systems. {IEEE}
  Trans. Dependable Secur. Comput. pp. 1--17 (2022)

\bibitem{chen2022towards}
Chen, G., Zhao, Z., Song, F., Chen, S., Fan, L., Wang, F., Wang, J.: Towards
  understanding and mitigating audio adversarial examples for speaker
  recognition. {IEEE} Trans. Dependable Secur. Comput. pp. 1--17 (2022)

\bibitem{CSSDvnn19}
Choi, A., Shi, W., Shih, A., Darwiche, A.: Compiling neural networks into
  tractable boolean circuits. In: Proceedings of the AAAI Spring Symposium on
  Verification of Neural Networks (2019)

\bibitem{CousotC77}
Cousot, P., Cousot, R.: Abstract interpretation: {A} unified lattice model for
  static analysis of programs by construction or approximation of fixpoints.
  In: Conference Record of the Fourth {ACM} Symposium on Principles of
  Programming Languages. pp. 238--252 (1977)

\bibitem{duncan2020relative}
Duncan, K., Komendantskaya, E., Stewart, R., Lones, M.: Relative robustness of
  quantized neural networks against adversarial attacks. In: Proceedings of the
  International Joint Conference on Neural Networks. pp.~1--8 (2020)

\bibitem{Ehl17}
Ehlers, R.: Formal verification of piece-wise linear feed-forward neural
  networks. In: Proceedings of the 15th International Symposium on Automated
  Technology for Verification and Analysis. pp. 269--286 (2017)

\bibitem{ElboherGK20}
Elboher, Y.Y., Gottschlich, J., Katz, G.: An abstraction-based framework for
  neural network verification. In: Proceedings of the 32nd International
  Conference on Computer Aided Verification. pp. 43--65 (2020)

\bibitem{EEF0RXPKS18}
Eykholt, K., Evtimov, I., Fernandes, E., Li, B., Rahmati, A., Xiao, C.,
  Prakash, A., Kohno, T., Song, D.: Robust physical-world attacks on deep
  learning visual classification. In: Proceedings of the {IEEE} Conference on
  Computer Vision and Pattern Recognition. pp. 1625--1634 (2018)

\bibitem{GMDTCV18}
Gehr, T., Mirman, M., Drachsler{-}Cohen, D., Tsankov, P., Chaudhuri, S.,
  Vechev, M.T.: {AI$^2$:} safety and robustness certification of neural
  networks with abstract interpretation. In: Proceedings of the {IEEE}
  Symposium on Security and Privacy. pp. 3--18 (2018)

\bibitem{GiacobbeHL20}
Giacobbe, M., Henzinger, T.A., Lechner, M.: How many bits does it take to
  quantize your neural network? In: Proceedings of the 26th International
  Conference on Tools and Algorithms for the Construction and Analysis of
  Systems. pp. 79--97 (2020)

\bibitem{DSQ}
Gong, R., Liu, X., Jiang, S., Li, T., Hu, P., Lin, J., Yu, F., Yan, J.:
  Differentiable soft quantization: Bridging full-precision and low-bit neural
  networks. Proceedings of the IEEE/CVF International Conference on Computer
  Vision pp. 4851--4860 (2019)

\bibitem{TFLiteWebpage}
Google: Tensorflow lite. \url{https://www.tensorflow.org/lite} (2022)

\bibitem{GuoWZZSW21}
Guo, X., Wan, W., Zhang, Z., Zhang, M., Song, F., Wen, X.: Eager falsification
  for accelerating robustness verification of deep neural networks. In:
  Proceedings of the 32nd {IEEE} International Symposium on Software
  Reliability Engineering. pp. 345--356 (2021)

\bibitem{Gurobi}
Gurobi: A most powerful mathematical optimization solver.
  \url{https://www.gurobi.com/} (2018)

\bibitem{HanMD15}
Han, S., Mao, H., Dally, W.J.: Deep compression: Compressing deep neural
  network with pruning, trained quantization and huffman coding. In:
  Proceedings of the 4th International Conference on Learning Representations
  (2016)

\bibitem{scaleQNN21}
Henzinger, T.A., Lechner, M., Zikelic, D.: Scalable verification of quantized
  neural networks. In: Proceedings of the AAAI Conference on Artificial
  Intelligence. vol.~35, pp. 3787--3795 (2021)

\bibitem{HDYDMJSVNSK12}
Hinton, G., Deng, L., Yu, D., Dahl, G.E., Mohamed, A., Jaitly, N., Senior, A.,
  Vanhoucke, V., Nguyen, P., Sainath, T.N., Kingsbury, B.: Deep neural networks
  for acoustic modeling in speech recognition: The shared views of four
  research groups. {IEEE} Signal Process. Mag.  \textbf{29}(6),  82--97 (2012)

\bibitem{HKWW17}
Huang, X., Kwiatkowska, M., Wang, S., Wu, M.: Safety verification of deep
  neural networks. In: Proceedings of the 29th International Conference on
  Computer Aided Verification. pp. 3--29 (2017)

\bibitem{JacobKCZTHAK18}
Jacob, B., Kligys, S., Chen, B., Zhu, M., Tang, M., Howard, A.G., Adam, H.,
  Kalenichenko, D.: Quantization and training of neural networks for efficient
  integer-arithmetic-only inference. In: Proceedings of the {IEEE} Conference
  on Computer Vision and Pattern Recognition. pp. 2704--2713 (2018)

\bibitem{julian2019deep}
Julian, K.D., Kochenderfer, M.J., Owen, M.P.: Deep neural network compression
  for aircraft collision avoidance systems. Journal of Guidance, Control, and
  Dynamics  \textbf{42}(3),  598--608 (2019)

\bibitem{jung2019learning}
Jung, S., Son, C., Lee, S., Son, J., Han, J.J., Kwak, Y., Hwang, S.J., Choi,
  C.: Learning to quantize deep networks by optimizing quantization intervals
  with task loss. In: Proceedings of the IEEE/CVF Conference on Computer Vision
  and Pattern Recognition. pp. 4350--4359 (2019)

\bibitem{KTSLSF14}
Karpathy, A., Toderici, G., Shetty, S., Leung, T., Sukthankar, R., Li, F.:
  Large-scale video classification with convolutional neural networks. In:
  Proceedings of 2014 {IEEE} Conference on Computer Vision and Pattern
  Recognition. pp. 1725--1732 (2014)

\bibitem{KBDJK17}
Katz, G., Barrett, C.W., Dill, D.L., Julian, K., Kochenderfer, M.J.: Reluplex:
  An efficient {SMT} solver for verifying deep neural networks. In: Proceedings
  of the 29th International Conference on Computer Aided Verification. pp.
  97--117 (2017)

\bibitem{KatzHIJLLSTWZDK19}
Katz, G., Huang, D.A., Ibeling, D., Julian, K., Lazarus, C., Lim, R., Shah, P.,
  Thakoor, S., Wu, H., Zeljic, A., Dill, D.L., Kochenderfer, M.J., Barrett,
  C.W.: The marabou framework for verification and analysis of deep neural
  networks. In: Proceedings of the 31st International Conference on Computer
  Aided Verification. pp. 443--452 (2019)

\bibitem{MNIST}
LeCun, Y., Cortes, C.: Mnist handwritten digit database (2010)

\bibitem{LiLYCHZ19}
Li, J., Liu, J., Yang, P., Chen, L., Huang, X., Zhang, L.: Analyzing deep
  neural networks with symbolic propagation: Towards higher precision and
  faster verification. In: Proceedings of the 26th International Symposium on
  Static Analysis. pp. 296--319 (2019)

\bibitem{li2021robust}
Li, J., Drummond, R., Duncan, S.R.: Robust error bounds for quantised and
  pruned neural networks. In: Proceedings of the 3rd Annual Conference on
  Learning for Dynamics and Control. pp. 361--372 (2021)

\bibitem{LiLHY0ZXH20}
Li, R., Li, J., Huang, C., Yang, P., Huang, X., Zhang, L., Xue, B., Hermanns,
  H.: Prodeep: a platform for robustness verification of deep neural networks.
  In: Proceedings of the 28th {ACM} Joint European Software Engineering
  Conference and Symposium on the Foundations of Software Engineering. pp.
  1630--1634 (2020)

\bibitem{LinTA16}
Lin, D.D., Talathi, S.S., Annapureddy, V.S.: Fixed point quantization of deep
  convolutional networks. In: Proceedings of the 33nd International Conference
  on Machine Learning. pp. 2849--2858 (2016)

\bibitem{lin2018defensive}
Lin, J., Gan, C., Han, S.: Defensive quantization: When efficiency meets
  robustness. In: Proceedings of the International Conference on Learning
  Representations (2019)

\bibitem{LXSSXM22}
Liu, J., Xing, Y., Shi, X., Song, F., Xu, Z., Ming, Z.: Abstraction and
  refinement: Towards scalable and exact verification of neural networks. CoRR
  \textbf{abs/2207.00759} (2022)

\bibitem{LiuSZW20}
Liu, W., Song, F., Zhang, T., Wang, J.: Verifying relu neural networks from a
  model checking perspective. J. Comput. Sci. Technol.  \textbf{35}(6),
  1365--1381 (2020)

\bibitem{LomuscioM17}
Lomuscio, A., Maganti, L.: An approach to reachability analysis for
  feed-forward {ReLU} neural networks. CoRR  \textbf{abs/1706.07351} (2017)

\bibitem{mistry2022milp}
Mistry, S., Saha, I., Biswas, S.: An {MILP} encoding for efficient verification
  of quantized deep neural networks. IEEE Transactions on Computer-Aided Design
  of Integrated Circuits and Systems (Early Access)  (2022)

\bibitem{MohammadinejadP21}
Mohammadinejad, S., Paulsen, B., Deshmukh, J.V., Wang, C.: {DiffRNN}:
  Differential verification of recurrent neural networks. In: Proceedings of
  the 19th International Conference on Formal Modeling and Analysis of Timed
  Systems. pp. 117--134 (2021)

\bibitem{moore2009introduction}
Moore, R.E., Kearfott, R.B., Cloud, M.J.: Introduction to interval analysis,
  vol.~110. Siam (2009)

\bibitem{nagel2020up}
Nagel, M., Amjad, R.A., Van~Baalen, M., Louizos, C., Blankevoort, T.: Up or
  down? adaptive rounding for post-training quantization. In: Proceedings of
  the International Conference on Machine Learning. pp. 7197--7206 (2020)

\bibitem{nagel2021white}
Nagel, M., Fournarakis, M., Amjad, R.A., Bondarenko, Y., van Baalen, M.,
  Blankevoort, T.: A white paper on neural network quantization. arXiv preprint
  arXiv:2106.08295  (2021)

\bibitem{Nakamura05}
Nakamura, A.: An efficient query learning algorithm for ordered binary decision
  diagrams. Inf. Comput.  \textbf{201}(2),  178--198 (2005)

\bibitem{narodytska2018verifying}
Narodytska, N., Kasiviswanathan, S.P., Ryzhyk, L., Sagiv, M., Walsh, T.:
  Verifying properties of binarized deep neural networks. In: Proceedings of
  the {AAAI} Conference on Artificial Intelligence. pp. 6615--6624 (2018)

\bibitem{OdenaOAG19}
Odena, A., Olsson, C., Andersen, D.G., Goodfellow, I.J.: Tensorfuzz: Debugging
  neural networks with coverage-guided fuzzing. In: Proceedings of the 36th
  International Conference on Machine Learning. pp. 4901--4911 (2019)

\bibitem{paulsen2020reludiff}
Paulsen, B., Wang, J., Wang, C.: Reludiff: Differential verification of deep
  neural networks. In: 2020 IEEE/ACM 42nd International Conference on Software
  Engineering (ICSE). pp. 714--726. IEEE (2020)

\bibitem{PaulsenWWW20}
Paulsen, B., Wang, J., Wang, J., Wang, C.: {NeuroDiff:} scalable differential
  verification of neural networks using fine-grained approximation. In:
  Proceedings of the 35th {IEEE/ACM} International Conference on Automated
  Software Engineering. pp. 784--796 (2020)

\bibitem{PCYJ17}
Pei, K., Cao, Y., Yang, J., Jana, S.: Deepxplore: Automated whitebox testing of
  deep learning systems. In: Proceedings of the 26th Symposium on Operating
  Systems Principles. pp. 1--18 (2017)

\bibitem{PT10}
Pulina, L., Tacchella, A.: An abstraction-refinement approach to verification
  of artificial neural networks. In: Proceedings of the 22nd International
  Conference on Computer Aided Verification. pp. 243--257 (2010)

\bibitem{ddlearning19B}
Shih, A., Darwiche, A., Choi, A.: Verifying binarized neural networks by
  angluin-style learning. In: Proceedings of the 2019 International Conference
  on Theory and Applications of Satisfiability Testing. pp. 354--370 (2019)

\bibitem{ddlearning19A}
Shih, A., Darwiche, A., Choi, A.: Verifying binarized neural networks by local
  automaton learning. In: Proceedings of the AAAI Spring Symposium on
  Verification of Neural Networks (2019)

\bibitem{SinghGPV19}
Singh, G., Ganvir, R., P{\"{u}}schel, M., Vechev, M.T.: Beyond the single
  neuron convex barrier for neural network certification. In: Proceedings of
  the Annual Conference on Neural Information Processing Systems. pp.
  15072--15083 (2019)

\bibitem{SGPV19}
Singh, G., Gehr, T., P{\"{u}}schel, M., Vechev, M.T.: An abstract domain for
  certifying neural networks. Proceedings of the ACM on Programming Languages
  (POPL)  \textbf{3},  41:1--41:30 (2019)

\bibitem{SongLCFL21}
Song, F., Lei, Y., Chen, S., Fan, L., Liu, Y.: Advanced evasion attacks and
  mitigations on practical ml-based phishing website classifiers. Int. J.
  Intell. Syst.  \textbf{36}(9),  5210--5240 (2021)

\bibitem{TianPJR18}
Tian, Y., Pei, K., Jana, S., Ray, B.: Deeptest: automated testing of
  deep-neural-network-driven autonomous cars. In: Proceedings of the 40th
  International Conference on Software Engineering. pp. 303--314 (2018)

\bibitem{TranBXJ20}
Tran, H., Bak, S., Xiang, W., Johnson, T.T.: Verification of deep convolutional
  neural networks using imagestars. In: Proceedings of the International
  Conference on Computer Aided Verification. pp. 18--42 (2020)

\bibitem{TranLMYNXJ19}
Tran, H., Lopez, D.M., Musau, P., Yang, X., Nguyen, L.V., Xiang, W., Johnson,
  T.T.: Star-based reachability analysis of deep neural networks. In:
  Proceedings of the 3rd World Congress on Formal Methods. pp. 670--686 (2019)

\bibitem{WPWYJ18}
Wang, S., Pei, K., Whitehouse, J., Yang, J., Jana, S.: Formal security analysis
  of neural networks using symbolic intervals. In: Proceedings of the 27th
  {USENIX} Security Symposium. pp. 1599--1614 (2018)

\bibitem{FSDChip}
WikiChip: Fsd chip - tesla.
  \url{https://en.wikichip.org/wiki/tesla_(car_company)/fsd_chip} (Accessed
  April 30, 2022)

\bibitem{YLLHWSXZ20}
Yang, P., Li, R., Li, J., Huang, C., Wang, J., Sun, J., Xue, B., Zhang, L.:
  Improving neural network verification through spurious region guided
  refinement. In: Groote, J.F., Larsen, K.G. (eds.) Proceedings of 27th
  International Conference on Tools and Algorithms for the Construction and
  Analysis of Systems. pp. 389--408 (2021)

\bibitem{ZhangHML22}
Zhang, J.M., Harman, M., Ma, L., Liu, Y.: Machine learning testing: Survey,
  landscapes and horizons. {IEEE} Trans. Software Eng.  \textbf{48}(2),  1--36
  (2022)

\bibitem{toolWEB}
Zhang, Y., Song, F., Sun, J.: {QEBVerif}.
  \url{https://github.com/S3L-official/QEBVerif} (2023)

\bibitem{BDD4BNN}
Zhang, Y., Zhao, Z., Chen, G., Song, F., Chen, T.: {BDD4BNN:} {A} {BDD}-based
  quantitative analysis framework for binarized neural networks. In:
  Proceedings of the 33rd International Conference on Computer Aided
  Verification. pp. 175--200 (2021)

\bibitem{ZZCSC23}
Zhang, Y., Zhao, Z., Chen, G., Song, F., Chen, T.: Precise quantitative
  analysis of binarized neural networks: A bdd-based approach. ACM Transactions
  on Software Engineering and Methodology  \textbf{32}(3) (2023)

\bibitem{zhang2022qvip}
Zhang, Y., Zhao, Z., Chen, G., Song, F., Zhang, M., Chen, T., Sun, J.: {QVIP}:
  An {ILP}-based formal verification approach for quantized neural networks.
  In: Proceedings of the 37th IEEE/ACM International Conference on Automated
  Software Engineering. pp. 82:1--82:13 (2023)

\bibitem{ZhaoCWYS021}
Zhao, Z., Chen, G., Wang, J., Yang, Y., Song, F., Sun, J.: Attack as defense:
  characterizing adversarial examples using robustness. In: Proceedings of the
  30th {ACM} {SIGSOFT} International Symposium on Software Testing and
  Analysis. pp. 42--55 (2021)

\bibitem{ZZCSCL22}
Zhao, Z., Zhang, Y., Chen, G., Song, F., Chen, T., Liu, J.: {CLEVEREST:}
  accelerating cegar-based neural network verification via adversarial attacks.
  In: Proceedings of the 29th International Symposium on Static Analysis. pp.
  449--473 (2022)

\end{thebibliography}
\appendix
 \clearpage

\section{An illustrating example of {\scshape DeepPoly}}\label{app_sec:deepPoly}

Consider the DNN $\mathcal{N}_e$ given in Fig.~\ref{fig:nnVS}. Neuron $\bs{x}^2_1$ (resp. $\bs{x}^2_2$) can be seen as two nodes $\bs{x}^2_{1,0}$ and $\bs{x}^2_{1,1}$ (resp. $\bs{x}^2_{2,0}$ and $\bs{x}^2_{2,1}$). Consider input region  $R((0.6,0.4),0.2)=\{(\bs{x}^1_1,\bs{x}^1_2)\in \mathbb{R}^2 \mid 0.4\le \bs{x}^1_1 \le 0.8, 0.2\le \bs{x}^1_2 \le 0.6\}$. Lower/upper bounds for the input variables are: $l^1_1=0.4$, $u^1_1=0.8$, $l^1_2=0.2$ and $u^1_2=0.6$.

We first get the lower/upper bound of nodes $\bs{x}^2_{1,0}$ and $\bs{x}^2_{2,0}$ in the form of linear expressions of the input variables: $\bs{a}_{1,0}^{2,\le}=\bs{a}_{1,0}^{2,\ge}=1.2\bs{x}^1_1-0.2\bs{x}^1_2$, $\bs{a}_{2,0}^{2,\le}=\bs{a}_{2,0}^{2,\ge}=-0.7\bs{x}^1_1+0.8\bs{x}^1_2$, and compute the concrete bounds as $l^2_{1,0}=1.2\times l^1_1 -0.2 u^1_2=0.36$, $u^2_{1,0}=1.2 u^1_1 -0.2 l^1_2=0.92$,
	$l^2_{2,0}=-0.7 u^1_1 + 0.8 l^1_2= -0.4$, $u^2_{2,0}=-0.7 l^1_1 + 0.8 u^1_2=0.2$.

Then, we can get the following abstract elements for nodes $\bs{x}^2_{1,1}$ and $\bs{x}^2_{2,1}$ based on the above ReLU transforms:
	\begin{itemize}
		\item $\bs{a}_{1,1}^{2,\le}=\bs{a}_{1,1}^{2,\ge}=\bs{x}^2_{1,0}$, $l^2_{1,1}=l^2_{1,0}=0.36$, $u^2_{1,1}=u^2_{1,0}=0.92$;
		\item $\bs{a}_{2,1}^{2,\le}=0$, $\bs{a}_{2,1}^{2,\ge}=\frac{1}{3}(\bs{x}^2_{2,0}+0.4)$, $l^2_{2,1}=0$, $u^2_{2,1}=u^2_{2,0}=0.2$;
\end{itemize}

Finally, we compute the lower/upper bounds for the output neuron as $l^3_0= 0.3 \bs{a}^{2,\le}_{1,1}+0.7 \bs{a}^{2,\le}_{2,1}= 0.3\bs{x}^2_{1,0}=0.3\bs{a}^{2,\le}_{1,0}=0.3l^2_{1,0}=0.108$, $u^3_0= 0.3 \bs{a}^{2,\ge}_{1,1}+0.7 \bs{a}^{2,\ge}_{2,1}= 0.3\bs{x}^2_{1,0}+\frac{0.7}{3}(\bs{x}^2_{2,0}+0.4)=0.3\bs{a}^{2,\ge}_{1,0}+\frac{0.7}{3}(\bs{a}^{2,\ge}_{2,0}+0.4)=\frac{0.59}{3}\bs{x}^1_1+\frac{0.38}{3}\bs{x}^1_2+\frac{0.28}{3}=\frac{0.59}{3} u^1_1+\frac{0.38}{3} u^1_2+\frac{0.28}{3}\approx 0.327$.

\section{Proof of Theorem~\ref{them:framework}}
\label{sec:proofthem:framework}
\begin{proof}
    Since Alg.~\ref{alg:AffTrs} is obviously sound, i.e.,
    the results over-approximate the underlying real output bounds, we next only need to prove that the activation transformation given in Alg.~\ref{alg:ActTrs} is also sound.

    Let $\Delta x=\text{clamp}(\tilde{\bs{y}}^i_j,0,t)-\text{ReLU}(\bs{y}^i_j)$ be the difference of neurons $\tilde{\bs{x}}^i_j$ and $\bs{x}^i_j$ from the QNN and DNN after applying the activation function. Note that, here we use $\tilde{\bs{y}}^i_j$ (resp. $\bs{y}^i_j$) to denote the value of the neuron before an activation function in QNN (resp. DNN). Let $\Delta=\tilde{\bs{y}}^i_j-\bs{y}^i_j$.

    First, consider \textbf{Case 1} ($\text{UB}(S^{in}(\bs{x}^i_j))\le 0$). The neuron of DNN is always  deactivated as 0. Hence, the output difference $\delta_{i,j}$ is $S(\tilde{\bs{x}}^i_j)$.

    Next, we consider the following cases when the neuron in DNN is always activated, i.e., $\text{LB}(S^{in}(\bs{x}^i_j))> 0$:
    \begin{itemize}
        \item \textbf{Case 2-1\&2-2} ($\text{UB}(\SinQ)\le 0$ or $\text{LB}(\SinQ)\ge t$): Since the neuron in QNN is always deactivated or clamped to $t$, we can get $\delta_{i,j}=-\SinD$ or $\delta_{i,j}=t-\SinD$.
        \item \textbf{Case 2-3} ($\text{LB}(\SinQ)\ge 0$ and $\text{UB}(\SinQ)\le t$): Both of two neurons in QNN and DNN are activated and not clamped. Therefore, we have $\delta_{i,j}=\delta^{in}_{i,j}$.
        \item \textbf{Case 2-4} ($\text{LB}(\SinQ)< 0$ and $0\le \text{UB}(\SinQ)\le t$): Since $\tilde{\bs{y}}^i_j$ is always smaller than $t$, we have $\Delta x=\text{max}(\tilde{\bs{y}}^i_j,0)-\bs{y}^i_j=\text{max}(\tilde{\bs{y}}^i_j-\bs{y}^i_j, -\bs{y}^i_j)$. Then, $\delta=\text{max}(\delta^{in}_{i,j},-\SinD)$.
        \item \textbf{Case 2-5} ($0\le \text{LB}(\SinQ)<t$ and $\text{UB}(\SinQ)> t$):  Since $\tilde{\bs{y}}^i_j$ is always larger than 0, we have $\Delta x=\text{min}(\tilde{\bs{y}}^i_j,t)-\bs{y}^i_j=\text{min}(\tilde{\bs{y}}^i_j-\bs{y}^i_j, t-\bs{y}^i_j)$. Hence, we have $\delta_{i,j}=\text{min}(\delta^{in}_{i,j},t-\SinD)$.
        \item \textbf{Case 2-6} (Otherwise): By case 2-4 \& 2-5, we have $\Delta x =\text{max}(\text{min}(\tilde{\bs{y}}^i_j,t),0)-\bs{y}^i_j=\text{max}(\text{min}(\Delta,t-\bs{y}^i_j), -\bs{y}^i_j)$. Hence, we have $\delta_{i,j}=\text{max}(\text{min}(\delta^{in}_{i,j},t-\SinD), -\SinD)$.
    \end{itemize}
    Finally, we consider the following cases when the neuron in DNN can be either activated or deactivated:
    \begin{itemize}
        \item \textbf{Case 3-1\&3-2} ($\text{UB}(\SinQ)\le 0$ or $\text{LB}(\SinQ)\ge t$): Similar to above, we have $\delta_{i,j}=-S(\bs{x}^i_j)$ or $t-S(\bs{x}^i_j)$ directly.
        \item \textbf{Case 3-3} ($\text{LB}(\SinQ)\ge 0$ and $\text{UB}(\SinQ)\le t$): $\Delta x ^i_j=\tilde{\bs{y}}^i_j-\text{max}(\bs{y}^i_j,0)=\tilde{\bs{y}}^i_j+\text{min}(-\bs{y}^i_j,0)=\text{min}(\tilde{\bs{y}}^i_j,\Delta)$. Then, we have $\delta_{i,j}=\text{min}(\SinQ,\delta^{in}_{i,j})$.

        \item \textbf{Case 3-4} ($\text{LB}(\SinQ)< 0$ and $0\le\text{UB}(\SinQ)\le t$): We rewrite $\Delta x $ as $\Delta x=\text{max}(\tilde{\bs{y}}^i_j,0)-\text{max}(\bs{y}^i_j,0)$:
        \begin{itemize}
            \item If $\Delta\le 0$, $\Delta x =\text{max}(\bs{y}^i_j+\Delta,0)-\text{max}(\bs{y}^i_j,0)\le 0$. Then, we have $\Delta x =0$ when $\bs{y}^i_j\le 0$, and $\Delta x =\text{max}(\bs{y}^i_j+\Delta, 0)-\bs{y}^i_j = \text{max}(\Delta, -\bs{y}^i_j)\le 0$ when $\bs{y}^i_j\ge 0$. Therefore, we have $\text{max}(\text{LB}(\delta^{in}_{i,j}), -\text{UB}(\SinD)) \le \Delta x \le 0$
            \item If $\Delta\ge 0$, $\Delta x =\text{max}(\tilde{\bs{y}}^i_j,0)-\text{max}(\tilde{\bs{y}}^i_j-\Delta,0)\ge 0$. Then, we have $\Delta x =0$ when $\tilde{\bs{y}}^i_j\le 0$, and $\Delta x =\tilde{\bs{y}}^i_j- \text{max}(\tilde{\bs{y}}^i_j-\Delta,0) = \tilde{\bs{y}}^i_j+ \text{min}(\Delta -\tilde{\bs{y}}^i_j,0)=\text{min}(\Delta, \tilde{\bs{y}}^i_j)\ge 0$ when $\bs{y}^i_j\ge 0$. Therefore, we have $0 \le \Delta x \le \text{min}(\text{UB}(\delta^{in}_{i,j}), \text{UB}(\SinQ))$
        \end{itemize}
        Thus we have $\text{LB}(\delta_{i,j})=0$ if $\text{LB}(\delta^{in}_{i,j})\ge 0$, and $\text{max}(\text{LB}(\delta^{in}_{i,j}), -\text{UB}(\SinD))$ otherwise. $\text{UB}(\delta_{i,j})=0$ if $\text{UB}(\delta^{in}_{i,j})\le 0$, and $\text{min}(\text{UB}(\delta^{in}_{i,j}), \text{UB}(\SinQ))$ otherwise.

        \item \textbf{Case 3-5} ($0\le\text{LB}(\SinQ)\le t$ and $\text{UB}(\SinQ)> t$): we rewrite $\Delta x $ as $\text{min}(\tilde{\bs{y}}^i_j,t)-\text{max}(\bs{y}^i_j,0)$. Then, $\Delta x =t-\text{max}(\bs{y}^i_j,0)=\text{min}(t-\bs{y}^i_j,t)$ when $\tilde{\bs{y}}^i_j \ge t$, and $\Delta x =\tilde{\bs{y}}^i_j - \text{max}(\tilde{\bs{y}}^i_j -\Delta,0) = \tilde{\bs{y}}^i_j + \text{min}(\Delta-\tilde{\bs{y}}^i_j,0) = \text{min}(\Delta,\tilde{\bs{y}}^i_j)$ when $\tilde{\bs{y}}^i_j \le t$. Specifically, when $\tilde{\bs{y}}^i_j \ge t$:
        \begin{itemize}
            \item If $\Delta\le t$, then $\bs{y}^i_j =\tilde{\bs{y}}^i_j - \Delta \ge 0$, and we  have $\Delta x =t-\bs{y}^i_j=t-\tilde{\bs{y}}^i_j +\Delta\le \Delta \le t$;
            \item If $\Delta\ge t$, then $\Delta x =t-\text{max}(\bs{y}^i_j,0)\le t \le \Delta$.
        \end{itemize}

        Therefore, we can have $t-\text{UB}(\SinD) \le  \Delta x  \le \{\Delta,t\}$ for $\tilde{\bs{y}}^i_j\le t$, and  $\text{min}(\text{LB}(\delta^{in}_{i,j}),\text{LB}(\SinQ))\le \Delta x  \le \{\Delta, t\}$ for $\tilde{\bs{y}}^i_j\le t$.
        Finally, we have $\text{LB}(\delta_{i,j}) = \text{min}\big(\text{LB}(\delta^{in}_{i,j}),\text{LB}(\SinQ),t-\text{UB}(\SinD)\big)$ and $\text{UB}(\delta_{i,j})=\text{min}\big(\text{UB}(\delta^{in}_{i,j}),t\big)$.

        \item \textbf{Case 3-6} (Otherwise):
        \begin{itemize}
            \item If $\tilde{\bs{y}}^i_j<0$, $\text{UB}(\delta_{i,j})=0$ by case 3-1.
            \item If $\tilde{\bs{y}}^i_j\ge 0$, $\text{UB}(\delta_{i,j})=\text{min}\big(\text{UB}(\delta^{in}_{i,j}),t\big)$ by case 3-5.
            \item If $\tilde{\bs{y}}^i_j\ge t$, $\text{LB}(\delta_{i,j})=t-\text{UB}(\SinD)$ by case 3-2;
            \item If $\tilde{\bs{y}}^i_j\le t$, $\text{LB}(\delta_{i,j})=0$ if $\text{LB}(\delta^{in}_{i,j})\ge 0$, and $\text{max}(\text{LB}(\delta^{in}_{i,j}), -\text{UB}(\SinD))$ otherwise by case 3-4.
        \end{itemize}
        Then, we get the lower bound as \[
        \text{LB}(\delta_{i,j})=\text{min}\big(t-\text{UB}(\SinD), 0, \text{max}(\text{LB}(\delta^{in}_{i,j}), -\text{UB}(\SinD))\big),
        \] and upper bound as \[\text{UB}(\delta_{i,j})=\text{max}\big(\text{min}(\text{UB}(\delta^{in}_{i,j}),t),0\big)=\text{clamp}(\text{UB}(\delta^{in}_{i,j}),0,t).\]
    \end{itemize}
    Hence, $[lb,ub]$ returned by Alg.~\ref{alg:ActTrs} is a sound result. Since $\big(S^{in}(\tilde{\bs{x}}^i_j) \cap [0,t]) - (S^{in}(\bs{x}^i_j)\cap [0,+\infty)\big)$ is also a sound difference interval.
    We conclude the proof.
\end{proof}

\section{Proof of Theorem~\ref{them:abQNN}}
\label{sec:proofthem:abQNN}
\begin{proof}
	If $\hat{l}^i_{j,1}\ge \up$, then $\hat{\bs{a}}_{j,2}^{i,\le}=\hat{\bs{a}}_{j,2}^{i,\ge}=\hat{\bs{x}}^i_{j,2}=\up$, thus
	$\up = \hat{l}^i_{j,2} = \hat{\bs{a}}_{j,2}^{i,\le} = \hat{\bs{x}}^i_{j,2} = \hat{\bs{a}}_{j,2}^{i,\ge} = \up =\hat{u}^i_{j,2}$.
	If $\hat{u}^i_{j,1}\le \up$, then $ \hat{l}^i_{j,2} = \hat{l}^i_{j,1} \le \hat{\bs{a}}_{j,1}^{i,\le} = \hat{\bs{a}}_{j,2}^{i,\le} \le \hat{\bs{x}}^i_{j,2} \le \hat{\bs{a}}_{j,2}^{i,\ge} =\hat{\bs{a}}_{j,1}^{i,\ge} \le \hat{u}^i_{j,1}= \hat{u}^i_{j,2}$.
	Otherwise, we have $\hat{l}^i_{j,1} < \up$, $\hat{u}^i_{j,1} > \up$, and therefore $\hat{l}^i_{j,2}= \hat{l}^i_{j,1} \le \hat{\bs{x}}^i_{j,2} \le \lambda \hat{u}^i_{j,1} + \mu = \hat{u}^i_{j,2}$.
\end{proof}

\section{An illustrating example of Symbolic-based DRA}\label{app_sec:sym_DRA}
%

Now we give illustration examples for interval-based DRA (cf. Section~\ref{sec:qveb}) and symbolic-based DRA (cf. Section~\ref{sec:abstract}), both of which are based on DNN $\mathcal{N}_e$ and QNN $\widehat{\mathcal{N}}_e$ given in Figure~\ref{fig:nnVS}. The quantization configurations for the weights, outputs of the input layer, and hidden layer are $\mathcal{C}_w=\langle \pm,4,2\rangle$, $\mathcal{C}_{in}=\langle +,4,4\rangle$ and $\mathcal{C}_h=\langle +,4,2\rangle$. We set the input region as $R((9,6),3)=\{(x,y)\in \mathbb{Z}^2 \mid 6\le x\le 12, 3\le y\le 9\}$.

First, we get the abstract element $\mathcal{A}^2_{j,s}=\langle \bs{a}^{2,\le}_{j,s}, \bs{a}^{2,\ge}_{j,s}, l^2_{j,s},u^2_{j,s} \rangle$ for $j\in\{1,2\}$ and $s\in\{0,1\}$ for DNN $\mathcal{N}_e$ as follows:
	\begin{itemize}
		\item $\bs{x}^2_1 \rightarrow \bs{x}^2_{1,0}, \bs{x}^2_{1,1}$:
		\begin{itemize}
		\item $\mathcal{A}^2_{1,0}=\langle 1.2\bs{x}^1_1 - 0.2\bs{x}^1_2, \ 1.2\bs{x}^1_1 - 0.2\bs{x}^1_2, \ 0.36, \ 0.92 \rangle$;
		\item $\mathcal{A}^2_{1,1}=\langle \bs{x}^2_{1,0},\ \bs{x}^2_{1,0},\ 0.36,\ 0.92 \rangle$.
		\end{itemize}

		\item $\bs{x}^2_2 \rightarrow \bs{x}^2_{2,0}, \bs{x}^2_{2,1}$:
		\begin{itemize}
		\item $\mathcal{A}^2_{2,0}=\langle -0.7\bs{x}^1_1 + 0.8\bs{x}^1_2, \ -0.7\bs{x}^1_1 + 0.8\bs{x}^1_2, \ -0.4, \ 0.2 \rangle$;
		
		\item $\mathcal{A}^2_{2,1}=\langle 0,\ \frac{1}{3}\bs{x}^2_{2,0} +\frac{0.4}{3},\ 0,\ 0.2 \rangle$.
		\end{itemize}
	\end{itemize}
	
	After substituting every variable in $\bs{a}^{2,\le}_{1,1}$, $\bs{a}^{2,\ge}_{1,1}$, $\bs{a}^{2,\le}_{2,1}$, and $\bs{a}^{2,\ge}_{2,1}$ until no further substitution is possible, we have the following forms of linear combination of the input variables:
	\begin{itemize}
		
		\item $\mathcal{A}^{2,\ast}_{1,0}=\langle 1.2\bs{x}^1_1 - 0.2\bs{x}^1_2, \ 1.2\bs{x}^1_1 - 0.2\bs{x}^1_2, \ 0.36, \ 0.92 \rangle$;
		\item $\mathcal{A}^{2,\ast}_{1,1}=\langle 1.2\bs{x}^1_1 - 0.2\bs{x}^1_2, \ 1.2\bs{x}^1_1 - 0.2\bs{x}^1_2, \ 0.36,\ 0.92 \rangle$;

		\item $\mathcal{A}^{2,\ast}_{2,0}=\langle -0.7\bs{x}^1_1 + 0.8\bs{x}^1_2, \  -0.7\bs{x}^1_1 + 0.8\bs{x}^1_2, \ -0.4, \ 0.2 \rangle$;
		
		\item $\mathcal{A}^{2,\ast}_{2,1}=\langle 0,\ -\frac{0.7}{3}\bs{x}^1_1 + \frac{0.8}{3}\bs{x}^1_2+\frac{0.4}{3},\ 0,\ 0.2 \rangle$.
	\end{itemize}

	Then,  we get the abstract element $\widehat{\mathcal{A}}^2_{j,p}=\langle \hat{\bs{a}}^{2,\le}_{j,p}, \hat{\bs{a}}^{2,\ge}_{j,p}, \hat{l}^2_{j,p},\hat{u}^2_{j,p} \rangle$ for $j\in\{1,2\}$ and $p\in\{0,1,2\}$ for QNN $\widehat{\mathcal{N}}_e$ as follows:

	\begin{itemize}
		\item $\hat{\bs{x}}^2_1 \rightarrow \hat{\bs{x}}^2_{1,0}, \hat{\bs{x}}^2_{1,1}, \hat{\bs{x}}^2_{1,2}$:
		\begin{itemize}
		\item $\widehat{\mathcal{A}}^2_{1,0}=\langle \frac{1}{16}(5\hat{\bs{x}}^1_1- \hat{\bs{x}}^1_2)-0.5, \ \frac{1}{16}(5\hat{\bs{x}}^1_1-
		\hat{\bs{x}}^1_2)+0.5,\ 0.8125,\ 4.0625 \rangle$;
		\item $\widehat{\mathcal{A}}^2_{1,1}=\langle \hat{\bs{x}}^2_{1,0},\ \hat{\bs{x}}^2_{1,0}, \ 0.8125,\ 4.0625\rangle$;
		\item $\widehat{\mathcal{A}}^2_{1,2}=\langle \hat{\bs{x}}^2_{1,1},\ \hat{\bs{x}}^2_{1,1}, \ 0.8125,\ 4.0625\rangle$.
		\end{itemize}

		\item $\hat{\bs{x}}^2_2 \rightarrow \hat{\bs{x}}^2_{2,0}, \hat{\bs{x}}^2_{2,1}, \hat{\bs{x}}^2_{2,2}$:
		\begin{itemize}
		\item $\widehat{\mathcal{A}}^2_{2,0}=\langle \frac{1}{16}(-3\hat{\bs{x}}^1_1+3 \hat{\bs{x}}^1_2)-0.5, \ \frac{1}{16}(-3\hat{\bs{x}}^1_1+3 \hat{\bs{x}}^1_2)+0.5, \ -2.1875, \ 1.0625 \rangle$;
		\item $\widehat{\mathcal{A}}^2_{2,1}=\langle 0, \frac{17(\hat{\bs{x}}^2_{2,0}+2.1875)}{52}, 0, 1.0625 \rangle$;
		\item $\widehat{\mathcal{A}}^2_{2,2}=\langle \hat{\bs{x}}^2_{2,1}, \ \hat{\bs{x}}^2_{2,1}, \ 0, \ 1.0625\rangle$.
		\end{itemize}

	\end{itemize}

	After substituting every variable in $\hat{\bs{a}}^{2,\le}_{1,1}$, $\hat{\bs{a}}^{2,\ge}_{1,1}$, $\hat{\bs{a}}^{2,\le}_{1,2}$, $\hat{\bs{a}}^{2,\ge}_{1,2}$, $\hat{\bs{a}}^{2,\le}_{2,1}$, $\hat{\bs{a}}^{2,\ge}_{2,1}$, $\hat{\bs{a}}^{2,\le}_{2,2}$, and $\hat{\bs{a}}^{2,\ge}_{2,2}$ until no further substitution is possible, we have the following forms of linear combination of the input variables:

	\begin{itemize}
		
		\item $\widehat{\mathcal{A}}^{2,\ast}_{1,0}=\langle \frac{1}{16}(5\hat{\bs{x}}^1_1- \hat{\bs{x}}^1_2)-0.5, \ \frac{1}{16}(5\hat{\bs{x}}^1_1-
		\hat{\bs{x}}^1_2)+0.5,\ 0.8125,\ 4.0625 \rangle$;
		\item $\widehat{\mathcal{A}}^{2,\ast}_{1,1}=\langle \frac{1}{16}(5\hat{\bs{x}}^1_1- \hat{\bs{x}}^1_2)-0.5, \ \frac{1}{16}(5\hat{\bs{x}}^1_1-
		\hat{\bs{x}}^1_2)+0.5,\ 0.8125,\ 4.0625 \rangle$;
		\item $\widehat{\mathcal{A}}^{2,\ast}_{1,2}=\langle \frac{1}{16}(5\hat{\bs{x}}^1_1- \hat{\bs{x}}^1_2)-0.5, \ \frac{1}{16}(5\hat{\bs{x}}^1_1-
		\hat{\bs{x}}^1_2)+0.5,\ 0.8125,\ 4.0625 \rangle$;

		\item $\widehat{\mathcal{A}}^{2,\ast}_{2,0}=\langle \frac{1}{16}(-3\hat{\bs{x}}^1_1+3 \hat{\bs{x}}^1_2)-0.5, \ \frac{1}{16}(-3\hat{\bs{x}}^1_1+3 \hat{\bs{x}}^1_2)+0.5, \ -2.1875, \ 1.0625  \rangle$;
		\item $\widehat{\mathcal{A}}^{2,\ast}_{2,1}=\langle 0, \frac{\frac{17}{16}(-3\hat{\bs{x}}^1_1 + 3 \hat{\bs{x}}^1_2)+45.6875}{52}, \ 0, \ 1.0625 \rangle$;
		\item $\widehat{\mathcal{A}}^{2,\ast}_{2,2}=\langle 0, \frac{\frac{17}{16}(-3\hat{\bs{x}}^1_1 + 3 \hat{\bs{x}}^1_2)+45.6875}{52}, \ 0, \ 1.0625 \rangle$;
	\end{itemize}

	Therefore, we have the lower bounds $\Delta l^{2,\ast}_{1,0}$, $\Delta l^{2,\ast}_{2,0}$, and upper bounds $\Delta u^{2,\ast}_{1,0}$, $\Delta u^{2,\ast}_{2,0}$ of the difference interval $\delta^{in}_{2,1}$, $\delta^{in}_{2,2}$ for the hidden neurons based on the input region as well as $\hat{\bs{x}}^1_j = 15 \bs{x}^1_j$ for $j\in\{1,2\}$ as follows:
	\begin{itemize}
		\item $\Delta l^{2,\ast}_{1,0} = 2^{-2}(\hat{\bs{a}}^{2,\le,\ast}_{1,0}) - \bs{a}^{2,\ge,\ast}_{1,0}= 2^{-2} \big( \frac{1}{16}(5\hat{\bs{x}}^1_1- \hat{\bs{x}}^1_2)-0.5\big) - (1.2\bs{x}^1_1 - 0.2\bs{x}^1_2 )$, and $\text{LB}(\Delta l^{2,\ast}_{1,0})= -0.168125$;
		\item $\Delta u^{2,\ast}_{1,0} = 2^{-2}(\hat{\bs{a}}^{2,\ge,\ast}_{1,0}) - \bs{a}^{2,\le,\ast}_{1,0} = 2^{-2} \big( \frac{1}	{16}(5\hat{\bs{x}}^1_1- \hat{\bs{x}}^1_2)+0.5\big) - (1.2\bs{x}^1_1 - 0.2\bs{x}^1_2)$, and $\text{UB}(\Delta u^{2,\ast}_{1,0}) = 0.106875$.
		\item $\Delta l^{2,\ast}_{2,0} = 2^{-2}(\hat{\bs{a}}^{2,\le,\ast}_{2,0}) - \bs{a}^{2,\ge,\ast}_{2,0}= 2^{-2} \big( \frac{1}{16}(-3\hat{\bs{x}}^1_1 +3\hat{\bs{x}}^1_2)-0.5\big) - (-0.7\bs{x}^1_1 + 0.8\bs{x}^1_2 )$, and $\text{LB}(\Delta l^{2,\ast}_{2,0}) = -0.185625$;
		\item $\Delta u^{2,\ast}_{2,0} = 2^{-2}(\hat{\bs{a}}^{2,\ge,\ast}_{2,0}) - \bs{a}^{2,\le,\ast}_{2,0} =2^{-2} \big( \frac{1}{16}(-3\hat{\bs{x}}^1_1 +3\hat{\bs{x}}^1_2)+0.5\big) - (-0.7\bs{x}^1_1 + 0.8\bs{x}^1_2 )$, and $\text{UB}(\Delta u^{2,\ast}_{2,0}) = 0.104375$.
	\end{itemize}

	Note that, based on above, we can compute $S^{in}(\hat{\bs{x}}^2_1)=[0.8125,3.0625]$ and $S^{in}(\hat{\bs{x}}^2_1)=[-2.1875,1.0625]$ via symbolic interval analysis on QNN $\widehat{\mathcal{N}}_e$. Then, according to Alg. 3, we have:
	\begin{itemize}
		\item $\delta_{2,1}^{in}=\delta_{2,1}=[-0.168125,0.106875]$;
		\item $\delta_{2,2}=[-0.185625,0.104375]$:
		\begin{itemize}
			\item $\text{LB}(\delta_{2,2})=\text{max}\big(\text{LB}(\delta^{in}_{2,2}), - \text{UB}(S^{in}(\bs{x}^2_2))\big)= \text{max}(-0.185625, -0.2)=-0.185625$;
			\item $\text{UB}(\delta_{2,2})=\text{min}\big(\text{UB}(\delta^{in}_{2,2}), \text{UB}(S^{in}(\tilde{\bs{x}}^2_2))\big)= \text{min}(0.104375, 1.0625/4)= 0.104375$.
		\end{itemize}
	\end{itemize}

	We remark that for the output layers in DNN $\mathcal{N}_e$ and QNN $\widehat{\mathcal{N}}_e$, we also have $\bs{x}^3_1 = \bs{x}^3_{1,0} = \bs{x}^2_{1,1}+3\bs{x}^2_{2,1}$ and $\hat{\bs{x}}^3_1 = \hat{\bs{x}}^3_{1,0}=\hat{\bs{x}}^2_{1,2}+3\hat{\bs{x}}^2_{2,2}$. Hence, for the output layer, we have:
	\begin{itemize}
		\item $\bs{a}^{3,\le,\ast}_{1,0} = 0.3\times \bs{a}^{2,\le,\ast}_{1,1} + 0.7\times \bs{a}^{2,\le,\ast}_{2,1}$, $\bs{a}^{3,\ge,\ast}_{1,0} = 0.3\times \bs{a}^{2,\ge,\ast}_{1,1} + 0.7\times \bs{a}^{2,\ge,\ast}_{2,1}$;
		\item $\hat{\bs{a}}^{3,\le,\ast}_{1,0} = 2^{-2}(\hat{\bs{a}}^{2,\le,\ast}_{1,2} + 3 \times \hat{\bs{a}}^{2,\le,\ast}_{2,2})$, $\hat{\bs{a}}^{3,\ge,\ast}_{1,0} = 2^{-2}(\hat{\bs{a}}^{2,\ge,\ast}_{1,2} + 3 \times \hat{\bs{a}}^{2,\ge,\ast}_{2,2})$.
	\end{itemize}

	Finally, we get the lower bound $\Delta l^{3,\ast}_{1,0}$ and upper bound $\Delta u^{3,\ast}_{1,0}$ for the output neurons as follows:
	\begin{itemize}
		\item $\Delta l^{3,\ast}_{1,0} = 2^{-2} (\hat{\bs{a}}^{3,\le,\ast}_{1,0})- \bs{a}^{3,\ge,\ast}_{1,0} = 2^{-4} (\hat{\bs{a}}^{2,\le,\ast}_{1,2} + 3 \times \hat{\bs{a}}^{2,\le,\ast}_{2,2})- (0.3\times \bs{a}^{2,\ge,\ast}_{1,1} + 0.7\times \bs{a}^{2,\ge,\ast}_{2,1})$, and $\text{LB}(\Delta l^{3,\ast}_{1,0}) = -0.19721875$;
		\item $\Delta u^{3,\ast}_{1,0} = 2^{-2} (\hat{\bs{a}}^{3,\ge,\ast}_{1,0})- \bs{a}^{3,\le,\ast}_{1,0} = 2^{-4} (\hat{\bs{a}}^{2,\ge,\ast}_{1,2} + 3 \times \hat{\bs{a}}^{2,\ge,\ast}_{2,2})- (0.3\times \bs{a}^{2,\le,\ast}_{1,1} + 0.7\times \bs{a}^{2,\le,\ast}_{2,1})$, and $\text{UB}(\Delta u^{3,\ast}_{1,0}) = 0.2045$.
	\end{itemize}
	
\smallskip
\noindent
{\bf Analysis of Results.}
For this verification task, we get different output intervals using two methods: i.e., $[-0.24459375, 0.117625]$ with interval-based DRA and $[-0.19721875, 0.2045]$ with symbolic-based DRA.
Note that, although the output quantization error interval computed by symbolic-based DRA is looser than that by the interval-based one, the symbolic-based method return preciser results for the hidden neurons, and such a comparison result is quite similar to the cases of P1-8 and P1-10 in Table~\ref{tab:DRA}.

\begin{table}[t]
	\centering
	\caption{Differential Reachability Analysis of Larger Attack Radius on MNIST.}
	\label{tab:largerDRA}
	\setlength{\tabcolsep}{1pt} 
	\scalebox{0.59}{
		\begin{tabular}{c|c|ccc|ccc|ccc|ccc|ccc}
			\toprule
			
			~ &  ~ &\multicolumn{3}{c|}{P1, $r=6$}&\multicolumn{3}{c|}{P2, $r=6$}&\multicolumn{3}{c|}{P3, $r=6$}&\multicolumn{3}{c|}{P4, $r=6$}&\multicolumn{3}{c}{P5, $r=6$}\\
			\multirow{-2}*{$Q$} &  \multirow{-2}*{Method} & H\_Diff & O\_Diff & \#S/T & H\_Diff & O\_Diff & \#S/T & H\_Diff & O\_Diff & \#S/T & H\_Diff & O\_Diff & \#S/T & H\_Diff & O\_Diff & \#S/T \\ \midrule
			
			\cellcolor{white} &  Naive & 125.9 & 14.26 & 9/0.87 & 430.7 & 40.06 & 0/1.70  & 866.8 & 60.68 & 0/2.80 & 988.1 & 47.84 & 0/13.06 & 10,661 & 256.1 & 0/146.3 \\
			
			\rowcolor{gray!20}
			\cellcolor{white}{4} & \tool (Int)  &  56.06 & 12.14 & 25/0.85 & 371.2 & 40.06 & 0/1.73  & 819.8 & 60.68 & 0/2.81 & 923.7 & 47.84 & 0/13.19 & 10,533 & 256.1 & 0/147.8 \\
			
			\cellcolor{white} & \tool (Sym)  & 57.71 & 6.07 & \hl{68/1.51} & 262.4 & 16.51 & \hl{1/3.48} & 654.8 & 44.57 & 0/6.37 & 798.8 & 25.87 & \hl{1/27.69} &8,985 & 478.0 & 0/476.4 \\
			\midrule
			
			\cellcolor{white} &  Naive & 134.8& 16.0 & 5/0.84 & 477.4 & 46.80 & 0/1.71 & 932.8 & 64.73 & 0/2.78 & 1,321 & 69.48 & 0/13.07 & 12,444 & 324.8 & 0/145.1 \\
			
			\rowcolor{gray!20}
			\cellcolor{white}{6} & \tool (Int)  & 18.25&3.99& 93/0.86 & 182.8 & 27.15 & 0/1.73 & 681.9 & 64.73 & 0/2.81 & 748.4 & 50.95 & 0/13.09 & 11,932& 324.8 & 0/145.4 \\
			
			\cellcolor{white}  &  \tool (Sym)  & 18.04 & 3.71 & \hl{96/1.51}  & 138.3 &  9.04 & \hl{45/3.32} & 386.1 & 18.88 & \hl{3/5.72} & 451.2 & 12.86 & \hl{36/26.10} & 7,139 & 331.9 & 0/416.9 \\
			\midrule
			
			\cellcolor{white} &  Naive & 135.4 &  16.13 & 4/0.84  & 482.3 & 47.58 & 0/1.71  & 941.2 & 65.49 & 0/2.79 & 1,343 & 71.28 & 0/13.03 & 12,567 & 329.4 & 0/142.8 \\
			
			\rowcolor{gray!20}
			\cellcolor{white}{8} &\tool (Int)   & 7.97& 1.68& \hl{126/0.84} & 86.18 & 13.55 & 11/1.72 & 550.7 & 63.20 & 0/2.80 & 368.5 & 25.80 & 2/13.12 & 11,796 & 329.4 & 0/144.1 \\
			
			\cellcolor{white} &   \tool (Sym)  & 7.67  & 3.09 & 105/1.51 & 105.2 & 7.32 & \hl{63/3.26} & 303.8 & 14.23 & \hl{14/5.51} & 342.5 & 9.72 & \hl{60/25.42} & 6165 & 263.4 & 0/380.2 \\
			\midrule
			
			\cellcolor{white} &  Naive & 135.4 & 16.14 & 5/0.85  & 483.0 & 47.66 & 0/1.70 & 942.4 & 65.64 & 0/2.80 & 1,342 & 71.25 & 0/13.04 & 12,592 & 330.2 & 0/143.0 \\
			
			\rowcolor{gray!20}
			\cellcolor{white}{10} & \tool (Int) & 5.38 & 1.10 & \hl{131/0.85}  & 59.88 & 9.74 & 31/1.71 & 427.7 & 53.27 & 0/2.81 & 250.3 & 17.95 & 4/13.10 & 11,747 & 330.2 & 0/143.2 \\
			
			\cellcolor{white} &  \tool (Sym) & 5.12 & 2.95 & 107/1.49  & 96.95 & 6.91 & \hl{68/3.22}  & 282.6 & 13.14  & \hl{18/5.53} & 315.8 & 8.98 & \hl{64/25.55} & 5,911 & 249.7 & 0/373.2 \\
			
			\midrule \midrule
			
			~ &  ~ &\multicolumn{3}{c|}{P1, $r=10$}&\multicolumn{3}{c|}{P2, $r=10$}&\multicolumn{3}{c|}{P3, $r=10$}&\multicolumn{3}{c|}{P4, $r=10$}&\multicolumn{3}{c}{P5, $r=10$}\\
			\multirow{-2}*{$Q$} &  \multirow{-2}*{Method} & H\_Diff & O\_Diff & \#S/T & H\_Diff & O\_Diff & \#S/T & H\_Diff & O\_Diff & \#S/T & H\_Diff & O\_Diff & \#S/T & H\_Diff & O\_Diff & \#S/T \\ \midrule
			
			\cellcolor{white} &  Naive &  206.0 &  24.54 & 0/0.93 & 684.6 & 62.99 & 0/1.95 & 1,289 & 78.86 & 0/3.33 & 1,787 & 87.43 & 0/15.83 & 56,648 & 2,146 & 0/198.7 \\
			
			\rowcolor{gray!20}
			\cellcolor{white}{4} & \tool (Int)  & 90.51 & 20.16 & 0/0.94 & 586.3 & 62.99 & 0/1.96 & 1,212 & 78.86 & 0/3.36 & 1,684 & 87.43 & 0/16.11 & 56,454 & 2,146 & 0/198.9 \\
			
			\cellcolor{white} & \tool (Sym)  &  92.50 & 10.68 & \hl{31/1.69} & 461.5 & 32.15 & 0/4.05 & 1,085 & 92.70 & 0/7.54 & 1,469 & 51.53 & 0/33.09 & 55,814 & 2,687 & 0/579.7 \\
			\midrule
			
			\cellcolor{white} &  Naive & 219.8 & 27.27 & 0/0.92 &  733.1 &  69.62 & 0/1.97  & 1,366 & 82.52 & 0/3.32 & 2,331 & 123.2 & 0/15.96 & 58,274 & 2,214 & 0/198.9 \\
			
			\rowcolor{gray!20}
			\cellcolor{white}{6} & \tool (Int)  &  30.26 & 6.70 & \hl{74/0.93} & 330.6 & 49.54 & 0/2.00 & 1,001 & 82.52 & 0/3.40 & 1,585 & 109.2 & 0/16.04 & 57,714 & 2,214 & 0/200.1 \\
			
			\cellcolor{white}  &  \tool (Sym)  &  29.84 & 8.10 & 61/1.68 & 311.4 & 22.82 & 0/3.88 & 885.2 & 60.29 & 0/7.02 & 1,121 & 38.62 & \hl{2/32.20} & 55,702 & 2,716 & 0/576.1 \\
			\midrule
			
			\cellcolor{white} &  Naive &  220.8 & 27.52 & 0/0.92 & 738.9 & 70.54 & 0/1.97 & 1,375 & 83.27 & 0/3.32 & 2,366 & 126.1 & 0/15.85 & 58,412 & 2,218 & 0/196.9 \\
			
			\rowcolor{gray!20}
			\cellcolor{white}{8} &\tool (Int)   &  13.81 & 2.91 & \hl{102/0.93} & 148.3 & 23.14 & 0/1.97 & 815.6 & 83.27 & 0/3.34 & 693.9 & 48.16 & 0/16.07 & 57,077 & 2,218 & 0/197.7 \\
			
			\cellcolor{white} &   \tool (Sym)  &  13.25 & 7.37 & 71/1.68 & 267.1 & 20.34 & 0/3.82 & 807.6 & 51.37 & 0/6.87 & 968.9 & 33.75 & \hl{3/31.77} & 55,391 & 2,666 & 0/567.7 \\
			\midrule
			
			\cellcolor{white} &  Naive & 220.9 & 27.55 & 0/0.92 & 739.9 &  70.67 & 0/1.96 & 1,377 & 83.42 & 0/3.31  & 2,367 & 126.1 & 0/16.02 & 58,443 & 2,219 & 0/197.6 \\
			
			\rowcolor{gray!20}
			\cellcolor{white}{10} & \tool (Int) & 9.65 & 1.95 & \hl{119/0.93}  & 99.04 & 15.76 & \hl{10/1.98}  & 597.9 & 71.12 & 0/3.34  & 397.4 & 27.76 & 0/16.13 & 56,858 & 2,219 & 0/198.4 \\
			
			\cellcolor{white} &  \tool (Sym) &  9.18 & 7.21 & 76/1.69  & 256.1 & 19.74 & 1/3.84  & 786.3 & 49.15 & \hl{1/6.83}  & 929.4 & 32.47 & \hl{3/31.59} & 55,289 & 2,637 & 0/559.5 \\
			
			\bottomrule
		\end{tabular}
	}
\end{table}

\section{More Experiments of DRA on MNIST}\label{app_sec:minist_DRA}

In this section, we use the same benchmarks and  settings on MNIST dataset as in Section~\ref{sec:RQ1} but with larger radii to evaluate the efficiency and effectiveness of DRA component in \tool. 

Table~\ref{tab:largerDRA} shows the results. 
 We can observe that DRA in \tool can scale to larger radii, such as $r=6$ and $r=10$. \tool (Sym) still perform best with most tasks proved but with less efficiency. Furthermore, we find that the attack radius has little impact on the propagation time of the difference interval but significantly influences the analysis results. Nonetheless, \tool remains capable of accomplishing many tasks solely through DRA, even when $r=10$.

\begin{table}
	\centering
	\caption{Differential Reachability Analysis on the 16 input regions of ACAS Xu.}
	\label{tab:app_prop_dra}
	\scalebox{0.7}{
	 \begin{tabular}{c|c|ccc|ccc|ccc|ccc}
	  \toprule
	  ~ &  ~ &\multicolumn{3}{c|}{Region 1}&\multicolumn{3}{c|}{Region 2}&\multicolumn{3}{c|}{Region 3}&\multicolumn{3}{c}{Region 4}\\
	  \multirow{-2}*{$Q$} &  \multirow{-2}*{Method} & H\_Diff & O\_Diff & \#S/T & H\_Diff & O\_Diff & \#S/T & H\_Diff & O\_Diff & \#S/T & H\_Diff & O\_Diff & \#S/T \\ \midrule
	
	
	  \cellcolor{white} &  Naive & 8,876,043 & 30,942 & 0/1.31 & 4,749,652 & 19,902 & 0/1.43 & 841.2 & 1.58 & 0/1.00 & 957.4 & 0.22 & \hl{2/0.88} \\
	  \rowcolor{gray!20}
	  \cellcolor{white}{4} & \tool (Int) & 8,875,909 & 30,942 & 0/1.27 & 4,749,504 & 19,902 & 0/1.54 & 839.2 & 1.58 & 0/0.90 & 955.7 & 0.22 & \hl{2/0.92} \\
	
	  \cellcolor{white}& \tool (Sym) & 8,875,990 & 30,942 & 0/3.99 & 4,749,602 & 19,902 & 0/3.88 & 937.5 & 1.33 & 0/3.16 & 1,046 & 0.22 & \hl{2/3.20} \\
	  \midrule
	
	  \cellcolor{white} &  Naive & 8,876,098 & 30,942 & 0/1.31 & 4,749,710 & 19,902 & 0/1.60 & 890.7 & 2.33 & 0/1.08 & 998.7 & 0.47 & 0/0.87 \\
	  \rowcolor{gray!20}
	  \cellcolor{white}{6} & \tool (Int) & 8,875,823 & 30,942 & 0/1.49 & 4,729,938 & 19,902 & 0/1.58 & 869.8 & 2.33 & 0/1.04 & 976.3 & 0.47 & 0/0.97 \\
	  \cellcolor{white} &  \tool (Sym)  & 8,876,031 & 31,097 & 0/3.77 & 4,749,648 & 19,991 & 0/4.11 & 779.2 & 14.87 & 0/3.56 & 916.6 & 10.16 & 0/3.02 \\
	  \midrule
	
	  \cellcolor{white} &  Naive & 8,876,114 & 30,943 & 0/1.38 & 4,749,729 & 19,902 & 0/1.38 & 920.4 & 3.55 & 0/1.05 & 1,018 & 1.33 & 0/0.86 \\
	  \rowcolor{gray!20}
	  \cellcolor{white}{8} &\tool (Int)  & 8,873,253 & 30,943 & 0/1.46 & 4,729,521 & 19,902 & 0/1.53 & 861.1 & 3.55 & 0/1.04 & 948.4 & 1.33 & 0/1.02 \\

	  \cellcolor{white} & \tool (Sym)  & 8,876,030 & 31,758 & 0/4.01 & 4,749,649 & 20,329 & 0/3.76 & 542.1 & 5.45 & 0/2.70 & 775.6 & 13.10 & 0/2.80 \\
	  \midrule
	
	  \cellcolor{white} & Naive & 8,876,117 & 30,943 & 0/1.31 & 4,749,733 & 19,903 & 0/1.45 & 921.7 & 3.73 & 0/1.04 & 1,023 & 1.72 & 0/0.92 \\
	  \rowcolor{gray!20}
	  \cellcolor{white}{10} & \tool (Int) & 8,872,649 & 30,943 & 0/1.64 & 4,729,185 & 19,903 & 0/1.62 & 843.5 & 3.73 & 0/1.01 & 935.3 & 1.72 & 0/0.94 \\
	  \cellcolor{white} & \tool (Sym) & 8,876,030 & 31,985 & 0/4.19 & 4,749,649 & 20,367 & 0/4.16 & 319.7 & 1.45 & 0/2.49 & 564.2 & 3.83 & 0/2.48 \\ \midrule \midrule

      ~ &  ~ &\multicolumn{3}{c|}{Region 5}&\multicolumn{3}{c|}{Region 16}&\multicolumn{3}{c|}{Region 26}&\multicolumn{3}{c}{Region 7}\\
	  \multirow{-2}*{$Q$} &  \multirow{-2}*{Method} & H\_Diff & O\_Diff & \#S/T & H\_Diff & O\_Diff & \#S/T & H\_Diff & O\_Diff & \#S/T & H\_Diff & O\_Diff & \#S/T \\ \midrule
	
	
	  \cellcolor{white} &  Naive & 9,667 & 111.2 & 0/1.39 & 46276 & 461.6 & 0/1.51 & 40025 & 343.1 & 0/1.49 & 78,468 & 446.7 & 0/1.48 \\
	  \rowcolor{gray!20}
	  \cellcolor{white}{4} & \tool (Int) & 9,658 & 111.2 & 0/1.35 & 44,815 & 461.6 & 0/1.51 & 39,987 & 343.1 & 0/1.59 & 77,417 & 446.7 & 0/1.51 \\
	
	  \cellcolor{white} & \tool (Sym) & 9,756 & 250.0 & 0/3.70 & 46266 & 779.1 & 0/3.87 & 40,009 & 751.9 & 0/4.00 & 78,462 & 446.7 &  0/3.75 \\
	  \midrule
	
	  \cellcolor{white} &  Naive & 9,741 & 114.1 & 0/1.45 & 46,331 & 464.0 & 0/1.46 & 40,088 & 345.5 & 0/1.46 & 78,550 & 448.3 & 0/1.49 \\
	  \rowcolor{gray!20}
	  \cellcolor{white}{6} & \tool (Int) &  9,708 & 114.1 & 0/1.49 & 46,080 & 464.0 & 0/1.40 & 39,937 & 345.5 & 0/1.52 & 77,368 & 448.3 & 0/1.44 \\
	  \cellcolor{white} &  \tool (Sym)  & 9,707 & 328.6 & 0/3.89 & 46,254 & 595.5 & 0/3.96 & 39,988 & 478.9 & 0/4.39 & 78,562 & 459.5 & 0/3.87 \\
	  \midrule
	
	  \cellcolor{white} & Naive & 9,751 & 115.5 & 0/1.50 & 46,344 & 464.9 & 0/1.64 & 40,102 & 346.5 & 0/1.55 & 78,576 & 449.2 & 0/1.49 \\
	  \rowcolor{gray!20}
	  \cellcolor{white}{10} & \tool (Int) & 9,656 & 115.5 & 0/1.47 & 46,106 & 464.9 & 0/1.65 & 39,902 & 346.5 & 0/1.59 & 76,976 & 449.2 & 0/1.62 \\
	  \cellcolor{white} & \tool (Sym) & 9,597 & 133.1 & 0/3.84 & 46,227 & 629.3 & 0/4.03 & 39,953 & 496.5 & 0/3.95 & 78,564 & 494.3 & 0/4.03 \\
	  \midrule
	
	  \cellcolor{white} & Naive & 9,753 & 115.7 & 0/1.46 & 46,346 & 465.1 & 0/1.50 & 40,105 & 346.6 & 0/1.66 & 78,580 & 449.3 & 0/1.46 \\
	  \rowcolor{gray!20}
	  \cellcolor{white}{10} & \tool (Int) & 9,584 & 115.7 & 0/1.48 & 46,058 & 465.1 & 0/1.49 & 39,890 & 346.6 & 0/1.53 & 77,058 & 449.3 & 0/1.57 \\
	  \cellcolor{white} & \tool (Sym) & 9,529 & 125.8 & 0/3.58 & 46,221 & 629.5 & 0/4.16 & 39,941 & 476.7 & 0/3.84 & 78,564 & 489.5 & 0/4.17 \\ \midrule \midrule

      ~ &  ~ &\multicolumn{3}{c|}{Region 8}&\multicolumn{3}{c|}{Region 9}&\multicolumn{3}{c|}{Region 10}&\multicolumn{3}{c}{Region 11}\\
	  \multirow{-2}*{$Q$} &  \multirow{-2}*{Method} & H\_Diff & O\_Diff & \#S/T & H\_Diff & O\_Diff & \#S/T & H\_Diff & O\_Diff & \#S/T & H\_Diff & O\_Diff & \#S/T \\ \midrule
	
	
	  \cellcolor{white} &  Naive & 1,230,035 & 395.5 & 0/1.51 & 11,196 & 93.76 & 0/1.38 & 64,270 & 452.7 & 0/1.36 & 9,667 & 111.2 & 0/1.31 \\
	  \rowcolor{gray!20}
	  \cellcolor{white}{4} & \tool (Int) & 1,215,608 & 395.5 & 0/1.55 & 11,176 &  93.76 & 0/1.43 & 64,207 & 452.7 & 0/1.39 & 9,658 & 111.2 & 0/1.42 \\
	
	  \cellcolor{white}& \tool (Sym) & 1,230,016 & 395.5 & 0/3.96 & 11,209 & 93.76 & 0/4.21 & 64,267 & 673.9 & 0/4.11 & 9,756 & 249.7 & 0/3.75 \\
	  \midrule
	
	  \cellcolor{white} &  Naive & 1,230,088 & 395.5 & 0/1.55 & 11,242 & 95.51 & 0/1.53 & 64,347 & 452.9 & 0/1.39 & 9,741 & 114.1 & 0/1.30 \\
	  \rowcolor{gray!20}
	  \cellcolor{white}{8} &\tool (Int) & 1,206,959 & 395.5 & 0/1.57 & 11,155 & 95.51 & 0/1.52 & 64,228 & 452.9 & 0/1.46 & 9,708 & 114.1 & 0/1.42 \\

	  \cellcolor{white} & \tool (Sym)  & 1,230,023 & 395.5 & 0/4.10 & 11,134 & 115.1 & 0/3.88 & 64,247 & 497.1 & 0/3.79 & 9,707 & 328.6 & 0/3.58 \\
	  \midrule
	
	  \cellcolor{white} &  Naive & 1,230,102 & 395.5 & 0/1.49 & 11,256 & 97.14 & 0/1.51 & 64,360 & 454.5 & 0/1.44 & 9,751 & 115.5 & 0/1.31 \\
	  \rowcolor{gray!20}
	  \cellcolor{white}{8} &\tool (Int)  & 1,205,601 & 395.5 & 0/1.59 & 11,125 & 97.14 & 0/1.40 & 64,197 & 454.5 & 0/1.45 & 9,656 & 115.5 & 0/1.37 \\

	  \cellcolor{white} & \tool (Sym)  & 1,230,014 & 395.5 & 0/3.98 & 11,033 & 118.4 & 0/3.92 & 64,188 & 580.0 & 0/3.70 & 9,597 & 133.1 & 0/3.45 \\
	  \midrule
	
	  \cellcolor{white} & Naive & 1,230,104 & 395.6 & 0/1.51 & 11,258 & 97.39 & 0/1.54 & 64,363 & 454.6 & 0/1.35 & 9,753 & 115.7 & 0/1.31 \\
	  \rowcolor{gray!20}
	  \cellcolor{white}{10} & \tool (Int) & 1,205,191 & 395.6 & 0/1.22 & 11,086 & 97.39 & 0/1.40 & 64,152 & 454.6 & 0/1.35 & 9,584 & 115.7 & 0/1.44 \\
	  \cellcolor{white} & \tool (Sym) & 1,230,010 & 404.6 & 0/3.34 & 11,002 & 119.4 & 0/3.78 & 64,174 & 562.0 & 0/3.60 & 9,529 & 125.8 & 0/3.35  \\ \midrule \midrule

      ~ &  ~ &\multicolumn{3}{c|}{Region 12}&\multicolumn{3}{c|}{Region 13}&\multicolumn{3}{c|}{Region 14}&\multicolumn{3}{c}{Region 15}\\
	  \multirow{-2}*{$Q$} &  \multirow{-2}*{Method} & H\_Diff & O\_Diff & \#S/T & H\_Diff & O\_Diff & \#S/T & H\_Diff & O\_Diff & \#S/T & H\_Diff & O\_Diff & \#S/T \\ \midrule
	
	
	  \cellcolor{white} &  Naive & 197,086 & 2,220 & 0/1.38 & 308,389 & 3,097 & 0/1.37 & 24,661 & 502.3 & 0/1.35 & 13,267 & 187.4 & 0/1.39 \\
	  \rowcolor{gray!20}
	  \cellcolor{white}{4} & \tool (Int) & 196,819 & 2220 & 0/1.44 & 303,993 & 3097 & 0/1.46 & 24,643 & 502.3 & 0/1.42 & 13,249 & 187.4 & 0/1.41 \\
	
	  \cellcolor{white}& \tool (Sym) & 197,033 & 2,220 & 0/3.76 & 308,364 & 3,427 & 0/3.80 & 24,679 & 986.0 & 0/3.80 & 13,294 & 187.4 & 0/3.86 \\
	  \midrule
	
	  \cellcolor{white} &  Naive & 197,155 & 2,223 & 0/1.39 & 308,449 & 3,099 & 0/1.47 & 24,713 & 506.2 & 0/1.44 & 13,315 & 193.9 & 0/1.35 \\
	  \rowcolor{gray!20}
	  \cellcolor{white}{6} & \tool (Int) & 196,907 & 2,223 & 0/1.51 & 306,624 & 3,099 & 0/1.42 & 24,633 & 506.2 & 0/1.39 & 13,230 & 193.9 & 0/1.43 \\
	  \cellcolor{white} &  \tool (Sym)  & 197,083 & 2,552 & 0/3.73 & 308,393 & 3,905 & 0/3.88 & 24,645 & 1,280 & 0/3.69 & 13,258 & 274.4 & 0/3.88 \\
	  \midrule
	
	  \cellcolor{white} &  Naive & 197,167 & 2,225 & 0/1.40 & 308,462 & 3,100 & 0/1.37 & 24,724 & 506.8 & 0/1.33 & 13,327 & 193.1 & 0/1.35 \\
	  \rowcolor{gray!20}
	  \cellcolor{white}{8} &\tool (Int)  & 196,809 & 2,225 & 0/1.41 & 307,200 & 3,100 & 0/1.44 & 24,598 & 506.8 & 0/1.41 & 13,193 & 193.1 & 0/1.44 \\

	  \cellcolor{white} & \tool (Sym)  & 197,082 & 3,002 & 0/3.92 & 308,397 & 3,386 & 0/3.89 & 24,591 & 680.0 & 0/3.60 & 13,202 & 199.5 & 0/3.60 \\
	  \midrule
	
	  \cellcolor{white} & Naive & 197,171 & 2,225 & 0/1.43 & 308,465 & 3,100 & 0/1.46 & 24,727 & 507.0 & 0/1.35 & 13,330 & 193.3 & 0/1.37 \\
	  \rowcolor{gray!20}
	  \cellcolor{white}{10} & \tool (Int) & 196,777 & 2,225 & 0/1.51 & 307,136 & 3,100 & 0/1.41 & 24,559 & 507.0 & 0/1.44 & 13,135 & 193.3 & 0/1.43 \\
	  \cellcolor{white} & \tool (Sym) & 197,082 & 2,886 & 0/3.72 & 308,401 & 3,770 & 0/4.12 & 24,570 & 616.8 & 0/3.53 & 13,179 & 199.3 & 0/3.52 \\

	  \bottomrule
	 \end{tabular}
	}
\end{table}

\section{More Experiments of \tool on ACAS Xu}

In contrast to the adversarial input regions considered in the main text, in this section, we use DNNs and input regions generated from the 16 properties given in~\cite{WPWYJ18}, and for each property, we use one of the ACAS Xu networks as our benchmarks, resulting 16 input regions and 80 verification tasks for each quantization bit size, i.e., $\epsilon=\{0.05,0.1,0.2,0.3,0.4\}$ for each input region.

\subsection{Differential Reachability Analysis}\label{app_sec:acas_DRA}
We first evaluate our DRA algorithms on these 16*5*4=320 tasks. The experimental results are given in Table~\ref{tab:app_prop_dra}.
Columns (H\_Diff) and (O\_Diff) averagely give the sum ranges of the difference intervals of all the hidden neurons and the output neuron of the predicted class, respectively.

We observe that \tool only successfully proved tasks in Region 4 out of all 16 input regions. However, we find that either \tool(Int) or \tool(Sym)  produces a more accurate output interval for the hidden and output neurons than the naive method. Note that, the output intervals of hidden neurons computed by symbolic interval analysis for DNNs in ACAS Xu can be very large when the range of the input region is large. However, the output intervals of hidden neurons for QNN are always limited by the quantization grid limit, i.e., $[0,\frac{2^Q-1}{2^{Q-2}}]$ for $Q\in\{4,6,8,10\}$. Hence, the difference intervals computed by these three methods can be too conservative when the input region is large.

\subsection{Verification with \tool}\label{app_sec:acas_tool}
We next evaluate \tool on relatively small to medium sizes of input regions, i.e., Regions 3, 4, 5, 9, 11, and 15.

Table~\ref{tab:app_qebv_acas} shows the verification results of \tool within 1 hour per task. Columns (\#Verified) and (Time)
give the number of successfully verified tasks and the average verification time of all the solved tasks in seconds, respectively.
Columns (DRA) give the verification results of proved tasks using only the differential reachability analysis, i.e., \tool(Sym).
Columns (DRA + MILP) give the results by a full verification process in \tool as shown in Figure~\ref{fig:overview}, i.e., we first use DRA and then use MILP solving if DRA fails.
Columns (DRA + MILP + Diff) are similar to Columns (DRA + MILP) except that
linear constraints of the difference intervals of hidden neurons got from DRA are added into the MILP encoding.

We observe that, with an MILP-based verification method \tool further verifies more tasks on which DRA fails. Similar to results in Figure~\ref{fig:qebVerif}, the effectiveness of the added linear constraints of the difference intervals varies on the MILP solving efficiency on different tasks.

\section{Complete Experimental Results of Figure~\ref{fig:qebVerif}}\label{app_sec:complete_res}
Tables~\ref{tab:app_exp2_acas} and \ref{tab:app_exp2_mnist} below give the complete verification results for the experiments in Section~\ref{sec:exp2}. Rows (\#Verified) and (Time)
give the number of successfully verified tasks and the average verification time of all the solved tasks in seconds.
Recall that only the number of successfully
proved tasks is given for DRA due to its incompleteness.

\begin{table}[t]
	\centering
	\caption{Verification Results of \tool on the 6 regions of ACAS Xu, where T.O. denotes time out.}
	
	\label{tab:app_qebv_acas}
	\setlength{\tabcolsep}{3pt} 
	\scalebox{0.75}{
		\begin{tabular}{c|c|cccc|cccc|cccc}
			\toprule
			\multicolumn{2}{c|}{Region}  & \multicolumn{4}{c|}{3} & \multicolumn{4}{c|}{4} & \multicolumn{4}{c}{5} \\ \midrule
			\multicolumn{2}{c|}{Q} & 4 & 6 & 8 & 10 & 4 & 6 & 8 & 10 & 4 & 6 & 8 & 10  \\ \midrule
			
			\rowcolor{gray!20}
			\cellcolor{white} & \#Proved & 0 & 0 & 0 & 0 & 2 & 0 & 0 & 0 & 0 & 0 & 0 & 0 \\
			\cellcolor{white}{\multirow{-2}*{DRA}}  & Time & 3.03 & 2.81 & 2.80 & 2.48 & 3.23 & 2.94 & 2.87 & 2.53 & 3.79 & 3.72 & 3.52 & 3.48  \\ \midrule
			\rowcolor{gray!20}
			\cellcolor{white} & \#Verified & 5 & 2 & 0 & 0 & 3 & 3 & 2 & 3 & 5 & 4 & 3 & 3 \\
			\cellcolor{white}{\multirow{-2}*{DRA+MILP}}  & Time & 4.52 & 4.76 & -- & -- & 255.0 & 966.9 & 8.39 & 268.2 & 407.2 & 432.1 & 20.51 & 32.87 \\ \midrule
			
			\rowcolor{gray!20}
			\cellcolor{white}{DRA+MILP} & \#Verified & 5 & 2 & 0 & 0 & 3 & 5 & 2 & 3 & 5 & 5 & 2 & 3  \\
			\cellcolor{white}{+Diff}  & Time & 5.54 & 6.17 & -- & -- & 126.1 & 859.9 & 10.02 & 416.4 & 18.44 & 139.6 & 10.61 & 14.22 \\ \midrule \midrule
			
			\multicolumn{2}{c|}{Region}  & \multicolumn{4}{c|}{9} & \multicolumn{4}{c|}{11} & \multicolumn{4}{c}{15} \\ \midrule
			\multicolumn{2}{c|}{Q} & 4 & 6 & 8 & 10 & 4 & 6 & 8 & 10 & 4 & 6 & 8 & 10  \\ \midrule
			
			\rowcolor{gray!20}
			\cellcolor{white} & \#Proved & 0 & 0 & 0 & 0 & 0 & 0 & 0 & 0 & 0 & 0 & 0 & 0 \\
			\cellcolor{white}{\multirow{-2}*{DRA}}  & Time & 3.17 & 3.27 & 3.23 & 3.19 & 3.30 & 3.21 & 3.18 & 3.02 & 3.27 & 3.26 & 3.23 & 3.26 \\ \midrule
			\rowcolor{gray!20}
			\cellcolor{white} & \#Verified & 4 & 2 & 2 & 3 & 5 & 4 & 3 & 3 & 5 & 4 & 4 & 4 \\
			\cellcolor{white}{\multirow{-2}*{DRA+MILP}}  & Time & 758.5 & 107.0 & 36.15 & 1,045 & 335.2 & 362.3 & 17.41 & 26.93 & 12.21 & 80.92 & 1,146 & 55.67 \\ \midrule
			
			\rowcolor{gray!20}
			\cellcolor{white}{DRA+MILP} & \#Verified & 2 & 2 & 2 & 2 & 5 & 5 & 2 & 3 & 4 & 5 & 5 & 0 \\
			\cellcolor{white}{+Diff}  & Time & 28.78 & 26.95 & 26.67 & 186.8 & 15.54 & 111.8 & 9.01 & 12.34 & 49.64 & 3.47 & 3.50 & T.O. \\

			\bottomrule
		\end{tabular}
	}
\end{table}

\begin{table}
	\centering
	\caption{Verification Results of \tool on ACAS Xu.}
    \label{tab:app_exp2_acas}
	 \scalebox{0.8}{
	 \begin{tabular}{c|c|cccc|cccc|cccc}
	  \toprule
	   \multicolumn{2}{c|}{}  & \multicolumn{4}{c|}{$r=3$} & \multicolumn{4}{c|}{$r=6$} & \multicolumn{4}{c}{$r=13$} \\ \midrule
	  \multicolumn{2}{c|}{Q} & 4 & 6 & 8 & 10 & 4 & 6 & 8 & 10 & 4 & 6 & 8 & 10 \\ \midrule
	
	  \rowcolor{gray!20}
      \cellcolor{white} & \#Proved & 0 & 10 & 24 & 25 & 0 & 9 & 18 & 22 & 0 & 5 & 8 & 9  \\
	  \cellcolor{white}{\multirow{-2}*{DRA}}  & Time & 2.06 & 1.49 & 1.09 & 1.05 & 2.71 & 1.58 & 1.17 & 1.13 & 2.20 & 1.36 & 1.20 & 1.15  \\ \midrule
	  \rowcolor{gray!20}
      \cellcolor{white} & \#Verified & 25 & 25 & 25 & 25 & 25 & 25 & 21 & 23 & 14 & 6 & 8 & 9  \\
	  \cellcolor{white}{\multirow{-2}*{DRA+MILP}}  & Time & 3.79 & 4.28 & 58.54 & 1.05 & 44.69 & 750.0 & 539.5 & 17.11 & 1,034 & 2,994 & 1.20 & 1.15  \\ \midrule

      \rowcolor{gray!20}
      \cellcolor{white}{DRA+MILP} & \#Verified & 25 & 25 & 25 & 25 & 25 & 24 & 23 & 25 & 16 & 8 & 10 & 10 \\
	  \cellcolor{white}{+Diff}  & Time & 3.33 & 3.87 & 2.92 & 1.05 & 26.43 & 621.4 & 305.1 & 3.76 & 650.6 & 1,793 & 173.3 & 2.96 \\
	
	  \bottomrule
	 \end{tabular}
	 }\vspace{-10mm}
\end{table}

\begin{table}
	\centering
	\caption{Verification Results of \tool on MNIST.}
    \label{tab:app_exp2_mnist}
	\setlength{\tabcolsep}{4pt} 
	 \scalebox{0.6}{
	 \begin{tabular}{c|c|cccc|cccc|cccc|cccc}
	  \toprule
	   \multicolumn{2}{c|}{Arch}   &  \multicolumn{4}{c|}{P1} &  \multicolumn{4}{c|}{P2} & \multicolumn{4}{c|}{P3} & \multicolumn{4}{c}{P4}\\ \midrule
	  \multicolumn{2}{c|}{Q} & 4 & 6 & 8 & 10 & 4 & 6 & 8 & 10 & 4 & 6 & 8 & 10 & 4 & 6 & 8 & 10 \\ \midrule
	
	  \rowcolor{gray!20}
      \cellcolor{white} & \#Proved & 88 & 130 & 136 & 139 & 49 & 88 & 108 & 112 & 1 & 70 & 86 & 92 & 35 & 102 & 128 & 131 \\
	  \cellcolor{white}{\multirow{-2}*{DRA}}  & Time &  1.34 & 1.37 & 1.36 & 1.37 & 2.91 & 2.87 & 2.84 & 2.86 & 5.03 & 4.67 & 4.58 & 4.58 & 23.08 & 21.92 & 21.58 & 21.53 \\ \midrule
	  \rowcolor{gray!20}
      \cellcolor{white} & \#Verified &  150 & 148& 150 &150 &132 &138 &145 &145 & 86& 115 &122 &126 & 108 & 103 & 130 & 134  \\
	  \cellcolor{white}{\multirow{-2}*{DRA+MILP}}  & Time & 21.32& 2.56 &3.08 &42.88 &95.75 &46.13 &96.76 &134.6 & 171.6 & 197.1 &137.7 & 93.96 & 365.9 & 83.06 & 2,356 & 444.7 \\ \midrule

      \rowcolor{gray!20}
      \cellcolor{white}{DRA+MILP} & \#Verified & 150&149&150 &150 &133 &138 & 146 &145 &85 &119 &135 &138 & 113 & 108 & 145 & 147 \\
	  \cellcolor{white}{+Diff}  & Time & 31.50& 27.51 &3.45 &125.7 &143.4 &16.49 &19.50 &9.19 & 167.9& 152.7 &184.3 &309.5 & 274.6 & 698.7 & 412.2 & 94.36 \\

	  \bottomrule
	 \end{tabular}
	 }
 \vspace{-8mm}
\end{table}

%
%
%
%
\end{document}